\documentclass{article}

\newif\ifarxiv
\arxivtrue
\newif\ifnotarxiv
\notarxivfalse

\usepackage[round]{natbib}

\usepackage[utf8]{inputenc} 
\usepackage[T1]{fontenc}    
\usepackage{url}            
\usepackage{amsfonts}       
\usepackage{xcolor}         
\usepackage{fullpage}
\usepackage{xcolor}
\definecolor{DarkGreen}{rgb}{0.1,0.5,0.1}
\definecolor{DarkRed}{rgb}{0.5,0.1,0.1}
\definecolor{DarkBlue}{rgb}{0.1,0.1,0.5}
\definecolor{Gray}{rgb}{0.2,0.2,0.2}
\usepackage[pdftex]{hyperref}
\hypersetup{
    unicode=false,          
    pdftoolbar=true,        
    pdfmenubar=true,        
    pdffitwindow=false,      
    pdfnewwindow=true,      
    colorlinks=true,       
    linkcolor=DarkBlue,          
    citecolor=DarkBlue,        
    filecolor=DarkRed,      
    urlcolor=DarkBlue,          
    pdftitle={Allocation Requires Prediction Only if Inequality Is Low},
    pdfauthor={Ali Shirali, Rediet Abebe, Moritz Hardt},
}

\usepackage{amsmath,amssymb,amsthm}
\usepackage{mathtools}
\usepackage{centernot}                      
\usepackage{bbm}                            
\usepackage{proof-at-the-end}

\usepackage{booktabs}       
\usepackage{makecell}
\usepackage{multirow}
\usepackage{multicol}
\usepackage[capitalize,noabbrev]{cleveref}

\usepackage{graphicx}
\usepackage{subfigure}
\usepackage{standalone}
\usepackage{tikz}
\usetikzlibrary{patterns}
\usetikzlibrary{calc}

\usepackage{authblk}

\theoremstyle{plain}
\newtheorem{thm}{Theorem}[section]
\newtheorem{proposition}[thm]{Proposition}
\newtheorem{lemma}[thm]{Lemma}
\newtheorem{corollary}[thm]{Corollary}
\theoremstyle{definition}
\newtheorem{definition}[thm]{Definition}
\newtheorem{property}{Property}
\theoremstyle{remark}
\newtheorem{remark}[thm]{Remark}
\newtheorem{example}[thm]{Example}
\newtheorem{assumption}[thm]{Assumption}
\crefname{thm}{Theorem}{Theorems}

\usepackage{amsmath,amsfonts,bm}

\newcommand{\One}{\mathbbm{1}}
\def\vrho{{\bm{\rho}}}

\def\barT{{\overline{T}}}
\def\vT{{\bm{T}}}
\newcommand{\C}{\mathbb{C}}
\def\ogamma{{\overline{\gamma}}}
\def\ugamma{{\underline{\gamma}}}

\newcommand{\dif}{\mathop{}\!\mathrm{d}}

\makeatletter
\newcommand{\spx}[1]{%
  \if\relax\detokenize{#1}\relax
    \expandafter\@gobble
  \else
    \expandafter\@firstofone
  \fi
  {^{#1}}%
}
\makeatother

\newcommand\pd[3][]{\frac{\partial\spx{#1}#2}{\partial#3\spx{#1}}}

\newcommand{\od}[3][]{\frac{\dif\spx{#1}#2}{\dif#3\spx{#1}}}

\def\eqref#1{equation~\ref{#1}}

\def\1{\bm{1}}

\DeclareMathAlphabet{\mathsfit}{\encodingdefault}{\sfdefault}{m}{sl}
\SetMathAlphabet{\mathsfit}{bold}{\encodingdefault}{\sfdefault}{bx}{n}

\def\gD{{\mathcal{D}}}

\def\gG{{\mathcal{G}}}
\def\gH{{\mathcal{H}}}

\def\gK{{\mathcal{K}}}

\def\gN{{\mathcal{N}}}
\def\gO{{\mathcal{O}}}
\def\gP{{\mathcal{P}}}

\def\gR{{\mathcal{R}}}
\def\gS{{\mathcal{S}}}

\def\gU{{\mathcal{U}}}

\def\gW{{\mathcal{W}}}
\def\gX{{\mathcal{X}}}

\def\sH{{\mathbb{H}}}

\def\sT{{\mathbb{T}}}

\newcommand{\E}{\mathbb{E}}

\newcommand{\R}{\mathbb{R}}

\newcommand{\Var}{\mathrm{Var}}

\renewcommand{\textstyle}{}

\title{Allocation Requires Prediction Only if Inequality Is Low}

\author[1]{Ali Shirali}
\author[2]{Rediet Abebe\thanks{Equal contribution}}
\author[3]{Moritz Hardt\textsuperscript{\textasteriskcentered}}
\affil[1]{University of California, Berkeley}
\affil[2]{Harvard Society of Fellows}
\affil[3]{Max Planck Institute for Intelligent Systems, T\"ubingen and T\"ubingen AI Center}

\date{}

\begin{document}

\maketitle

\begin{abstract}

Algorithmic predictions are emerging as a promising solution concept for efficiently allocating societal resources. Fueling their use is an underlying assumption that such systems are necessary to identify individuals for interventions. We propose a principled framework for assessing this assumption: Using a simple mathematical model, we evaluate the efficacy of prediction-based allocations in settings where individuals belong to larger units such as hospitals, neighborhoods, or schools. We find that prediction-based allocations outperform baseline methods using aggregate unit-level statistics only when between-unit inequality is low and the intervention budget is high. Our results hold for a wide range of settings for the price of prediction, treatment effect heterogeneity, and unit-level statistics' learnability. Combined, we highlight the potential limits to improving the efficacy of interventions through prediction. 

\end{abstract}
\section{Introduction}
\label{sec:intro}

Predictive systems are emerging as a promising solution concept for allocating societal resources. Often, such systems output individual risk scores for undesirable outcomes like eviction, poor health, and school dropout to assist decision-makers in targeting individuals with greater precision. In doing so, an underlying assumption is that individual predictions are necessary for efficient identification. See, e.g., \citep{bruce2011track, cuccaro2017risk, kleinberg2018human}. 

By contrast, decision-makers can use baseline allocation schemes by proportioning their resources using aggregate welfare information such as neighborhood eviction, hospital readmission, or school dropout rates. These larger units (i.e., neighborhoods, hospitals, and schools) can then target individuals within the units crudely. While such baseline methods can be cheaper, easier to implement, and less contested, they are also feared to be wasteful compared to prediction-based allocations.\footnote{See the discussion in~\citep{johnson2022bureaucratic,moon2024human}.}

\begin{figure}[ht!]
    \centering
    \includegraphics{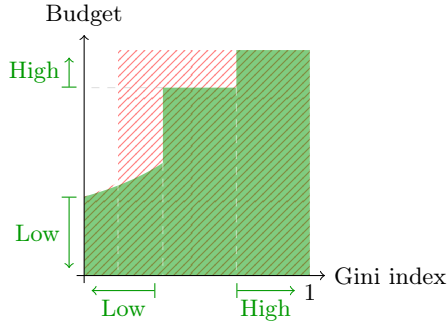}
    \caption{Sufficient conditions for a dominant unit-level allocation (green) and nondominated unit-level allocation (red). We define a high budget in relation to the cost of treating the whole population and a low budget in relation to the cost of prediction. See \cref{fig:ula_regimes} for a quantitative version of this figure.}
    \label{fig:ula_regimes_simplified}
\end{figure}

We propose a principled framework for assessing this fundamental assumption and evaluating the efficacy of individual prediction-based systems. We present a simple mathematical model to compare prediction-based versus unit-based allocations. The former method relies on individuals' predicted welfare, while the latter uses aggregate unit-level statistics and coarse information within the units. We show that prediction leads to superior allocations only when between-unit inequality is low, and the allocation budget is high. (See \cref{fig:ula_regimes_simplified} for a high-level view of inequality and budget regimes covered in our results.) Our analyses cover a broad range of settings for the price of prediction, treatment effect heterogeneity, and unit-level statistics' learnability. 

\subsection{Our Results}
At a high level, we contrast two allocation mechanisms: Individual-level allocation (ILA) predicts welfare, which falls in~$[0,1]$, and treats individuals in ascending order of their predicted welfare. Unit-level allocation (ULA), roughly speaking, sorts units according to their average welfare and intervenes in increasing order of average welfare. We assume administrators within each unit target the intervention to a fraction of the individuals. They are able to avoid giving resources to the top~$q \ge 0$ fraction, allowing for some errors. To start, we consider an intervention with a fixed treatment effect~$\delta$. The welfare is capped at~$1$ so individuals of welfare above $1-\delta$ see diminished effects. 

\paragraph{Individual-Level versus Unit-Level Allocations.} We characterize each unit~$k$ with a parameter~$\rho_k$ that expresses the fraction of high-welfare individuals in the unit. We measure inequality in terms of the Gini index of the $\rho_k$-parameters. Our main results establish the fundamental role of inequality in understanding the relative effectiveness of the mechanisms. We prove that high inequality ensures the relative efficiency of ULA as long as the budget is not excessively large. This is true even when ILA uses perfect predictions (after paying some upfront cost). 

What defines the high inequality regime depends on the fraction~$q$ and the average of $\rho_k$s, denoted by~$\bar{\rho}$. The quantity~$\bar{\rho}$ is also the overall proportion of individuals with high welfare (above $1-\delta$). Roughly speaking, a Gini index below $1 - q/\bar{\rho}$ is low, and a Gini index above $1 - q/(4\bar{\rho})$ is high. When inequality is neither high nor low, ULA will still be efficient for a budget up to a constant ratio of the cost of treating everyone. In all these cases, unless the cost of prediction is minuscule, ULA \emph{dominates} ILA (\cref{fig:ula_regimes_simplified}). 

Necessary conditions for ILA to dominate ULA turn out to be significantly stronger. Put differently, under mild conditions we can show that ILA does \emph{not} dominate ULA. For example, in the natural regime of a Gini index around $1/2$, ULA remains nondominated even if~$q=0$, that is, within-unit allocation treats everyone.

\paragraph{Extension to Heterogeneous Treatment Effects.}
In \cref{sec:het}, we extend our results to the case of heterogeneous treatment effects. We continue to assume that ILA uses welfare prediction as the targeting mechanism. Without further assumptions, the heterogeneity gives ILA an edge since distinguishing between individuals with close welfare through prediction can potentially yield significant gains. We limit this advantage by considering a Lipschitz continuous effect curve. Additionally, we make our analysis more realistic by restricting the concentration of welfare values. Both assumptions help avoid contrived worst-case scenarios.

Without heterogeneity, there exists a sufficient level of prediction accuracy such that more accurate predictors cannot generate more value. However, with heterogeneous treatment effects, even small changes in accuracy can be translated into value. Therefore, in the analysis of a heterogeneous treatment effect, we incorporate prediction accuracy into our framework by assuming a model of independently noisy predictions. 
Under these assumptions, we obtain similar results as in the homogeneous case.
  
\paragraph{Extension to Learning Unit-Level Statistics.}
ULA hinges on the availability of some unit-level aggregate statistics, such as $\rho_k$s, to prioritize different units. We generally assume the cost of obtaining these statistics is negligible. In \cref{sec:learning}, we revisit this assumption and place our results in a learning-theoretic context. Although some information about individuals may be available for free, their welfare may be costly to measure. Therefore, we explore how to use readily available data to learn unit-level statistics sufficient for implementing ULA. First, we demonstrate that if an individual-level predictor can be obtained at a reasonable cost, a predictor for unit-level statistics can be obtained at a much lower cost. Next, we consider different noisy observation welfare models and discuss efficient learnability. In particular, in a general nonparametric setting, the relative cost of learning becomes negligible if the number of treated units is sufficiently large.

\paragraph{Summary.}
Our results suggest that allocation by prediction may not compare favorably to basic allocation schemes that use aggregate unit statistics in cases where inequality exists, or resources are limited. Our evaluation framework and analyses surface inequality as a fundamental mechanism linking prediction and intervention. We hope that our theoretical framework provides the tools necessary to investigate prediction as a solution concept for allocation.
\subsection{Related Work}

ILA, as a form of resource allocation based on prediction, is widely adopted across various domains~\citep{kube2023community,kube2023fair,toros2018prioritizing,mashiat2024beyond,chan2012optimizing,mukhopadhyay2017prioritized,faria2017getting}. However, there is still minimal evidence of its efficacy~\citep{mac2019efficacy,moon2024human}. For instance, \citet{dews} recently demonstrated that early warning systems, when utilized to target interventions at individual students by predicting the risk of dropout, have shown little improvement despite the use of accurate predictors. \citet{dews} and \citet{hardt2023is} also point out how risk scores draw on unit-level information. Closely related, \citet{perdomo2023relative} asks how improvements in welfare arising from better predictions compare to those of other policy levers, such as expanding access to resources. In line with our results, it may not help investing in greater accuracy when resources are limited.

Refer to \cref{sec:additional_related} for a complete review of related work.

\section{Our Model}
\label{sec:model}

There are $M$~units with $N$~individuals each. Individual~$i$ has a welfare of~$w_i \in [0, 1]$.
Our goal is to improve social welfare, defined as the sum of individual welfares. To do so, we can intervene on any individual at a unit cost~$c$ subject to a total budget constraint~$B$. We may target individuals by predicting their current welfare and allocating resources to those with lower welfare. We call this mechanism \emph{individual-level allocation} (ILA). Alternatively, we may target units instead of individuals and delegate within-unit allocation to unit administrators, through what we call \emph{unit-level allocation} (ULA). 

\paragraph{The Cost of Prediction.}
We assume access to a welfare prediction~$\hat{w}_i$ for each individual~$i$ that satisfies the uniform error bound: $|\hat{w}_i - w_i| \le \epsilon$. We have to pay a total price~$p(\epsilon)$ to obtain such a predictor. The price~$p$ is a decreasing function of~$\epsilon$. Our results about the limitations of ILA hold even if we give ILA the advantage that after paying some fixed price, predictions are perfect. In other words, there is a fixed price of prediction~$p=p(\epsilon)$ for any $\epsilon\in[0,1)$.

\paragraph{Intervention Effect.} 
Let $\tilde{w}_i$ denote the counterfactual welfare of individual~$i$ had she received treatment. Then $\tau_i = \tilde{w}_i - w_i$ denotes the treatment effect. In our simplest model, we assume $\tilde{w}_i = \min \, \{w_i + \delta, 1\}$. This is the case of an intervention that has a fixed treatment effect~$\delta$ capped at the maximum welfare. Our results extend to heterogeneous treatment effects.

\paragraph{Individual-Level Allocation (ILA).}
The \emph{individual-level allocation} gives the resources to the individuals with the lowest estimated welfare~$\hat{w}_i$ until the budget constraint is met. Formally, sort individuals ascendingly in terms of~$\hat{w}_i$ with any tie-breaking and let $s(i)$ be the individual in the $i^\text{th}$~place. Given a budget of~$B$ and an $\epsilon$-error welfare predictor, we can treat $I = \lfloor \frac{B - p(\epsilon)}{c} \rfloor$~individuals. Treating the first $I$~individuals after sorting, the \emph{value} of ILA is
\begin{equation}
    V_\text{ind} = \textstyle\sum_{i \in [I]} \tau_{s(i)}
    \,.
\end{equation}

\paragraph{Unit-Level Allocation (ULA).}
The \emph{unit-level allocation} targets those units that are performing worst in terms of some aggregate measurement of their welfare. More precisely, we define a statistic~$\rho_k$ for unit~$k$ that captures the well-being of individuals in unit~$k$ and allocate the resources to the units with the smallest $\rho_k$-values.
Let~$\gU_k$ denote the set of individuals under unit~$k$. Define $\rho_k$ to be the proportion of individuals in $\gU_k$ with a welfare more than $1 - \delta,$ i.e., 
\begin{equation}
    \rho_k = \frac{1}{N} \textstyle\sum_{i \in \gU_k} \One\{w_i > 1 - \delta\}
    \,.
\end{equation}
We assume $\rho_k$ is known and freely available. We revisit this assumption in \cref{sec:learning}, where we show learning $\rho_k$s can be done efficiently with minimal impact on the allocation value. We denote the profile of all units with a vector~$\vrho$.

After choosing to intervene on a unit, the allocation of interventions within each unit will be handled by unit administrators, such as school officials. We need to assume that administrators are able to avoid a blatantly wasteful allocation of resources. Specifically, we assume administrators are able to avoid intervening on individuals of highest welfare most of the time. Quantitatively, the allocation avoids the top~$q$ fraction of individuals according to welfare, possibly misclassifying a $q'$ fraction of all individuals, where $q'\le q.$ For example, with~$N=100$ individuals, $q=0.25$, and $q'=0.1$, a total of $75$~individuals will receive resources. Out of these, at most $10$~individuals may erroneously belong to the top~$25$ highest welfare individuals.

Given a budget of~$B$, we can treat $K = \lfloor \frac{B}{N (1-q) c} \rfloor$~units. Let $T_k$ be the average treatment effect of unit~$k$. Sort the units ascendingly in terms of~$\rho_k$ with any tie-breaking and let $s(k)$ be the unit in the $k^\text{th}$~place. The \emph{value} of ULA is 
\begin{equation}
\label{eq:def_V_unit}
    V_\text{unit} = \textstyle\sum_{k \in [K]} N\, T_{s(k)}
    \,.
\end{equation}

\paragraph{Dominance Notion.}
Given two allocation mechanisms, such as ILA and ULA, the next question is how to compare them for a fixed available budget. Ideally, this comparison should be agnostic to the specific unit profile~$\vrho$, as long as $\vrho$ is within a set of unit profiles of our interest. Furthermore, we would like the comparison to be agnostic to how individual welfares are distributed within a unit~$k$, as long as welfares are consistent with~$\rho_k$. Here we define a dominance notion that accomplishes both desiderata.
\begin{definition}[Dominance notion]
A solution concept~$S_1$ (e.g., ULA) dominates another solution concept~$S_2$ (e.g., ILA) for a fixed budget~$B$ and for a set of feasible unit profiles~$\gP$, if for every profile~$\vrho \in \gP$, for every within-unit welfare distribution consistent with~$\vrho$, and for every tie-breaking in algorithms, the value of~$S_1$ is at least as the value of~$S_2$. We denote this dominance relation by $S_1 \succeq S_2$. If there exists a profile~$\vrho \in \gP$ for which $S_1$ has a larger value than~$S_2$, we say $S_1$ strictly dominates~$S_2$ and denote it by $S_1 \succ S_2$. We use $\alpha \cdot S_1 \succeq S_2$ to show that the comparison is made by multiplying the value of~$S_1$ by a constant~$\alpha$.
\end{definition}

This notion of dominance is a strong test if $\gP$ is broad. Nonetheless, as we discuss in \cref{sec:dominant_ula}, surprisingly ULA dominates ILA in this strong sense for a typical~$\gP$. For instance, in the presence of high inequality across units which is a common empirical observation in many settings, there will be enough units with very low~$\rho_k$ that can be targeted by ULA. On the other hand, as we discuss in \cref{sec:nondominated_ula}, for a sufficiently broad~$\gP$, ULA can never be dominated. Therefore, the dominance notion can provide a useful characterization of~$\gP$ for which one solution concept is dominant or nondominated assuming a typical~$\gP$.

\section{Illustrating the Key Idea}

\ifarxiv
We start with a visualization of the key idea behind our results, illustrating how ULA can outperform ILA in simulated and real-world settings. We will then introduce the underlying theory in a special case.

\subsection{Graphical Illustration}
Consider units of equal size where individual welfares are drawn from a beta distribution with unique parameters for each unit. \cref{fig:sim_beta} depicts such units, sorted according to their median welfare. Within each unit, individuals are sorted based on their welfare and plotted with their position on the horizontal axis and their welfare on the vertical axis. We have implemented a within-unit allocation with parameters~$q=0.3$ and $q'\approx0$ for simplicity. The top~$q$ fraction of individuals at each unit will be excluded during within-unit allocation and appear as faded in the plot. 

Suppose prediction costs $20\%$ of the budget. In the figure, we show the values realized by ULA and ILA from each unit. For simplicity, we neglect ULA's value from high-welfare individuals with $w > 1-\delta$. As the figure suggests, in the presence of high inequality between units, ULA with an effective within-unit allocation mostly avoids individuals with high welfare and is almost optimal. However, due to the costly prediction, ILA, despite more accurate targeting, cannot reach ULA's value.

\begin{figure}[h]
    \centering
    \includegraphics[width=0.95\linewidth]{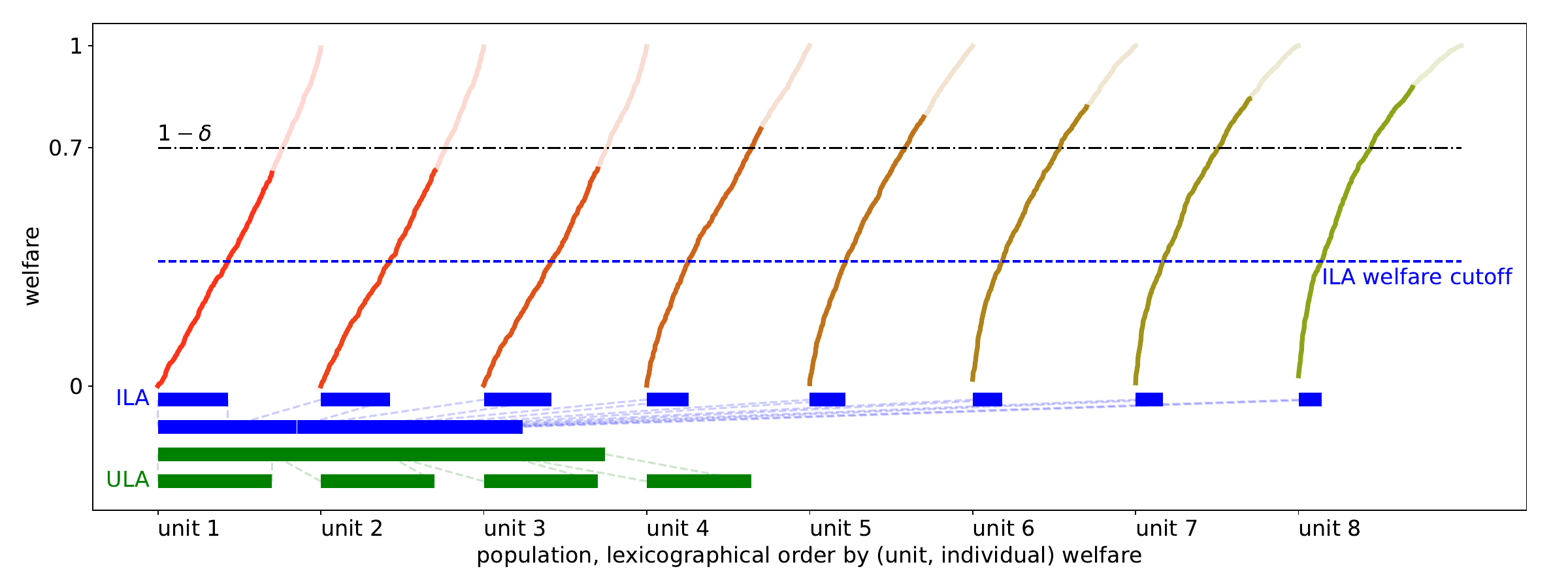}
    \caption{ULA outperforms ILA in a simulated setting with high inequality. The treatment effect is $\delta=0.3$, and half of the units are treated, with $20\%$ of the budget required for prediction. Within each unit, individuals have independent beta-distributed welfare. They are sorted based on their welfare and plotted according to their position and welfare. A within-unit allocation with $q=0.3$ and $q'\approx0$ is considered, and individuals among the top~$q$ fraction at each unit are faded. The values obtained by ILA and ULA from each unit are depicted with horizontal bars, and their total values compared. For ease of presentation, we assume ULA gets zero value from treating units above welfare $1-\delta.$}
    \label{fig:sim_beta}
\end{figure}

The observation replicates on real-world data. Here, we utilize the American Community Survey Children's Education Tabulation,\footnote{\url{https://nces.ed.gov/programs/edge/Demographic/ACS}} an annually updated custom data collection of demographic, economic, social, and housing characteristics about school-age children and their families, developed from the U.S. Census Bureau’s 2017-2021 data. We limit our analysis to certain school districts (units) and relevant students (individuals), namely those who reside within the territory of the district. As a proxy for welfare, we examine household income (inflation-adjusted), which is reported in 10 income brackets ranging from less than $\$10$k to more than $\$200$k. We consider the highest bracket as having a welfare of $0.95$ and the lowest bracket as having a welfare of $0.05$. 

\cref{fig:acs_ed_demo_LA} displays data from representative school districts of similar size in the greater Los Angeles (LA) area (excluding the LA Unified School District due to its sheer size). 
We consider a similar parameter setting as in the synthetic data example with $\delta=0.3$, $q=0.3$, $q'\approx0$, $p(\epsilon) = 0.2 \cdot B$, and a budget~$B$ sufficient to treat half of the units. As the figure suggests, in the presence of high inequality, as is the case for the chosen 8 districts, ULA largely avoids high-welfare students. In contrast, ILA with costly prediction cannot achieve comparable value. 

The superiority of ULA primarily stems from the high inequality between the different school districts. The same conclusions hold in other high-inequality regions. For example, \cref{fig:demo_NY} illustrates the same point for school districts in New York. Conversely, in regions of low inequality, ULA may not achieve the same value as ILA. To give an example, \cref{fig:demo_UT} shows that for Utah's school districts, ILA comes out ahead. When considering the Gini index of the $\rho_k$s, the inequality in the chosen New York state school districts is almost twice that of LA and four times that of Utah.

\begin{figure}[h]
    \centering
    \includegraphics[width=0.9\linewidth]{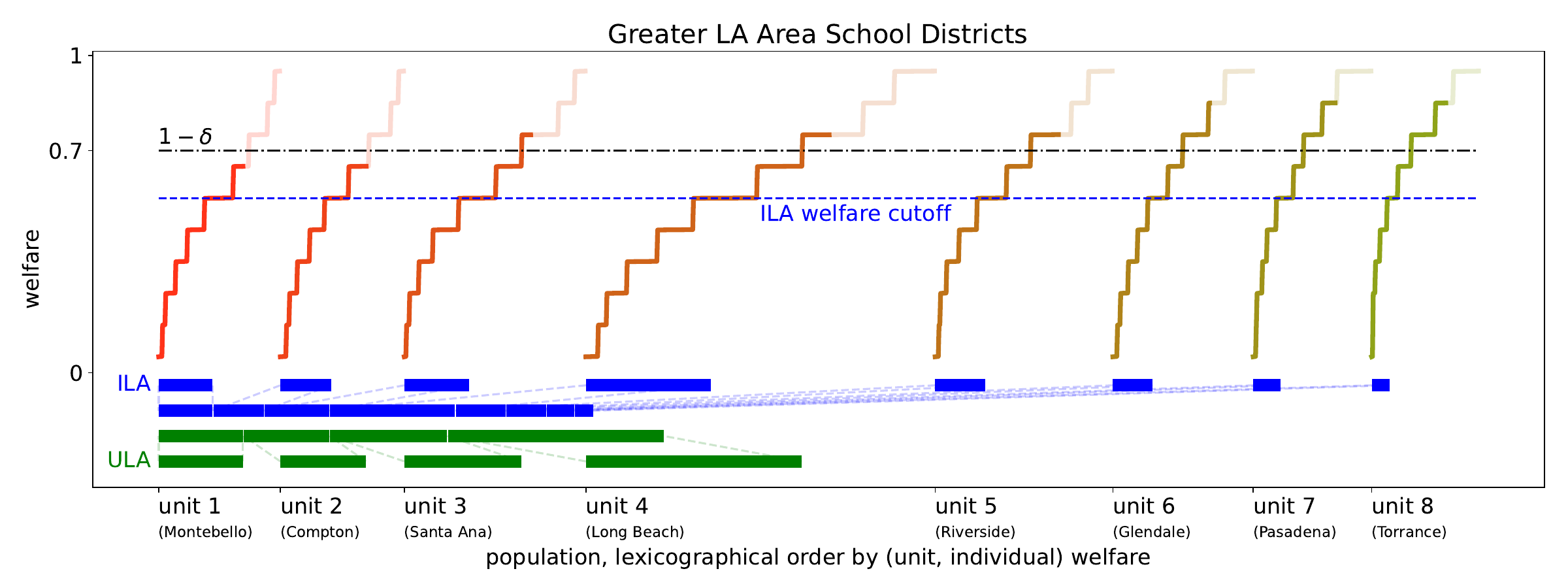}
    \caption{ULA outperforms ILA in a real-world high inequality setting. Eight school districts from the greater Los Angeles (LA) area are considered. Household income, in $10$~brackets, is used as a proxy for individual welfare. A within-unit allocation with $q=0.3$ and $q'\approx0$ is considered, where individuals among the top~$q$ fraction at each unit appear faded. The values obtained by ILA and ULA from each unit are depicted with horizontal bars, and their overall values are compared.}
    \label{fig:acs_ed_demo_LA}
\end{figure}

\ifnotarxiv

\begin{figure}[h]
    \centering
    \includegraphics[width=0.9\linewidth]{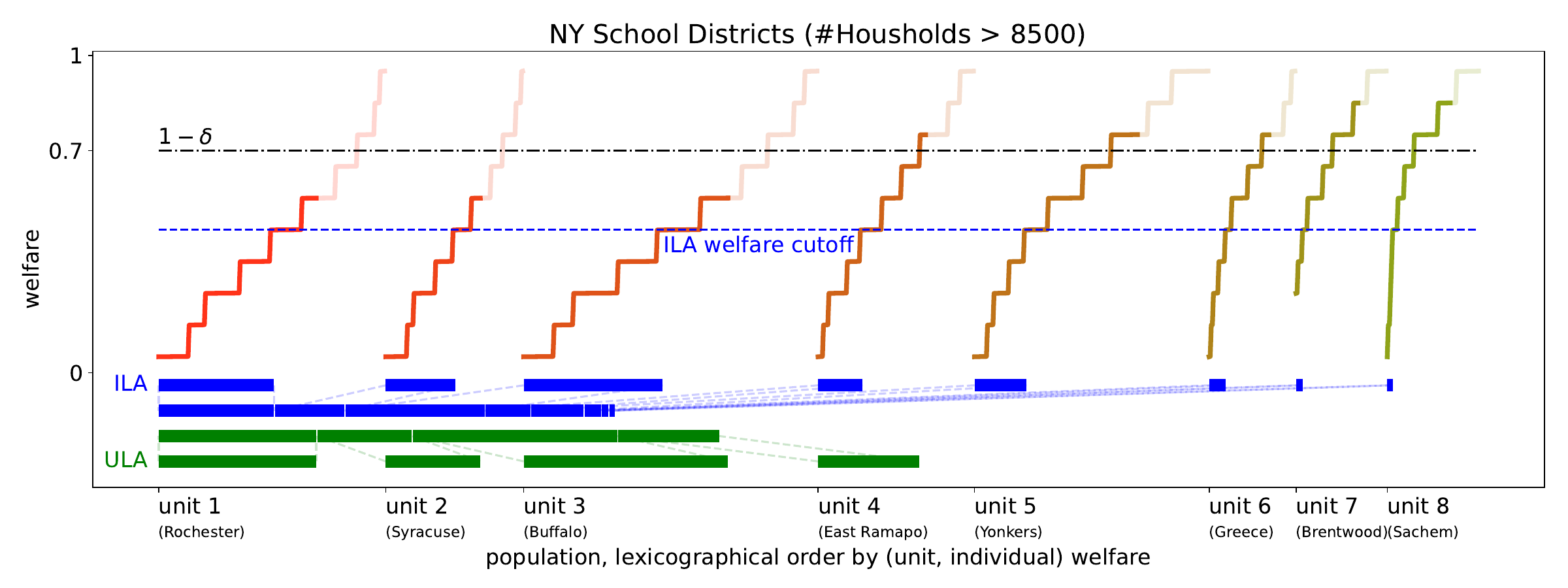}
    \caption{Similar to \cref{fig:acs_ed_demo_LA}, ULA outperforms ILA in a real-world high inequality setting. New York State's school districts with a household population of $8,500$ or more are considered (excluding the New York City Department of Education due to its sheer size).}
    \label{fig:demo_NY}
\end{figure}
\begin{figure}[h]
    \centering
    \includegraphics[width=0.9\linewidth]{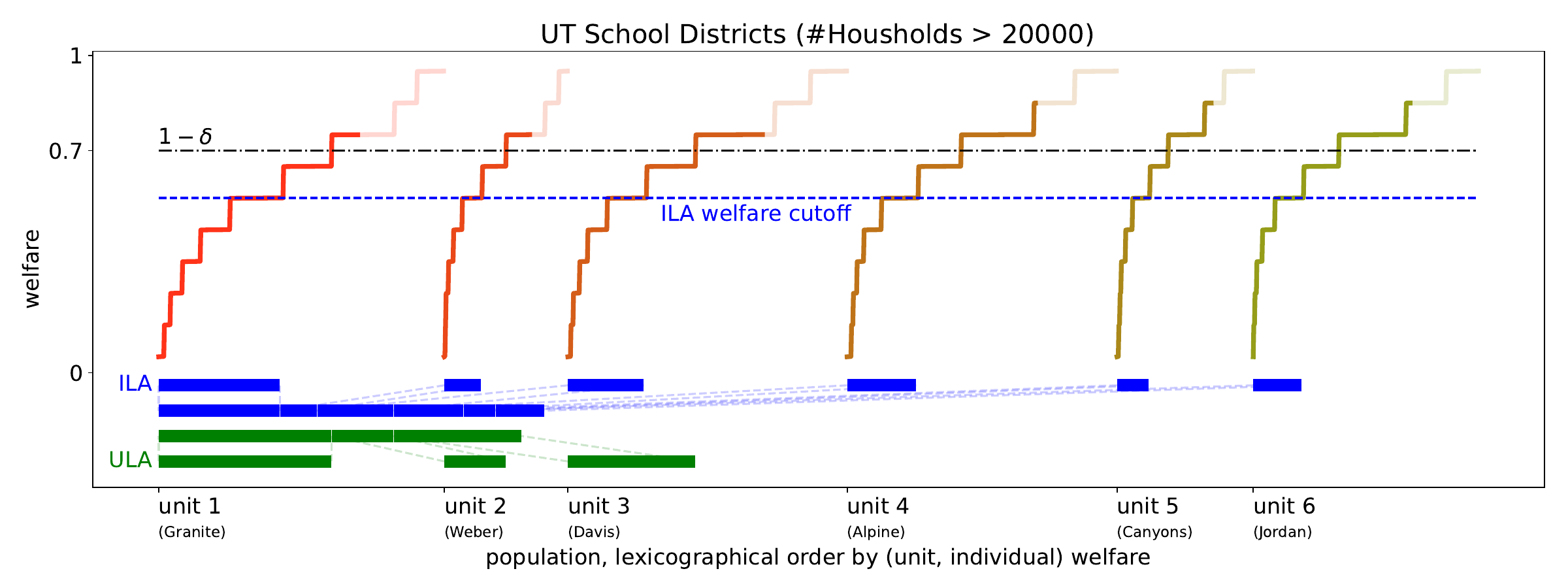}
    \caption{Unlike \cref{fig:acs_ed_demo_LA}, ILA outperforms ULA in a real-world low inequality setting. Utah's school districts with a household population of $20,000$ or more are considered.}
    \label{fig:demo_UT}
\end{figure}

\fi

\subsection{Theoretical Illustration}
\fi

We now present our theoretical argument in an illustrative special case as a warm-up exercise. This simple argument shows why ULA dominates ILA when inequality is high, or the budget is low. 
In the following, we consider cases where the cost of treating everyone~$C \coloneqq M N c$ is too high so that the budget~$B$ scales sublinearly or linearly with~$C$.

\ifarxiv
Recall that we associate each unit~$k$ with the fraction of its members with a welfare above $1-\delta$, denoted by~$\rho_k$. In calculating ULA's value, we lower bound the value from unit~$k$ by $(1 - \rho_k) \, \delta$. In other words, we neglect the value realized from high-welfare individuals. This approach obviates the need to talk about individual welfares, and the unit profile~$\vrho$ characterization will be sufficient to present a lower bound on ULA's value.
\fi

Consider $M$~units uniformly distributed around a mean value~$\bar{\rho}$: $\rho_1 = \bar{\rho} - g \, \bar{\rho}$ and $\rho_M = \bar{\rho} + g \, \bar{\rho}$, for a constant $g \in [0, 1]$. In this case,
\begin{equation}
\label{eq:rho_k_uniform}
    \rho_k = \bar{\rho} + 2 g \, \bar{\rho} \, \big(k - (M + 1)/2\big) / (M - 1)\,,\quad k \in [M]
    \,.
\end{equation}
Here, the parameter~$g$ controls how spread out the units are around~$\bar{\rho}$. We can interpret~$g$ as a measure of inequality across units as well. In particular, $g$ is proportional to the Gini index of~$\rho_k$s denoted by~$G_\rho$. By definition,
\begin{equation*}
    G_\rho \coloneqq \frac{\sum_{k,k'} |\rho_k - \rho_{k'}|}{2 M^2 \bar{\rho}} =
    2g\cdot \frac{\sum_{k' < k} (k - k')}{(M - 1) M^2}\,
    .
\end{equation*}
Then a direct calculation shows $G_\rho = g \, (1 + 1/M)/3$. 

Consider a within-unit allocation with $q > 0$ and $q' \approx 0$. If $\rho_k \le q$, ULA avoids all individuals in unit~$k$ that have welfare above $1-\delta$. Hence, every individual treated under ULA in unit~$k$ will experience the full~$\delta$ improvement. For uniformly distributed units, a large~$q$ such as $q \ge \bar{\rho}$ ensures this happens for at least half of the units. So, if the budget is not excessively large, ULA will be optimal. In particular, so long as predictions are not free, ULA dominates ILA.

\ifarxiv

Even when $q < \bar{\rho}$, either a low budget or high inequality ensures the dominance of ULA (\cref{fig:illustration}). To see the former, observe that a minimal inequality of $g > 1 - q/\bar{\rho}$ guarantees that at least for the first unit, $\rho_1 < q$. Similarly, for a sublinear budget~$B = o(C)$, a minimal inequality of $g \ge 1 - q/\bar{\rho} + o(1)$ ensures that all treated units have $\rho_k \le q$, implying the optimality of ULA (\cref{fig:illustration_low_B}). Even for a linear budget~$B = \Theta(C)$, a simple argument shows that with a sufficiently high inequality, say, $g \ge 2(1 - q/\bar{\rho})$, as long as the budget is not excessively large, such as not surpassing the amount for treating one-fourth of the units, the optimality of ULA is ensured (\cref{fig:illustration_high_g}).

\begin{figure}[h]
    \centering
    \subfigure[High inequality ensures for every treated unit, $\rho_k \le q$ even when many units are treated, thus ULA remains efficient.]{\includegraphics{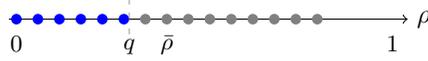}\label{fig:illustration_high_g}}
    \\
    \subfigure[Even when inequality is not high, so long as the budget is not large, ULA remains efficient.]{\includegraphics{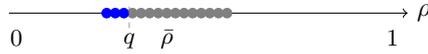}\label{fig:illustration_low_B}}
    \caption{In the special case of uniformly distributed units, each unit is represented by a dot. The first $K$ units (the targeted units) are indicated in blue.}
    \label{fig:illustration}
\end{figure}

\fi

Inequality plays a significant role in ULA, including cases where the within-unit allocation treats everyone ($q=q'=0$). When everyone is treated at unit~$k$, at least a $(1 - \rho_k)$~fraction of individuals experience a $\delta$~improvement. So, for $K$~treated units, $V_\text{unit} \ge \sum_{k \in [K]} N \delta (1 - \rho_k)$. Plugging $\rho_k$ from \cref{eq:rho_k_uniform} into this, a direct calculation yields
\begin{equation*}
    V_\text{unit} \ge K N \delta \left(1 - \bar{\rho} + g \, \bar{\rho} \, \frac{M - K}{M - 1} \right)
    \,.
\end{equation*}
Then, using $K \approx \frac{B}{N c}$, we obtain
\begin{align}
    \label{eq:value_decomposition_uniform}
    V_\text{unit} &\ge B \frac{\delta}{c} \Big(\underbrace{\,1 - \bar{\rho}\,}_{\substack{\text{value from} \\ \text{random targeting}}} +\quad \underbrace{\,g \, \bar{\rho} \, ( 1 - B/C )\,}_{\substack{\text{extra value} \\ \text{from inequality}}} \Big)
    \,.
\end{align}
\cref{eq:value_decomposition_uniform} decomposes the value of ULA: For a dollar spent, the maximum realizable value is $\delta/c$. If units were chosen randomly in ULA, on average, a unit of $\rho_k = \bar{\rho}$ had been treated, which in the worst case could decrease the rate budget is turning into value by a factor of $(1 - \bar{\rho})$. However, in the presence of inequality, the selected units will be those with smaller $\rho_k$s, and ULA performs better than random. This extra value is proportional to $g$ and $(1 - B/C)$.  

In contrast, consider ILA, where we pay a price of~$p$ for (say, perfect) prediction. Paying this price, ILA loses at least a value of $p\frac{\delta}{c}$ compared to the optimum. Therefore,
\begin{equation*}
    V_\text{ind} \le B\frac{\delta}{c} \big( 1 - p/B \big)
    \,.
\end{equation*}
Comparing this with the lower bound on ULA's value in \cref{eq:value_decomposition_uniform}, 
we have $\text{ULA} \succeq \text{ILA}$ so long as the fractional price of prediction satisfies
\begin{equation*}
    p/B \ge \bar{\rho} \, \big(1 - g \, (1 - B/C)\big)
    \,.
\end{equation*}
This requirement becomes weaker as inequality grows, and as budget shrinks.

Going beyond uniformly distributed units by considering arbitrarily distributed units, we will show in \cref{thm:dominant_ula_suff_conditions} that a similar observation of $\text{ULA} \succeq \text{ILA}$ holds in general if either inequality is high or the budget is not too large in the presence of a medium inequality. We give a qualitative view of the results in \cref{tab:dominant_ula_qualitatively}.

\begin{table*}[t]
    \centering
    \caption{Sufficient conditions for ULA to dominate ILA. Refer to \cref{thm:dominant_ula_suff_conditions} for accurate statements. For a similar table with quantitative values for low, medium, and high, refer to \cref{tab:dominant_ula_quantitative}.\smallskip}
    \begin{tabular}{lcccc}
        & \multirow{2}{*}{High inequality} & \multicolumn{2}{c}{Medium inequality} & \multirow{2}{*}{Low inequality} \\
        \cline{3-4}
        &                                  & Medium or lower budget & Large budget & \\
        \midrule
        High price of prediction & $\text{ULA} \succ \text{ILA}$ & $\text{ULA} \succ \text{ILA}$ & $\text{ULA} \succ \text{ILA}$ & $\text{ULA} \succ \text{ILA}$ \\
        \cline{4-5}
        Medium price of prediction & $\text{ULA} \succ \text{ILA}$ & $\text{ULA} \succ \text{ILA}$ & \multicolumn{2}{|c|}{\multirow{2}{*}{?}} \\
        Low price of prediction & \multicolumn{2}{c}{$\big(1+O(q')\big)\!\cdot\!\text{ULA} \succeq \text{ILA}$} & \multicolumn{2}{|c|}{} \\
        \cline{4-5}
    \end{tabular}
    \label{tab:dominant_ula_qualitatively}
\end{table*}
\section{Sufficient Conditions for Dominant Unit-Level Allocation}
\label{sec:dominant_ula}

We present sufficient conditions under which ULA dominates ILA. If we restrict feasible unit profiles to those meeting these conditions, $\text{ULA} \succeq \text{ILA}$.
To this end, we set upper bounds on~$V_\text{ind}$ and lower bounds on~$V_\text{unit}$ and then compare the bounds. 

\ifarxiv
\subsection{Upper bounding the Value of ILA}
\fi
Given a predictor with $\epsilon$~error that costs~$p(\epsilon)$ to be implemented, there remains a budget of $B - p(\epsilon)$ for intervention. If the predictor is sufficiently accurate, only individuals with welfare less than $(1-\delta)$ will be targeted, and the remaining budget will be used optimally. Therefore, optimistically,
\begin{equation}
\label{eq:V_ind_ub}
    V_\text{ind} \le \max \Big\{ \frac{B - p(\epsilon)}{c},\; 0 \Big\} \, \delta
    \,.
\end{equation}
This bound is tight if there are sufficient individuals with low welfare. Specifically, if there are $I = \lfloor \frac{B - p(\epsilon)}{c} \rfloor$ individuals with $w_i \le 1 - \delta - 2\epsilon$, the bound is tight (\cref{prop:ila_bound_is_tight}).

\ifarxiv
\subsection{Lower bounding the Value of ULA}
\fi
To set a lower bound on $V_\text{unit}$ (\cref{eq:def_V_unit}), we first lower bound $T_k$ in terms of the within-unit allocation parameters~$q$ and~$q'$ and the unit~$k$'s fraction of high welfare individuals~$\rho_k$.

\begin{theoremEnd}{lemma}[Treatment effect to loss conversion]
\label{lem:loss_definition}
For a $(q,q')$-within-unit allocation, we have
\begin{equation*}
    T_k/\delta \ge 1 - q - {\rm loss}(\rho_k)
    \,,
\end{equation*}
where we define
\begin{equation}
\label{eq:loss_definition}
    {\rm loss}(\rho_k) \coloneqq \begin{cases}
        \min \, \{\rho_k, q'\} \,, & \rho_k \le q\,, \\
        \rho_k - (q - q') \,, & \rho_k > q\,.
    \end{cases}
\end{equation}
\end{theoremEnd}
\begin{proofEnd}
    Recall that within an intervened unit, resources will be allocated to the lower $(1-q)$~fraction of individuals, potentially including a $q'$~fraction from the top~$q$ cut. Depending on the relative size of~$q$ to~$\rho_k$, two possibilities arise:
    \begin{itemize}
        \item $\rho_k \le q$: In the worst case, the top $\rho_k$~portion may erroneously receive the resources in place of the population in need. However, if $q' \le \rho_k$, only a proportion of size~$q'$ may be misclassified. The treatment on these individuals may have zero effect. Hence, $T_k/\delta \ge 1 - q - \min \, \{\rho_k, q'\}$.
    
        \item $\rho_k > q$: Since at least $(q-q')$~fraction of top individuals rightfully will not receive the resources, at most $\rho_k - (q - q')$~fraction of top individuals will be among treated. These individuals might experience a zero effect. Therefore, we have $T_k/\delta \ge 1- q - \big(\rho_k - (q - q')\big)$. 
    \end{itemize}
    In both cases, we can write the lower bound as $T_k/\delta \ge 1 - q - \text{loss}(\rho_k)$, where $\text{loss}$ is defined in \cref{eq:loss_definition}.
\end{proofEnd}
\ifarxiv
The proof is straightforward using the observation that if $\rho_k \le q$, within-unit allocation largely avoids top~$\rho_k$ individuals with $\tau_i < \delta$ by excluding the top~$q$ fraction. In the case of $\rho_k > q$, it at least avoids $(q-q')$ of high-welfare individuals with potentially small $\tau_i$s.

\fi 
\cref{lem:loss_definition} implies that instead of lower bounding $V_\text{unit}$, we may upper bound the total loss
\begin{equation*}
    \text{loss}_K(\vrho) \coloneqq \textstyle\sum_{k \in [K]}\text{loss}(\rho_{s(k)})
\end{equation*}
for any valid sorting~$s(\cdot)$ of the $\rho_k$s. The next lemma provides such a bound on $\text{loss}_K$.

\begin{theoremEnd}{lemma}
\label{lem:ub_loss_unit}
For a unit profile~$\vrho$, denote the mean and the Gini index of $\rho_k$s by $\bar{\rho}$ and $G_\rho$, respectively.
Assume the budget is not excessively large, so the number~$K$ of units treated by ULA does not exceed $M(1 - \bar{\rho})$. 
We have
\begin{equation*}
    {\rm loss}_K(\vrho) \le \begin{cases}
        K q', & q > q_c / u(K/M)\,, \\
        K (q_c -q + q') \,, & q \le q_c \, u(K/M)\,.
    \end{cases}
\end{equation*}
Here, $q_c$ is a critical value for~$q$ defined as
\begin{equation}
\label{eq:q_c}
    q_c \coloneqq \bar{\rho} \, \frac{1 - \bar{\rho} - G_\rho}{1 - \bar{\rho} - \bar{\rho} \, G_\rho}
    \,,
\end{equation}
and $u(K/M) \coloneqq 1 - 2(K/M)\frac{1-\bar{\rho}}{1 - \bar{\rho} - \bar{\rho}G_\rho} = 1 - O(K/M)$.
\end{theoremEnd}
\begin{proofEnd}
    Given a unit profile~$\vrho$, let $s(\cdot)$ be the ascending sorting of~$\rho_k$s. For notational simplicity, assume $s(k) = k$ without loss of generality. The largest possible loss of any unit profile~$\vrho$ with a mean of~$\bar{\rho}$ and Gini index of~$G_\rho$ is the solution of 
    the optimization problem
    \begin{align}
    \label{eq:linear_prog}
        \max_\vrho\quad& \sum_{k \in [K]} \text{loss}(\rho_k) \\
        \text{s.t.}\quad&\;\;\sum_{k \in [M]} \rho_{k} = M \bar{\rho}\,, &&\text{(Mean constraint)} \nonumber\\
        \label{eq:gini_index_constraint_eq}
        &\;\;\sum_{k \in [M]} k \rho_{k} - \frac{M(M + 1)}{2} \bar{\rho} - \frac{M^2}{2} \bar{\rho} G_\rho = 0\,, &&\text{(Gini index constraint)} \\
        &\;\;1 \ge \rho_{k} \ge 0\,, &&\text{(Valid range constraint)} \nonumber\\
        &\; \rho_{k+1} \ge \rho_{k}, \; \text{for } k\in[M-1]\,. &&\text{(Valid sorting constraint)} \nonumber
    \end{align}
    The second constraint (\cref{eq:gini_index_constraint_eq}) is obtained by rearranging terms in the Gini index formula. Since $\text{loss}(\cdot)$ is a piecewise linear function and the constraints are all linear functions of~$\vrho$, the above optimization problem is a linear program. We upper bound the solution to this problem in the following. 
    
    As the first step, we relax the Gini index constraint (\cref{eq:gini_index_constraint_eq}) to include every unit profile with a Gini index of~$G_\rho$ or more:
    \begin{equation}
    \label{eq:gini_index_constraint}
        \sum_{k \in [M]} k \rho_{k} - \frac{M(M + 1)}{2} \bar{\rho} - \frac{M^2}{2} \bar{\rho} G_\rho \ge 0
        \,.
    \end{equation}
    The solution of the linear program with this relaxed constraint will serve as an upper bound for~$\text{loss}_K(\vrho)$.
    
    Depending on the magnitude of $\rho_K$ (corresponding to the last treated unit) compared to $q$ and $q'$, three regimes can be defined:
    \begin{enumerate}
        \item $\rho_K < q'$
        \item $q' \le \rho_K < q$
        \item $q \le \rho_K$
    \end{enumerate}
    For a fixed $K$, we upper bound the solution of the linear program in each regime and take the maximum of these three upper bounds as a bound for the original problem.
    
    \paragraph{Second Regime.}
    Let $\vrho^*$ be an optimal solution in the second regime where $q' \le \rho^*_K < q$. We show a new solution~$\vrho$ can be constructed from~$\vrho^*$ with a similar mean, larger or equal Gini index, and larger or equal loss. Hence, this solution should also be optimal. Let $l = \min \, \{k: \rho^*_k > q'\}$ and $u = \max \, \{k: \rho^*_k < 1\}$. Define $\Delta = \min \, \{\rho^*_l - q', 1 - \rho^*_u\}$. This is the maximum amount that we can move $\rho^*_l$ back and $\rho^*_u$ forward while keeping them in the $[q', 1]$ range. Define the new values $\rho_l = \rho^*_l - \Delta$, $\rho_u = \rho^*_u + \Delta$, and $\rho_k = \rho^*_k$ for $k \notin \{l,u\}$. The new solution~$\vrho$ admits the same mean as $\vrho^*$, preserves the ordering, and increases the margin of the Gini index constraint (\cref{eq:gini_index_constraint}) by $(u - l)\Delta \ge 0$. Further, since $\rho^*_K < q$, we already know for $l \le k \le K$, $\text{loss}(\rho_k)$ is the minimum loss in range~$[q', 1]$ and this change can only increase $\text{loss}(\vrho)$. Therefore, $\vrho$ should also be an optimal solution. By repetitively applying this operation, we can obtain an optimal solution~$\vrho$ such that no more than one unit lies in $(q', 1)$. During this process, $\rho_K$ may enter the third regime, which is fine, as it shows the loss of the solution in the third regime is larger. In the rest of the proof, we continue in the second regime where $\rho_K < q$.
    
    Let us call the constructed unit profile, which we know is optimal, $\vrho^*$. We build on the same idea and further manipulate~$\vrho^*$ to obtain another optimal unit profile~$\vrho$. Let $l = \min \, \{k: \rho^*_k > 0\}$ and $u = \max \, \{k: \rho^*_k < q'\}$. Define $\Delta = \min \, \{\rho^*_l, q' - \rho^*_u\}$. This is the maximum amount that we can move $\rho^*_l$ back and $\rho^*_u$ forward while keeping them in the $[0, q']$ range. Define the new values $\rho_l = \rho^*_l - \Delta$, $\rho_u = \rho^*_u + \Delta$, and $\rho_k = \rho^*_k$ for $k \notin \{l,u\}$. The new solution~$\vrho$ again has the same mean as $\vrho^*$, preserves the ordering, and increases the margin of the Gini index constraint by $(u - l)\Delta \ge 0$. Moreover, since $\text{loss}(\cdot)$ is linear in the $[0, q']$ range and $u \le K$, this operation does not change the objective function. Therefore, if $\vrho^*$ was optimal, $\vrho$ also remains optimal. By repetitively applying this operation, we can obtain an optimal solution~$\vrho$ such that no more than one unit admits $0 < \rho_k < q'$.
    
    Neglecting a maximum of two units, in the optimal solution~$\vrho$ constructed so far, $\rho_k$ only takes a value from $\{0, q', 1\}$. Hence, we can represent this solution by three numbers $K_0$, $K_1$, and $K_2$ which are the number of units with $\rho_k$ of $0$, $q'$, and $1$, respectively. These numbers should be integers; however, neglecting the value of three units in maximum, we can assume they are nonnegative real numbers satisfying the sum constraint $K_0 + K_1 + K_2 = M$. The mean constraint for this solution is 
    \begin{equation*}
        K_1 q' + K_2 = M \bar{\rho}
        \,.
    \end{equation*}
    Solving for~$K_2$, we have $K_2 = M \bar{\rho} - K_1 q'$. The nonnegativity of~$K_2$ requires $K_1 \le M \bar{\rho} / q'$. Plugging~$K_2$ into the sum constraint and solving for~$K_0$, we obtain $K_0 = M (1 - \bar{\rho}) - K_1 (1 - q')$. The nonnegativity of~$K_0$ requires $K_1 \le M (1 - \bar{\rho})/(1 - q')$. The Gini index constraint for~$\vrho$ is
    \begin{equation*}
        \sum_{k \in [M]} k \rho_k = \frac{1}{2} (K_0 + 1 + K_0 + K_1) K_1 q' + \frac{1}{2}(M - K_2 + 1 + M) K_2 \ge \frac{1}{2} M (M + 1) \bar{\rho} + \frac{1}{2} M^2 \bar{\rho} G_\rho
        \,.
    \end{equation*}
    Plugging $K_0$ and $K_2$ into this and simplifying equations, we obtain
    \begin{equation*}
        \frac{K_1}{M} \le \sqrt{\frac{\bar{\rho}(1 - \bar{\rho} - G_\rho)}{q'(1 - q')}}
        \,.
    \end{equation*}
    Note that for values bounded between $0$ and $1$ with a mean of~$\bar{\rho}$, the Gini index cannot exceed~$(1 - \bar{\rho})$. Putting these together,
    \begin{equation*}
        0 \le \frac{K_1}{M} \le \min \, \Big\{ \frac{\bar{\rho}}{q'}, \frac{1 - \bar{\rho}}{1 - q'}, \sqrt{\frac{\bar{\rho}(1 - \bar{\rho} - G_\rho)}{q'(1 - q')}} \Big\}
        \,.
    \end{equation*}
    This imposes the following bounds on~$K_0$:
    \begin{equation}
    \label{eq:_proof_regime_2_K_0_lb}
        1 - \bar{\rho} \ge \frac{K_0}{M} \ge \max \, \Big\{ 0, 1 - \bar{\rho} - \bar{\rho}\frac{1-q'}{q'}, 1 - \bar{\rho} - \sqrt{\frac{1 - q'}{q'}\bar{\rho}(1 - \bar{\rho} - G_\rho)} \Big\}
        \,.
    \end{equation}
    In regime~$2$, we have $q' \le \rho_K < q$. This requires $K_0 \le K \le K_0 + K_1$. Substituting~$K_1$ in terms of~$K_0$ and performing a direct calculation, we obtain the following upper bounds on $K_0$:
    \begin{equation}
    \label{eq:_proof_regime_2_K_0_ub}
        \frac{K_0}{M} \le \frac{K}{M}, \;\;\;\;\; \frac{K_0}{M} \le \frac{1 - \bar{\rho}}{q'} - \frac{K}{M} \frac{1-q'}{q'}
        \,.
    \end{equation}
    For a solution to exist in regime~$2$, the upper bounds of \cref{eq:_proof_regime_2_K_0_ub} should be larger than the lower bound of \cref{eq:_proof_regime_2_K_0_lb}. One can verify that for $K \le 1 - \bar{\rho}$, the upper bound in the right-hand side of \cref{eq:_proof_regime_2_K_0_ub} is larger than the lower bound. Hence, there remains one condition for the solution of regime~$2$ to exist:
    \begin{equation*}
        \frac{K}{M} \ge 1 - \bar{\rho} - \min \, \Big\{\bar{\rho}\frac{1-q'}{q'}, \sqrt{\frac{1 - q'}{q'}\bar{\rho}(1 - \bar{\rho} - G_\rho)} \Big\}
        \,.
    \end{equation*}
    In regime~$2$, we can write the objective as
    \begin{equation*}
        \text{loss}_K(\vrho) = (K - K_0) q'
        \,.
    \end{equation*}
    Maximizing the objective then corresponds to minimizing~$K_0$ subject to \cref{eq:_proof_regime_2_K_0_lb}. It can be verified that when $q' < q_c = \bar{\rho} \frac{1 - \bar{\rho} - G_\rho}{1 - \bar{\rho} - \bar{\rho} G_\rho}$, the lower bound in \cref{eq:_proof_regime_2_K_0_lb} is zero, resulting in $\text{loss}_K(\vrho) = K q'$. When $q' \ge q_c$, the lower bound is at least $K_0 \ge 1- \bar{\rho} - \sqrt{\bar{\rho} \, (1 - \bar{\rho} - G_\rho) \, (1 - q')/q'}$. For $K_0$ equal to this lower bound, $\text{loss}_K(\vrho) \le (K - K_0) q'$. 
    
    \paragraph{Third Regime.}
    Let $\vrho^*$ be an optimal solution in the third regime where $q \le \rho^*_K$. We follow a similar approach and show that a new optimal solution~$\vrho$ can be constructed from~$\vrho^*$ that admits a specific structure. First of all, Let us upper bound $\text{loss}(\rho_k)$ with $\widetilde{\text{loss}}(\rho_k) = \max\,\{q', \rho_k - (q-q')\}$. We will consider maximizing $\widetilde{\text{loss}}_K(\vrho) \coloneqq \sum_{k \in [K]} \widetilde{\text{loss}}(\rho_k)$ instead of $\text{loss}_K(\vrho)$ in the linear program of \cref{eq:linear_prog}. Fix $\rho^*_K$. Let $l = \min \, \{k: k > 0\}$ and $u = \max \, \{k: \rho^*_k < \rho^*_K\}$. Define $\Delta = \min \, \{\rho^*_l, \rho^*_K - \rho^*_u\}$. This is the maximum amount we can move~$\rho^*_l$ back and $\rho^*_u$ forward while keeping them in the $[0, \rho^*_K]$ range. Define the new values $\rho_l = \rho^*_l - \Delta$, $\rho_u = \rho^*_u + \Delta$, and $\rho_k = \rho^*_k$ for $k \notin \{l,u\}$. The new solution~$\vrho$ admits the same mean as~$\vrho^*$, preserves the ordering, and increases the margin of the Gini index constraint by $(u - l)\Delta \ge 0$. Further, since $\widetilde{\text{loss}}(\cdot)$ is convex in~$[0, \rho^*_K]$, this change can only increase the loss. Therefore, $\vrho$ should be optimal as well. By repetitively applying this operation, we can obtain an optimal solution~$\vrho$ such that no more than one unit falls in $0 < \rho_k < \rho^*_K=\rho_K$.
    
    Let us call the constructed unit profile $\vrho^*$. Again, fix $\rho^*_K$. Let $l = \min \, \{k: k > \rho^*_K\}$ and $u = \max \, \{k: \rho^*_k < 1\}$. Define $\Delta = \min \, \{\rho^*_l - \rho^*_K, 1 - \rho^*_u\}$. This is the maximum amount we can move~$\rho^*_l$ back and $\rho^*_u$ forward while keeping them in the $[\rho^*_K, 1]$ range. Define the new values $\rho_l = \rho^*_l - \Delta$, $\rho_u = \rho^*_u + \Delta$, and $\rho_k = \rho^*_k$ for $k \notin \{l,u\}$. This new solution has the same mean as~$\vrho^*$ with a higher Gini index. It also leaves the objective function unaffected. Hence, the constructed solution is also optimal. By repetitively applying this operation, we can obtain an optimal solution~$\vrho$ such that for no more than one unit $\rho^*_K < \rho_k < 1$.
    
    For notational simplicity, denote $\rho^*_K$ by~$\rho$, a free parameter in the constructed solution. Neglecting a maximum of two units, $\rho_k$ in the constructed optimal solution can only take a value from~$\{0, \rho, 1\}$. So, we can represent this solution with~$\rho$ and three integers $K_0$, $K_1$, and $K_2$ which are the number of units with $\rho_k$ equal to $0$, $\rho$, and $1$, respectively. Although these numbers are integers, we can treat them as nonnegative real numbers, neglecting the effect of three units in maximum. We require $K_0 + K_1 + K_2 = M$. The mean constraint also requires
    \begin{equation*}
        K_1 \rho + K_2 = M \bar{\rho}
        \,.
    \end{equation*}
    Solving these two equations for~$K_1$ and $K_2$ in terms of~$K_0$, we obtain $K_1 = \big(M(1 - \bar{\rho}) - K_0\big)/(1-\rho)$ and $K_2 = \big(M(\bar{\rho} - \rho) + K_0 \rho \big)/(1 - \rho)$. The nonnegativity of $K_1$~and~$K_2$ imposes a bound on $K_0$: $1 - \bar{\rho}/\rho \le K_0/M \le 1 - \bar{\rho}$. The Gini index constraint requires
    \begin{equation*}
        \sum_{k \in [M]} k \rho_k = \frac{1}{2}(K_0+1+K_0+K_1)K_1 \rho + \frac{1}{2}(M - K_2 + 1 + M)K_2 \ge \frac{1}{2} M (M+1)\bar{\rho} + \frac{1}{2} M^2\bar{\rho} G_\rho
        \,.
    \end{equation*}
    Plugging $K_1$ and $K_2$ as a function of~$K_0$ into this constraint and simplifying equations, we obtain
    \begin{equation*}
        \frac{K_0}{M} \ge 1 - \bar{\rho} - \sqrt{\frac{1-\rho}{\rho}} \sqrt{\bar{\rho}(1 - \bar{\rho} - G_\rho)}
        \,.
    \end{equation*}
    Since we are in the third regime, $K_0 \le K$, and $K \le K_0 + K_1$. The latter imposes another upper bound on~$K_0$: $K_0 \le M(1-\bar{\rho})/\rho - K(1-\rho)/\rho$. Altogether, the following bounds are in place on~$K_0$:
    \begin{equation}
    \label{eq:_proof_third_regime_K_0_bounds}
        \min \, \Big\{\frac{K}{M}, 1-\bar{\rho}, \frac{1 - \bar{\rho}}{\rho} - \frac{K}{M}\frac{1-\rho}{\rho}\Big\} \ge \frac{K_0}{M} \ge \max \, \Big\{0, 1 - \frac{\bar{\rho}}{\rho}, 1 - \bar{\rho} - \sqrt{\frac{1-\rho}{\rho}} \sqrt{\bar{\rho}(1 - \bar{\rho} - G_\rho)} \Big\}
        \,.
    \end{equation}
    One can verify that for $K \le 1 - \bar{\rho}$, the second and third term of the upper bound in the left-hand side of \cref{eq:_proof_third_regime_K_0_bounds} are always larger than or equal to the lower bound. Hence, for the solution of the third region to exist, it is required $K/M \ge \max\,\{1 - \bar{\rho}/\rho, 1 - \bar{\rho} - \sqrt{\bar{\rho}(1 - \bar{\rho} - G_\rho)(1-\rho)/\rho}\}$. After a direct calculation, this imposes two upper bounds on~$\rho$:
    \begin{equation}
    \label{eq:_proof_regime_3_rho_ub}
        \rho \le \frac{\bar{\rho}}{1 - \frac{K}{M}}\,, \;\;\;\;\; \rho \le \bar{\rho}\frac{1 - \bar{\rho} - G_\rho}{1 - \bar{\rho} - \bar{\rho}G_\rho - 2\frac{K}{M}(1 - \bar{\rho} - \frac{K}{2M})}
        \,.
    \end{equation}
    Define
    \begin{equation}
    \label{eq:q_u_def}
        q_u(K/M) \coloneqq \min \, \Big\{ 
        \frac{\bar{\rho}}{1 - \frac{K}{M}},
        \bar{\rho}\frac{1 - \bar{\rho} - G_\rho}{1 - \bar{\rho} - \bar{\rho}G_\rho - 2\frac{K}{M}(1 - \bar{\rho} - \frac{K}{2M})} 
        \Big\}
        \,.
    \end{equation}
    Then the upper bounds of \cref{eq:_proof_regime_3_rho_ub} can be summarized as $\rho \le q_u(K/M)$. Since we are in regime~$3$ and $\rho \ge q$, a necessary condition in this regime is $q \le q_u(K/M)$.
    
    The upper bounded loss for the constructed~$\vrho$ in regime~$3$ is 
    \begin{equation}
    \label{eq:_proof_regime_3_loss}
        \widetilde{\text{loss}}_K(\vrho) = K_0 q' + (K-K_0)(\rho - q + q') = (K - K_0)(\rho  - q) + K q'
        \,.
    \end{equation}
    For a fixed~$\rho$, to maximize $\widetilde{\text{loss}}_K(\vrho)$, we have to minimize~$K_0$. From \cref{eq:_proof_third_regime_K_0_bounds} we can see
    \begin{equation}
    \label{eq:_proof_regime_3_K_0_final_lb}
        \frac{K_0}{M} \ge \begin{cases}
            1 - \bar{\rho} - \sqrt{\frac{1-\rho}{\rho}} \sqrt{\bar{\rho}(1 - \bar{\rho} - G_\rho)}\,, & \rho > q_c\,, \\
            0\,, & \text{o.w.}\,,
        \end{cases}
    \end{equation}
    where
    \begin{equation*}
        q_c \coloneqq \bar{\rho} \frac{1 - \bar{\rho} - G_\rho}{1 - \bar{\rho} - \bar{\rho} G_\rho}
        \,.
    \end{equation*}
    Note the tradeoff in the choice of $\rho$: \cref{eq:_proof_regime_3_loss} indicates that a larger~$\rho$ leads to a larger loss. However, \cref{eq:_proof_regime_3_K_0_final_lb} shows that a larger~$\rho$ also results in a larger~$K_0$, reducing the loss. Consider two possibilities:
    \begin{itemize}
        \item $\rho < q_c$: Since $\rho \ge q$ in regime~$3$, we should have $q < q_c$. The lower bound of \cref{eq:_proof_regime_3_K_0_final_lb} is not binding in this case, and $K_0$ can be as low as~$0$. Since $\rho \le q_c$, a direct result of \cref{eq:_proof_regime_3_loss} is
        \begin{equation*}
            \widetilde{\text{loss}}_K(\vrho) < K(q_c - q + q')
            \,.
        \end{equation*}

        \item $\rho \ge q_c$: If $B=o(C)$, we have $K/M = o(1)$ and neglecting $o(1)$, the second upper bound of \cref{eq:_proof_regime_3_rho_ub} reduces to $q_c$. Plugging $K_0=0$ and $\rho=q_c + o(\bar{\rho})$ into \cref{eq:_proof_regime_3_loss}, we obtain 
        \begin{equation*}
            \widetilde{\text{loss}}_K(\vrho) \le K(q_c - q + q')
            \,.
        \end{equation*}
        If $B=\Theta(C)$ and consequently $K/M=\Theta(1)$, we show if $q \le q_c/2$, the maximum loss occurs at $\rho=q_c$. To show this, substitute $K_0$ with its lower bound of \cref{eq:_proof_regime_3_K_0_final_lb} in $\widetilde{\text{loss}}_K(\vrho)$:
        \begin{equation*}
            \widetilde{\text{loss}}_K(\vrho) = \Big(K -  M(1 - \bar{\rho}) + M\sqrt{\frac{1-\rho}{\rho}} \sqrt{\bar{\rho}(1 - \bar{\rho} - G_\rho)}\Big)(\rho - q) + Kq'
            \,.
        \end{equation*}
        Take a derivative of $\widetilde{\text{loss}}_K$ with respect to~$\rho$ and simplify the equations:
        \begin{equation}
        \label{eq:_proof_regime_3_dloss_drho}
            \pd{}{\rho} \widetilde{\text{loss}}_K = K - M(1-\bar{\rho}) + M \sqrt{\frac{1-\rho}{\rho}} \sqrt{\bar{\rho}(1 - \bar{\rho} - G_\rho)} \Big(1 - \frac{1}{2} \frac{1 - q/\rho}{1-\rho} \Big)
            \,.
        \end{equation}
        First of all, observe that $\sqrt{(1-\rho)/\rho}$ and the last term inside the parentheses are decreasing in~$\rho$, hence $\widetilde{\text{loss}}_K$ is concave in the feasible range of~$\rho$. Define 
        \begin{equation}
        \label{eq:q_l_def}
            q_l(K/M) \coloneqq q_c\, u(K/M)
            \,.
        \end{equation}
        We next show if $q \le q_l(K/M)$, then $\pd{\widetilde{\text{loss}}_K}{\rho}$ is nonpositive at the left boundary where $\rho = q_c$. To observe this, plug $\rho = q_c$ into \cref{eq:_proof_regime_3_dloss_drho} and simplify the equations:
        \begin{equation*}
            \pd{}{\rho} \widetilde{\text{loss}}_K |_{\rho=q_c} = K - \frac{M}{2}(1 - \frac{q}{q_c}) \frac{1 - \bar{\rho} - \bar{\rho} G_\rho}{1-\bar{\rho}}
            \,.
        \end{equation*}
        Then, one can verify that the above derivative is nonpositive for $q \le q_l(K/M)$. Combining this with our observation of concavity shows that the maximum of~$\widetilde{\text{loss}}_K$ occurs at~$\rho=q_c$. For this choice of~$\rho$, we have
        \begin{equation*}
            \widetilde{\text{loss}}_K(\vrho) < K(q_c - q + q')
            \,.
        \end{equation*}
    \end{itemize}
    
    \paragraph{First Regime.} Following a similar argument as the third regime, we can show there exists an optimal solution~$\vrho$ such that neglecting two units, every other unit has $\rho_k$ from~$\{0, \rho, 1\}$, where $\rho \le q'$. We can again represent this solution with $K_0$, $K_1$, and $K_2$ corresponding to the number of units with $\rho_k=0, \rho,$~and~$1$, respectively. One can show exactly similar upper bounds on~$\rho$ (\cref{eq:_proof_regime_3_rho_ub}) and a similar lower bound for~$K_0$ (\cref{eq:_proof_regime_3_K_0_final_lb}). In the first regime, the objective at~$\vrho$ can be written as
    \begin{equation*}
        \text{loss}_K(\vrho) = (K - K_0)\rho
        \,.
    \end{equation*}
    If $\rho < q_c$, the zero lower bound for~$K_0$ can be achieved. In this case, 
    \begin{equation*}
        \text{loss}_K(\vrho) \le K \min \{q_c, q'\}
        \,.
    \end{equation*}
    If $\rho \ge q_c$, it is necessary to have $q' \ge q_c$. In this case, we can upper-bound the objective by
    \begin{equation*}
        \text{loss}_K(\vrho) \le K q'
        \,.
    \end{equation*}

    \paragraph{Worst of All Regimes.} 
    Finally, we consider the largest objective among all regimes as an upper bound on the optimal solution of the problem. For a sublinear budget $B = o(C)$,
    \begin{itemize}
        \item If $q \ge q' \ge q_c$, the largest loss comes from regime~$1$ where $\text{loss}_K(\vrho) \le K q'$.
    
        \item If $q \ge q_c > q'$, the largest loss comes from both regimes~$1$~and~$2$ where $\text{loss}_K(\vrho) \le K q'$.
    
        \item If $q_c > q \ge q'$, the largest loss comes from regime~$3$ where neglecting $o(1)$ we have $\text{loss}_K(\vrho) \le K (q_c - q + q')$.
    \end{itemize}
    For a linear budget $B = \Theta(C)$, using the definitions of $q_l(\cdot)$ and $q_u(\cdot)$ from \cref{eq:q_l_def,eq:q_u_def} 
    \begin{itemize}
        \item If $q > q_u(K/M)$ and $q' \ge q_c$, the largest loss comes from regime~$1$ where $\text{loss}_K(\vrho) \le K q'$.
    
        \item  If $q > q_u(K/M) \ge q_c > q'$, the largest loss comes from both regimes~$1$~and~$2$ where $\text{loss}_K(\vrho) \le K q'$.
    
        \item If $q_l(K/M) \ge q \ge q'$, the largest loss comes from regime~$3$ where $\text{loss}_K(\vrho) \le K (q_c - q + q')$.
    \end{itemize}
    This completes the proof. Note that in presenting the results in the main text, we used $q_c \, u(K/M) \ge q_u(K/M)$.

\end{proofEnd}
This lemma naturally gives rise to a critical value~$q_c$ for~$q$: Neglecting $O(K/M)$ in $u(K/M)$---as is the case for a sublinear budget, for instance--- $q_c$ defines two regimes for loss depending on whether $q > q_c$ or $q \le q_c$. Due to its importance, we define:
\begin{definition}[Effective within-unit allocation]
We say a $(q, q')$-within-unit allocation is effective if $q > q_c$ and least effective otherwise. In both cases, we assume $q' \le (1 - \bar{\rho}) q$.
\end{definition}
A direct calculation shows that $q > q_c$ in a sufficiently high inequality setting where $G_\rho/(1 - \bar{\rho}) > \frac{1 - q/\bar{\rho}}{1 - q}$.
It is also straightforward to show $q_c \le \bar{\rho} \, (1 - G_\rho)$. So, regardless of the level of inequality, $q > \bar{\rho}$ is sufficient to obtain $q > q_c$.
As an
immediate result of \cref{lem:ub_loss_unit}, we can lower bound $V_\text{unit}$. For brevity, we present this only for $B=o(C)$ next.
\begin{theoremEnd}[no proof end, proof here]{corollary}
\label{cor:lb_V_unit_sublinear_budget}
Under the conditions of \cref{lem:ub_loss_unit}, for $B = o(C)$, neglecting $o(1)$, ULA's value scales at least linearly with $B$. The rate at which budget converts to value has two regimes: If the within-unit allocation is effective,
\begin{equation}
\label{eq:lb_V_unit_sublinear_budget_eff_q}
    V_{\rm unit} \ge B \frac{\delta}{c} \big(1 - \frac{q'}{1-q} \big)
    \,,
\end{equation}
and if the within-unit allocation is least effective,
\begin{equation}
\label{eq:lb_V_unit_sublinear_budget_least_eff_q}
    V_{\rm unit} \ge B \frac{\delta}{c} \Big(1 - \frac{q_c - (q - q')}{1 - q} \Big)
    \,.
\end{equation}
\end{theoremEnd}
\begin{proof}
    Recall $T_k/\delta \ge 1 - q - \text{loss}(\rho_k)$. Then, summing the treatment effect over the first $K$~units, we have 
    \begin{equation*}
        V_\text{unit} = \textstyle\sum_{k \in [K]} N\, T_{s(k)} = K N \delta (1 - q) - N \delta \, \text{loss}_K(\vrho)
        \,.
    \end{equation*}
    Plugging the bound on $\text{loss}_K(\vrho)$ from \cref{lem:ub_loss_unit} with $u(K/M) \approx 1$ into this and using $K \approx \frac{B}{N(1-q)c}$ complete the proof.
\end{proof}
These bounds are insightful about the interaction of parameters. First, an effective within-unit allocation makes ULA almost optimal: Looking at \cref{eq:lb_V_unit_sublinear_budget_eff_q}, for a small~$q'$, ULA achieves the optimal rate of~$\delta/c$. The deficiency of ULA scales with $q'/(1-q)$. The dependency on~$q'$ reflects the imperfection in within-unit allocation. The scaling with~$1/(1-q)$ reflects the relative number of units that should be treated compared to the case of~$q=0$. While increasing~$q$ positively influences within-unit allocation towards the effective regime, it comes at the cost of searching for more units: By refusing to allocate to the top~$q$ of the low-welfare units, we may overlook their individuals in need, risking the search over the next units that may not have enough eligible individuals.

When the within-unit allocation is least effective, ULA deviates from optimal allocation. However, the signal from inequality still makes it better than randomly targeting units. Using $q' \le (1 - \bar{\rho}) q$ and $q_c \le \bar{\rho} \, (1 - G_\rho)$, we have
\begin{align}
    \label{eq:ula_extra_value_1}
    V_\text{unit} &\ge B \frac{\delta}{c} \Big(1 - \frac{q_c - (q - q')}{1 - q} \Big) 
    \ge B \frac{\delta}{c} (1 - q_c) \\
    \label{eq:ula_extra_value_2}
    &\ge B \frac{\delta}{c} \Big(\underbrace{1 - \bar{\rho}}_{\substack{\text{value from} \\ \text{random targeting}}} + \underbrace{\bar{\rho}\,G_\rho}_{\substack{\text{extra value} \\ \text{from inequality}}}\Big)
    \,.
\end{align}
\cref{eq:ula_extra_value_1} shows the least effective within-unit allocation is no worse than the case of $q=q'=0$. \cref{eq:ula_extra_value_2} shows ULA leverages inequality signal, attaining extra value proportional to $\bar{\rho} \, G_\rho$.
\ifarxiv

\fi
At this point, it is important to highlight the role of inequality in ULA. Examining \cref{eq:q_c}, we can see that greater inequality relaxes the requirement for an effective~$q$. Therefore, inequality significantly increases the effectiveness of omitting the top~$q$.
Moreover, even when within-unit allocation is least effective, inequality still informs ULA to avoid the high welfare units.
\ifarxiv
These observations underscore the significant role inequality plays in ULA.
\fi

\ifarxiv
\subsection{Establishing Sufficient Conditions}
\fi

\ifarxiv
ULA's value exceeds ILA's for sure if the lower bound on ULA's value surpasses the upper bound on ILA's. We can therefore combine our previous lemmas to give sufficient conditions for ULA to dominate ILA in the next theorem. 
\fi
\begin{theoremEnd}{thm}[Sufficient conditions for a dominant ULA]
\label{thm:dominant_ula_suff_conditions}
Consider ULA with a $(q,q')$-within-unit allocation and a budget~$B$ no more than the cost of treating $M(1-\bar{\rho})$~units. 
Define the normalized Gini coefficient~$\widehat{G}_\rho \coloneqq G_\rho / (1 - \bar{\rho}) \in [0, 1]$, and consider the inequality thresholds
\begin{equation*}
    \widehat{G}_\rho^{(1)} \coloneqq 1 - \frac{q}{\bar{\rho}} \frac{1 - \bar{\rho}}{1 - q}
    \;\;\le\;\; \widehat{G}_\rho^{(2)} \coloneqq 1 - \frac{1}{4} \frac{q}{\bar{\rho}} \frac{1 - \bar{\rho}}{1 - q}\,
    \,.
\end{equation*}
\begin{itemize}
    \item If $\widehat{G}_\rho > \widehat{G}_\rho^{(2)}$, within-unit allocation is effective, and as long as the budget is not excessively large (not surpassing the cost of treating $M(1-\bar{\rho})/2$~units), ULA achieves $\big(1 - q'/(1-q)\big)$ of the maximal value. This implies $\big(1 + O(q')\big)\!\cdot\!{\rm ULA} \succeq {\rm ILA}$. 
    \\
    If further $p(\epsilon)/B > q'/(1-q)$, we have ${\rm ULA} \succ {\rm ILA}\,$.
    
    \item If $\widehat{G}_\rho > \widehat{G}_\rho^{(1)}$, within-unit allocation is still effective, and we obtain similar results as long as the budget meets
    \begin{equation}
    \label{eq:dominant_ula_suff_conditions_linear_budget_low_B}
        \frac{B}{C} < (1 - q) \cdot \max \Big\{1 - \frac{\bar{\rho}}{q}, (1 - \bar{\rho})\big(1 - \sqrt{\frac{q_c}{q}}\big) \Big\}
        \,.
    \end{equation}
    \item In any case, if $p(\epsilon)$ consumes a small but sufficient part of the budget as $p(\epsilon)/B > q_c$, we have ${\rm ULA} \succ {\rm ILA}\,$.
\end{itemize}

\end{theoremEnd}
\begin{proofEnd}
    Recall from \cref{lem:ub_loss_unit} that $\text{loss}_K(\vrho) \le K q'$ if $q > q_u(K/M)$, where $q_u(K/M)$ is defined in \cref{eq:q_u_def} as
    \begin{equation}
    \label{eq:q_u_def_repeated}
        q_u(K/M) \coloneqq \min \, \Big\{ 
        \frac{\bar{\rho}}{1 - \frac{K}{M}},
        \bar{\rho}\frac{1 - \bar{\rho} - G_\rho}{1 - \bar{\rho} - \bar{\rho}G_\rho - 2\frac{K}{M}(1 - \bar{\rho} - \frac{K}{2M})} 
        \Big\}
        \,.
    \end{equation}
    Observe that $q_u(\cdot)$ is an increasing function as long as $K \le M(1 - \bar{\rho})$. So, for $K \le M(1-\bar{\rho})/2$, in order to have $q > q_u(K/M)$, it suffices to have $q > q_u\big((1 - \bar{\rho})/2\big)$. In particular, it suffices to have $q$~larger than the second term in the minimum in \cref{eq:q_u_def_repeated}:
    \begin{equation*}
        q > \bar{\rho}\frac{1 - \bar{\rho} - G_\rho}{1 - \bar{\rho} - \bar{\rho} G_\rho - \frac{3}{4}(1 - \bar{\rho})^2}\,
        \,.
    \end{equation*}
    Rearranging the terms, we find this condition can be met by a sufficiently high inequality: $G_\rho/(1-\bar{\rho}) > 1 - \frac{1}{4}\frac{q}{\bar{\rho}}\frac{1-\bar{\rho}}{1-q}$. 
    
    Even when inequality is not this high, a not-too-large budget can ensure $q > q_u(K/M)$. Solving for $K/M$ in $q > q_u(K/M)$, we obtain
    \begin{equation*}
        \frac{K}{M} < \max \, \Big\{ 1 - \frac{\bar{\rho}}{q},  1 - \bar{\rho} - \sqrt{(1 - \bar{\rho})^2 - (1 - \frac{q_c}{q})(1 - \bar{\rho} - \bar{\rho} G_\rho)} \Big\}
        \,.
    \end{equation*}
    The first term in the maximum is positive only when $q > \bar{\rho}$. Similarly, the second term is also positive only when $q > q_c$. Since $q_c \le \bar{\rho}$, a positive bound on $K/M$ requires $q > q_c$ which is equivalent to a sufficiently high inequality: $G_\rho/(1 - \bar{\rho}) > \frac{1 - q/\bar{\rho}}{1 - q} = 1 - \frac{q}{\bar{\rho}}\frac{1-\bar{\rho}}{1-q}$. In this case, the second term of the maximum will be decreasing in~$G_\rho$. Therefore, it suffices to set $G_\rho$ to its maximum of~$(1-\bar{\rho})$. Then, plugging $K/M \approx \frac{B}{(1-q)C}$ into this condition gives \cref{eq:dominant_ula_suff_conditions_linear_budget_low_B}. 
    
    In the third case, where inequality does not satisfy $G_\rho/(1 - \bar{\rho}) > \frac{1 - q/\bar{\rho}}{1 - q}$, within-unit allocation is least effective. In this case, \cref{lem:ub_loss_unit} suggests that for $q \le q_l(K/M) \coloneqq q_c \, u(K/M)$, the loss will be bounded by $K(q_c - q + q')$ which gives a similar lower bound on $V_\text{unit}$ as \cref{eq:lb_V_unit_sublinear_budget_least_eff_q}. Then a similar argument as the third case of \cref{thm:dominant_ula_suff_conditions_sublinear_budget} shows that ULA will dominate ILA if $p(\epsilon)/B > \frac{q_c - (q-q')}{1 - q}$. There remains to show $q \le q_l(K/M)$. For instance, if the budget is not larger than the amount for treating $M(1-\bar{\rho})/4$ units and $q \le q_c/2$, one can verify that regardless of the level of inequality, we will have $q \le q_l(K/M)$. In the worst case when such conditions don't hold, within-unit allocation may not be effective at all, and we can resort to a blind within-unit allocation corresponding to $q=q'=0$. This gives the sufficient condition of $p(\epsilon)/B > q_c$ for ULA dominance. 
\end{proofEnd}
\cref{tab:dominant_ula_qualitatively,tab:dominant_ula_quantitative} summarize this theorem. \cref{thm:dominant_ula_suff_conditions_sublinear_budget} is a stronger version for the case of sublinear budget~$B = o(C)$. 


\section{Necessary Conditions for Dominant Individual-Level Allocation}
\label{sec:nondominated_ula}

Previously, we showed that under weak assumptions ULA dominates ILA. Now, we show that ILA dominates ULA only when strong necessary conditions are met. Put differently, under weak conditions on the set of profiles~$\gP,$ we can show that ILA does not dominate ULA.
To show that ILA does not dominate ULA, it suffices to find a profile~$\vrho \in \gP$ for which ULA yields a larger value than ILA. 

To approach this result we construct a specific profile~$\vrho$ of small loss. The value achieved by this profile will then serve as a lower bound on ULA's value.

\begin{theoremEnd}{lemma}
\label{lem:ub_loss_unit_best_rho}
Consider ULA with a budget for treating $K$~units that is no more than $M(1-\bar{\rho})$. There exists a profile~$\vrho^*$ with mean~$\bar{\rho}$ and Gini index~$G_\rho$ or less, such that 
\begin{equation}
\label{eq:loss_unit_best_rho}
    {\rm loss}_K(\vrho^*) = K \, {\rm loss} \Big( \max \big\{ \bar{\rho} \, (1 - \frac{M}{K} G_\rho), \, 0 \big\} \Big)
    \,.
\end{equation}
\end{theoremEnd}
\begin{proofEnd}
    Consider the following unit profile:
    \begin{equation*}
        \rho_k^* = \begin{cases}
            \rho_l\,, & k \le K\,, \\
            \rho_u\,, & k > K\,.
        \end{cases}
    \end{equation*}
    Here, $\rho_k^*$s are concentrated around a lower ($\rho_l$) and an upper value ($\rho_u$). For a unit profile with mean~$\bar{\rho}$, we should have $K \rho_l + (M - K) \rho_u = M \bar{\rho}$. Solving for $\rho_u$, we obtain $\rho_u = (M \bar{\rho} - K \rho_l)/(M - K)$. A valid $\rho_u$ lies in the range $1 \ge \rho_u \ge \rho_l$. One can verify the upper bound of~$1$ is already satisfied if $K \le M(1 - \bar{\rho})$ and $\rho_l \ge 0$. A simple calculation also shows the lower bound is satisfied for $\rho_l \le \bar{\rho}$. Therefore, for a valid choice of $\rho_l$ and $\rho_u$, we just need $\bar{\rho} \ge \rho_l \ge 0$.  
    
    Our choice of $\rho_l$ and $\rho_u$ should also result in a Gini index of~$G_\rho$ or less:
    \begin{align*}
        \sum_{k \in [M]} k \rho_{k}^* - \frac{M(M + 1)}{2} \bar{\rho} - \frac{M^2}{2} \bar{\rho} G_\rho &\le 0 \\
        \iff K (K+1) \rho_l + (M + K + 1)(M - K) \rho_u - M(M + 1) \bar{\rho} - M^2 \bar{\rho} G_\rho &\le 0
        \,.
    \end{align*}
    Plugging $\rho_u$ in terms of~$\rho_l$ into this and simplifying equations, we obtain
    \begin{equation*}
        \rho_l \ge \bar{\rho} - \frac{M}{K} G \bar{\rho}
        \,.
    \end{equation*}
    Since $\text{loss}_K(\vrho^*) = K \, \text{loss}(\rho_l)$, we are interested in the smallest~$\rho_l$ possible, which is either the above bound or~$0$. Note that this choice satisfies $\bar{\rho} \ge \rho_l \ge 0$, and results in \cref{eq:loss_unit_best_rho}.

\end{proofEnd}
The proof is constructive.
\ifarxiv
\begin{remark}
Since ${\rm loss}(\cdot)$ is piecewise linear by definition, one can verify that ${\rm loss}_K(\vrho^*)$ in \cref{eq:loss_unit_best_rho} remains piecewise linear in~$K$ (and therefore also $B$). 
\end{remark}
\cref{eq:loss_unit_best_rho} already clarifies the role of inequality: 
\fi
Since $\text{loss}(\cdot)$ is an increasing function, $\text{loss}_K(\vrho^*)$ decreases with~$G_\rho$ to the point where the loss is $0$, and ULA becomes optimal. This happens for $G_\rho \ge K/M$. For a sublinear budget, this condition always holds asymptotically:
\ifarxiv
\begin{corollary}
For a sublinear budget $B = o(C)$, if the set of feasible unit profiles~$\gP$ does not exclude profiles with a Gini index of~$G_\rho$ or less, as long as $G_\rho \ge \frac{B}{(1-q) C} = o(1)$, there exists $\vrho^* \in \gP$ for which ULA is optimal.
\end{corollary}
\fi
The situation is more involved in the case of a linear budget. 
\ifarxiv
Looking at \cref{eq:loss_unit_best_rho} and the definition of $\text{loss}(\cdot)$ in \cref{eq:loss_definition}, for ULA it is desired to have $\bar{\rho} \, \big(1 - (M/K)G_\rho\big) \le q$, so the loss would not exceed~$K q'$. A small~$K$ corresponding to a small~$B$ and the inclusion of high-inequality profiles can make this happen. 
\fi
Next theorem, a counterpart of \cref{thm:dominant_ula_suff_conditions}, states sufficient conditions of $\text{ILA} \nsucc \text{ULA}$ in general:

\begin{theoremEnd}{thm}[Sufficient conditions for a nondominated ULA]
\label{thm:nondominated_ula_suff_conditions}
Consider ULA with a $(q,q')$-within-unit allocation and a budget~$B$ no more than the cost of treating $M(1-\bar{\rho})$~units. Consider a scenario in which the set of feasible unit profiles $\gP$ does not rule out profiles with a Gini index of $G_\rho$ or lower. Define the normalized Gini coefficient $\widehat{G}_\rho \coloneqq G_\rho / (1 - \bar{\rho}) \in [0, 1]$, and consider the inequality threshold
\begin{equation*}
    \widehat{G}_\rho^{(0)} \coloneqq \frac{1}{2} - \frac{1}{2}\frac{q}{\bar{\rho}}
    \,.
\end{equation*}
\begin{itemize}
    \item If $\widehat{G}_\rho \ge \widehat{G}_\rho^{(0)}$, as long as the budget is not excessively large (not exceeding the cost of treating $M(1-\bar{\rho})/2$~units), there exists $\vrho^* \in \gP$ for which ULA attains $\big(1 - q'/(1-q)\big)$ of the maximal value, implying ${\rm ILA} \nsucc \big(1 + O(q')\big)\!\cdot\!{\rm ULA}$. 
    \\If further $p(\epsilon)/B > q'/(1-q)$, we have ${\rm ILA} \nsucceq {\rm ULA}\,$.
    \item In any case, if $p(\epsilon)$ uses a small but sufficient part of the budget as $p(\epsilon)/B > \bar{\rho} \, (1 - \widehat{G}_\rho)$, we have ${\rm ILA} \nsucceq {\rm ULA}\,$.
\end{itemize}
\end{theoremEnd}
\begin{proofEnd}
    As \cref{lem:ub_loss_unit_best_rho} suggests, if $G_\rho \ge K/M$, there exists $\vrho^* \in \gP$ for which ULA's loss is zero. Assuming $K \le M(1 - \bar{\rho})/2$, then it suffices to have $G_\rho / (1 - \bar{\rho}) \ge 1/2$ for an optimal ULA.
    
    If inequality is not this high, we can still hope for a small loss if $\rho_\text{eff} \coloneqq \bar{\rho} \, (1 - (M/K)G_\rho)$ is less than or equal to~$q$. In this case, $\text{loss}(\rho_\text{eff}) \le q$ and $\text{loss}_K(\vrho^*) \le K q'$. 
    Rearranging terms, $\rho_\text{eff} \le q$ can be interpreted as a condition on~$G_\rho$: $G_\rho \ge (K/M)(1 - q/\bar{\rho})$. Using $K \le M(1 - \bar{\rho})/2$, it suffices to have $G_\rho / (1 - \bar{\rho}) \ge (1 - q/\bar{\rho})/2$. Then for a loss of $K q'$, the value of ULA will be at least as the lower bound in \cref{eq:lb_V_unit_sublinear_budget_eff_q}. Therefore, ULA achieves $\big(1 - q'/(1-q)\big)$ of the maximal value of $B \frac{\delta}{c}$, and we get $\text{ILA} \nsucc \big(1 + O(q') \big)\!\cdot\!\text{ULA}$. 
    
    We can get rid of the $O(q')$ term in the previous case if $p(\epsilon)/B$ is sufficiently large. For $\rho_\text{eff} \le q$, we can expand $\text{loss}(\rho_\text{eff})$ using \cref{eq:loss_definition}) and obtain
    \begin{equation*}
        \text{loss}_K(\vrho^*) \le K \, \min \, \{\rho_\text{eff}, q'\} = \min \big\{\bar{\rho} \, (K - M G_\rho), K q'  \big\}
        \,.
    \end{equation*}
    This corresponds to a reduction of 
    \begin{equation*}
        \min \Big\{ B \frac{\bar{\rho}}{1 - q} - \bar{\rho} \, G_\rho C, \, B \frac{q'}{1-q} \Big\} \frac{\delta}{c}
    \end{equation*}
    from ULA's value. Since ILA would lose at least a value of~$p(\epsilon) \frac{\delta}{c}$, if $p(\epsilon)/B > q'/(1-q)$ or $p(\epsilon) > \bar{\rho} \, (B/(1-q) - G_\rho C)$ holds, ULA will have a larger value than ILA for~$\vrho^*$, and we have $\text{ILA} \nsucceq \text{ULA}$.
    
    Finally, if $\rho_\text{eff} > q$, within-unit allocation is not effective and we can use a blind within-unit allocation instead. For $q=q'=0$, \cref{eq:loss_definition} simply gives $\text{loss}(\rho_\text{eff}) = \rho_\text{eff}$. Then a similar argument as the previous case shows ULA would beat ILA for~$\vrho^*$ if $p(\epsilon) > \bar{\rho} \, \big(B - G_\rho C \big)$. Since by assumption $K \le M (1 - \bar{\rho})$, we have $C \ge B/(1 - \bar{\rho})$. Therefore, it is sufficient to have $p(\epsilon)/B > \bar{\rho} \, \big(1 - G_\rho/(1 - \bar{\rho}) \big)$.

\end{proofEnd}
\cref{tab:nondominated_ula_qualitative} summarizes the theorem.
\begin{remark}
If $q > \bar{\rho}$, both $\widehat{G}_\rho^{(0)}$ and $\widehat{G}_\rho^{(1)}$ will be negative. Otherwise, $\widehat{G}_\rho^{(0)} \le \widehat{G}_\rho^{(1)}/2 \le \widehat{G}_\rho^{(2)}/2$. 
\end{remark}
Contrasting Theorems~\ref{thm:nondominated_ula_suff_conditions}~and~\ref{thm:dominant_ula_suff_conditions} using the above remark, a nondominated ULA requires weaker conditions than a dominant ULA as expected: Even for $q=q'=0$, all of the cases in \cref{thm:nondominated_ula_suff_conditions} are still plausible. 
\ifarxiv
To illustrate the results, \cref{fig:ula_regimes} gives a more precise version of \cref{fig:ula_regimes_simplified} in terms of the three different inequality thresholds that come out of \cref{thm:dominant_ula_suff_conditions,thm:nondominated_ula_suff_conditions}.

\begin{figure}[h]
    \centering
    \includegraphics{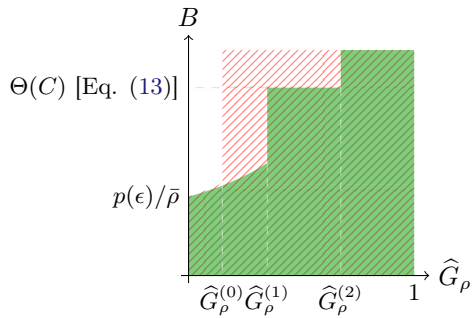}
    \caption{Sufficient conditions for dominant ULA (green) and nondominated ULA (red) for $q \le \bar{\rho}$.}
    \label{fig:ula_regimes}
\end{figure}
\fi

\begin{table*}[t]
    \centering
    \caption{Sufficient conditions for a nondominated ULA. Refer to \cref{thm:nondominated_ula_suff_conditions} for accurate statements. For a similar table with quantitative values for low, medium, and high, refer to \cref{tab:nondominated_ula_quantitative}. *In general, compared to \cref{tab:dominant_ula_qualitatively}, a high price of prediction and a medium or higher inequality require weaker conditions. \smallskip}
    \begin{tabular}{lcc}
                                    & Medium or higher* inequality      & Low inequality \\
        \midrule
        High* price of prediction   & $\text{ILA} \nsucceq \text{ULA}$  & $\text{ILA} \nsucceq \text{ULA}$ \\
        \cline{3-3}
        Medium price of prediction  & $\text{ILA} \nsucceq \text{ULA}$  & \multicolumn{1}{|c|}{\multirow{2}{*}{?}} \\
        Low price of prediction     & $\text{ILA} \nsucc \big(1+O(q')\big)\!\cdot\!\text{ULA}$ & \multicolumn{1}{|c|}{} \\
        \cline{3-3}
    \end{tabular}
    \label{tab:nondominated_ula_qualitative}
\end{table*}

\section{Heterogeneous Intervention Effects}
\label{sec:het}

Thus far, we assumed intervention has a fixed effect of~$\delta$ capped at the maximum welfare of~$1$. In this section, we allow for a general intervention effect~$\tau(w)$. We show that, under certain assumptions on the Lipschitzness of $\tau$ and limited concentration of welfare distribution, a sufficiently high inequality or a small budget ensures ULA outperforms ILA and, in fact, is close to the optimal allocation.

\paragraph{Notation and Preliminaries.} We assume $\tau(\cdot)$ is a decreasing function with $\tau(0) = \delta'$ and $\tau(1) = 0$. The average treatment effect at unit~$k$ with a $(q, q')$-within-unit allocation is denoted by~$T_k$. If everyone in~$k$ is treated, the average effect is denoted by~$T'_k$. The average and Gini index of $T'_k$s are represented by~$\barT'$ and~$G_T$, respectively. Our focus is on a setting with inequality~$G_T$ and mean treatment effect~$\barT'$ on everyone, aiming to bound the advantage that a heterogeneous $\tau$ can give to ILA. Such an approach is most effective when we make assumptions about welfare distribution. We denote the density of welfare at unit~$k$ by~$p_w^k$ and the overall welfare density by~$p_w$. Abusing notation, we use $V_\text{unit}$ and $V_\text{ind}$ to denote the expected ULA and ILA's values.

\ifarxiv
\subsection{Upper bounding the Value of ILA}
\fi

\cref{eq:V_ind_ub} shows that after paying for prediction, every remaining dollar can turn into value at a rate of~$\delta'/c$. With no assumption on welfare distribution, there would be enough individuals with a $0$-welfare and no one in $[0, 2\epsilon]$, making this bound tight. We avoid such cases by imposing the following assumption.

\begin{assumption}
\label{assump:overall_bounded_density}
The overall welfare density is bounded from above and below: $\gamma \ge p_w(w) \ge \ugamma$.   
\end{assumption}

Prediction accuracy was less relevant when the treatment effect was homogeneous and there were sufficient low-welfare individuals. However, in the case of a heterogeneous effect, distinguishing small differences in welfare is valuable. We propose a model for prediction error in the following.
Recall ILA sorts individuals in terms of their predicted welfare~$\hat{w}$ and allocates to those with the lowest welfare. This is equivalent to allocating to any individual~$i$ with $\hat{w}_i \le w_t$ for the largest possible choice of a threshold~$w_t$ that meets the budget constraint. We assume prediction errors are symmetric and independent. Formally, for an individual with welfare~$w$, we have $\Pr(\hat{w} \le w_t \mid w) = \sigma(w_t - w)$, where $\sigma(\Delta w)$ is $1$ for $\Delta w > \epsilon$, and is $0$ for $\Delta w < -\epsilon$. Denoting the derivative of~$\sigma$ by~$\sigma'$, we assume $1/\epsilon \ge |\sigma'(\Delta w)| \ge 1/(2\epsilon)$ for $|\Delta w| \le \epsilon$. The budget constraint requires the number of treated to not exceed $(B - p(\epsilon))/c$. We relax this constraint and assume the budget should be met in expectation.

When $\tau$ changes sharply, distinguishing individuals with a small difference in welfare gains significant value. We control this advantage by considering a Lipschitz continuous~$\tau$. By leveraging \cref{assump:overall_bounded_density} and Lipschitzness, next, we provide a tighter bound on ILA's value.

\begin{theoremEnd}{thm}
\label{thm:heter_ila_ub}
Consider ILA with an $\epsilon$-accurate welfare predictor ($\epsilon \le 1/2$) that costs~$p(\epsilon)$. Define the remaining budget as $\widetilde{B} \coloneqq B - p(\epsilon)$. Under \cref{assump:overall_bounded_density}, suppose $\widetilde{B}/C$ is no less than $\ugamma \epsilon$ and no more than~$\gamma (1 - \epsilon)$. For a concave and differentiable~$\tau$ with $l \ge |\od{\tau}{w}| \ge l/2$, we have
\begin{equation}
\label{eq:heter_ila_ub}
    V_{\rm ind} \le \widetilde{B} \frac{1}{c} \Big( \delta' - \frac{l}{8 \gamma} (\widetilde{B}/C + \ugamma \epsilon /2 ) \Big)
    \,.
\end{equation}
\end{theoremEnd}
\begin{proofEnd}
    The proof has two parts. First, for a fixed budget, we characterize the welfare density that maximizes $V_\text{ind}$ subject to \cref{assump:overall_bounded_density}. Then for such a class of densities, we upper bound $V_\text{ind}$ as a function of the budget. Throughout the proof, we denote the remaining budget by~$\widetilde{B} \coloneqq B - p(\epsilon)$ and define $\ogamma \coloneqq \gamma - \ugamma$.
    
    Given a welfare density~$p_w$ consistent with \cref{assump:overall_bounded_density}, define a new density $q_w(w) \coloneqq p_w(w) - u \, \delta(w - w_1) + u \, \delta(w - w_2)$, where $\delta(\cdot)$ is Dirac delta function and $u \ge 0$ is chosen such that \cref{assump:overall_bounded_density} is not violated. For a fixed~$w_t$, increasing~$u$ from~$0$ changes the budget by
    \begin{equation*}
        \frac{1}{C} \pd{\widetilde{B}}{u} = -\sigma(w_t - w_1) + \sigma(w_t - w_2)
        \,.
    \end{equation*}
    Without loss of generality, suppose $w_2 > w_1$. Then a nondecreasing~$\sigma$ implies $\pd{\widetilde{B}}{u} \le 0$. To spend the same budget, we need to increase~$w_t$:
    \begin{equation*}
        \frac{1}{C} \pd{\widetilde{B}}{w_t} = \E[\sigma'(w_t - w)] \ge 0
        \,.
    \end{equation*}
    Since we calculate the derivative at~$u=0$, the expectation is w.r.t.~$w \sim p_w$. Putting these together, a fixed spending requires
    \begin{equation*}
        \od{w_t}{u} = \frac{\sigma(w_t - w_1) - \sigma(w_t - w_2)}{\E[\sigma'(w_t - w)]}
        \,.
    \end{equation*}
    When we increase~$w_t$ in this way, the (expected) value of ILA changes by
    \begin{align}
    \label{eq:_proof_heter_ila_dv_du}
        \frac{1}{M N} \od{V_\text{ind}}{u} &= -\tau(w_1)\sigma(w_t - w_1) + \tau(w_2)\sigma(w_t - w_2) + \od{w_t}{u} \, \E[\tau(w) \sigma'(w_t - w)] \nonumber \\
        &= -\tau(w_1)\sigma(w_t - w_1) + \tau(w_2)\sigma(w_t - w_2) + \big(\sigma(w_t - w_1) - \sigma(w_t - w_2)\big) \frac{\E[\tau(w) \sigma'(w_t - w)]}{\E[\sigma'(w_t - w)]}
        \,.
    \end{align}
    Consider the special case of~$w_2 = 1$. At this point~$\tau(w_2) = 0$. We assume $\widetilde{B}$ is more than $\ugamma \epsilon$ and less than $\gamma (1 - \epsilon)$. This ensures for a tight choice of~$w_t$, $1 - \epsilon \ge w_t \ge \epsilon$ which implies $\sigma(w_t) = 1$ and $\sigma(w_t - 1) = 0$. We also relax \cref{assump:overall_bounded_density} at the right-most boundary and assume density can be concentrated at~$w=1$. Then \cref{eq:_proof_heter_ila_dv_du} shows a mass can move from~$w_1$ to~$1$ while increasing the value if and only if $\tau(w_1) < \E[\tau(w) \sigma'(w_t - w)]/\E[\sigma'(w_t - w)]$. Therefore, we can characterize the value-maximizing~$p_w$ for $w < 1$ by $p_w(w) = \gamma \, \One\{w \le a\} + \ugamma \, \One\{w > a\}$, where
    \begin{equation*}
        \tau(a) = \frac{\E[\tau(w) \sigma'(w_t - w)]}{\E[\sigma'(w_t - w)]}
        \,.
    \end{equation*}
    This is of course a recursive formula and not the full characterization. However, it allows us to bound~$a$ in terms of~$w_t$: 
    \begin{itemize}
        \item Using the concavity of~$\tau$ and the symmetry of~$\sigma'$, a simple argument in terms of the worst-case~$\sigma'$ shows
        \begin{equation}
        \label{eq:_proof_heter_ila_tau_a_lb_int}
            \tau(a) \ge \frac{\ugamma \int_{w_t - \epsilon}^{w_t + \epsilon} \frac{1}{2\epsilon} \tau(w) \dif w + \ogamma \int_{w_t - \epsilon/2}^a \frac{1}{\epsilon} \tau(w) \dif w}{\ugamma + \ogamma(a - w_t - \epsilon/2)/\epsilon}
            \,.
        \end{equation}
        Since $l \ge |\od{\tau}{w}| \ge l/2$, we can further lower bound the integrals in the numerator:
        \begin{align*}
            \frac{1}{2\epsilon} \int_{w_t - \epsilon}^{w_t + \epsilon} \tau(w) \dif w &\ge \frac{1}{2} \big(\tau(w_t) - l \epsilon/2 + \tau(t) + l \epsilon/4 \big) = \tau(w) - l \epsilon/8,\\ 
            \frac{1}{\epsilon} \int_{w_t - \epsilon/2}^a \tau(w) \dif w &\ge \big(\tau(a) + l(a - w_t + \epsilon/2)/4\big) (a - w_t + \epsilon/2)/\epsilon
            \,.
        \end{align*}
        Plugging these into \cref{eq:_proof_heter_ila_tau_a_lb_int} and simplifying equations, we obtain
        \begin{equation*}
            \tau(a) - \tau(w_t) \ge -\frac{l \epsilon}{8} + \frac{\ogamma}{\ugamma} \frac{l}{4} \frac{(a - w_t + \epsilon/2)^2}{\epsilon}
            \,.
        \end{equation*}
        Using $\tau(a) - \tau(w_t) \le l(w_t - a)$, we can further simplify the above equation and obtain
        \begin{equation*}
            a \le w_t - \alpha \epsilon
            \,,
        \end{equation*}
        where
        \begin{equation*}
            \alpha \coloneqq \frac{1}{2} + 2 (\ugamma/\ogamma) - \sqrt{4 (\ugamma/\ogamma)^2 + (5/2) (\ugamma/\ogamma)}
            \,.
        \end{equation*}
        Note that for $\gamma \ge 3 \ugamma$, we have $\alpha \in [0, 1/2]$.
    
        \item Using Jensen's inequality for a concave~$\tau$ and the symmetry of~$\sigma'$, one can show
        \begin{equation}
        \label{eq:_proof_heter_ila_tau_a_ub_int}
            \tau(a) \le \frac{\ugamma \tau(w_t) + \ogamma \int_{w_t - \epsilon}^a \frac{1}{2\epsilon} \tau(w) \dif w}{\ugamma + \ogamma(a - w_t - \epsilon)/(2\epsilon)}
            \,.
        \end{equation}
        Since $l \ge |\od{\tau}{w}| \ge l/2$, we can further upper bound the integral in the numerator:
        \begin{equation*}
            \frac{1}{2\epsilon} \int_{w_t - \epsilon}^a \tau(w) \dif w \le \big(\tau(a) + l (a - w_t + \epsilon)/2\big) (a - t + \epsilon)/(2 \epsilon)
            \,.
        \end{equation*}
        Plugging this into \cref{eq:_proof_heter_ila_tau_a_ub_int} and simplifying equations, we obtain
        \begin{equation*}
            \tau(a) - \tau(w_t) \le \frac{\ogamma}{\ugamma} \frac{l}{4} \frac{(a - w_t + \epsilon)^2}{\epsilon}
            \,.
        \end{equation*}
        Using $\tau(a) - \tau(w_t) \ge l(w_t - a)/2$, we can further simplify the above equation and obtain
        \begin{equation*}
            a \ge w_t - \beta \epsilon
            \,,
        \end{equation*}
        where
        \begin{equation*}
            \beta \coloneqq 1 + (\ugamma/\ogamma) - \sqrt{(\ugamma/\ogamma)^2 + 2 (\ugamma/\ogamma)}
            \,.
        \end{equation*}
    \end{itemize}
    
    Next, we upper bound~$V_\text{ind}$ for the partially characterized optimal~$p_w$. Using $|\od{\tau}{w}| \ge l/2$, for a nondecreasing~$\sigma$, we have
    \begin{equation*}
        \frac{1}{M N} V_\text{ind} \le \tau(0) \E[\sigma(w_t - w)] - \gamma \frac{l a^2}{4} \sigma(w_t - a) - \ugamma (w_t - a) \frac{l a}{2} \sigma(0) - \ugamma  \frac{l w_t}{2} \int_{w_t}^1 \sigma(w_t - w) \dif w 
        \,.
    \end{equation*}
    When all of the budget~$\widetilde{B}$ is spent, the first term is equal to~$\delta' \, \widetilde{B}/C$. Using $\sigma'(\Delta w) \ge 1/\epsilon$ and $w_t \le 1 - \epsilon$, we can lower bound the integral by~$\epsilon/8$. Since $a \le w_t$, we also have $\sigma(w_t - a) \ge \sigma(0) = 1/2$. Plugging these into the above equation, we obtain
    \begin{equation*}
        \frac{1}{M N} V_\text{ind} \le \widetilde{V}(w_t, a) \coloneqq \delta' \, \widetilde{B}/C - \gamma \frac{l a^2}{8} - \ugamma (w_t - a) \frac{l a}{4} - \ugamma  \frac{l w_t \epsilon}{16}
        \,.
    \end{equation*}
    When the budget constraint is met tightly, using the symmetry of~$\sigma$ and $1 - \epsilon \ge w_t \ge \epsilon$, we have
    \begin{equation*}
        \widetilde{B}/C = \E[\sigma(w_t - w)] \le \ogamma a + \ugamma w_t
        \,.
    \end{equation*}
    We call this the budget constraint. Maximizing $\widetilde(w_t, a)$ subject to this budget constraint and two additional constraints $a \le w_t - \alpha \epsilon$ and $a \ge w_t - \beta \epsilon$ shows that when $\ogamma \ge 4 \ugamma \beta$, the budget constraint is tight and $a = w_t - \alpha \epsilon$. In this case,
    \begin{equation*}
        \widetilde{V}(w_t, a) = \delta' \widetilde{B}/C - \frac{l}{8 \gamma} \Big((\widetilde{B}/C - \ugamma \alpha \epsilon)^2 + 2 \ugamma \alpha \epsilon (\widetilde{B}/C - \ugamma \alpha \epsilon) + \ugamma \epsilon (\widetilde{B}/C + \ogamma \alpha \epsilon)/2 \Big)
        \,.
    \end{equation*}
    Then a straightforward calculation shows that for $\ogamma \ge 2 \ugamma \alpha$, 
    \begin{equation*}
        \frac{1}{M N} V_\text{ind} \le \widetilde{V}(w_t, a) \le (\widetilde{B}/C) \Big( \delta' - \frac{l}{8 \gamma} (\widetilde{B}/C + \ugamma \epsilon /2 ) \Big)
        \,.
    \end{equation*}
\end{proofEnd}
Compared to the optimal rate of~$\delta'/c$, \cref{eq:heter_ila_ub} reflects two deficiencies: First, a larger budget implies that more individuals will be treated, including those with treatment effects farther from $\delta'$. Second, lower accuracy means that more individuals with high welfare values may be recognized as eligible.

\ifarxiv
\subsection{Lower bounding the Value of ULA}
\fi

ULA's previous bound can be reused in this general setting if, for a threshold~$w_t$, we lower bound $\tau$ with $\tilde{\tau}(w) = \delta \cdot \One \{w \le w_t\}$, where $\delta = \tau(w_t)$. This approach underestimates the effect on the low-welfare population by an amount as large as $\delta' - \delta$ and neglects the effect on the population above~$w_t$. So, there might be no good choice of~$w_t$ in this tradeoff. Next, we get around this by proposing ULA based on estimated unit-level average treatment effects instead of $\rho_k$s. Suppose ULA sorts units in terms of $T'_k$s and allocates to the top~$K$ units.
We assume such estimates are available for free. We will show in \cref{sec:learning} that under some assumptions on~$\tau$, the additional cost of learning reduces ULA's value negligibly. The following assumption helps us avoid contrived worst-case scenarios and tightly relate $T'_k$ and $T_k$.

\begin{assumption}
\label{assump:gamma_bounded_density}
For every unit~$k$, assume $p_w^k(w) \le \gamma$. 
\end{assumption}
\begin{theoremEnd}{lemma}[$T'_k$ to $T_k$ conversion]
\label{lem:min_T_gamma_bounded}
For a $(q, q')$-within-unit allocation, under \cref{assump:gamma_bounded_density}, 
we have
\begin{equation*}
    T_k \ge Q_\tau(T'_k - q'\delta'; \; q, q', \gamma)
    \,,
\end{equation*}
%
where
\begin{align*}
    Q_\tau(t; q, q', \gamma) &\coloneqq \gamma \, \Gamma_{(1 - q - q')/\gamma}\Big( \Gamma^{-1}_{(1 - 2q')/\gamma}\big(t/\gamma\big)\Big)\,, \\
    \Gamma_a(w) &\coloneqq \int_{w}^{w + a} \tau(x) \dif x
    \,.
\end{align*}
\end{theoremEnd}
\begin{proofEnd}
    Our goal is to find a welfare density that minimizes treatment effect~$T_k$ while adhering to~$T'_k$ and \cref{assump:gamma_bounded_density}. Define $w_{(\alpha)}^k$ as the $\alpha^\text{th}$ quantile of welfare at unit~$k$. For notational brevity, we drop the superscript~$k$ in the following. Let $p^*_w$ be the optimal welfare density with the $\alpha^\text{th}$ quantile denoted by~$w^*_{(\alpha)}$. We show a new valid welfare density~$p_w$ can be constructed from~$p^*_w$ that gives a~$T_k$ lower or equal to what $p^*_w$ could give. Fix $w^*_{(q')}$, $w^*_{(1-q)}$, and $w^*_{(1-q')}$. Define four regimes for welfare:
    \begin{equation*}
        \gR_1 = [0, w^*_{(q')}), \; \gR_2 = [w^*_{(q')}, w^*_{(1-q)}), \; \gR_3 = [w^*_{(1-q)}, w^*_{(1-q')}), \; \gR_4 = [w^*_{(1-q')}, 1]
        \,.
    \end{equation*}
    Note that during within-unit allocation, in the worst case, individuals from the top $q'$~fraction will replace those in the bottom $q'$~fraction and receive treatment. In this case, $T_k$ will only come from the individuals with welfare in $\gR_2$ and $\gR_4$. Since $\tau(\cdot)$ is nonincreasing, for fixed boundaries of the regimes, our aim is to move the mass within these $\gR_2$ and $\gR_4$ towards larger values while adhering to \cref{assump:gamma_bounded_density}. As the first step, we relax the problem and assume \cref{assump:gamma_bounded_density} only needs to be held in $\gR_2$ and $\gR_3$. The minimum of~$T_k$ under this relaxation serves as a lower bound for the original problem. If there exist $w^+ \in \gR_2 \cup \gR_4$, $w^- \in \gR_1 \cup \gR_3$, $\Delta^+ \ge 0$, and $\Delta^- \ge 0$ such that
    \begin{enumerate}
        \item Either $\Delta^+$ or $\Delta^-$ is not zero ,
        \item $w^+ + \Delta^+$ and $w^- - \Delta^-$ lie in the same regimes as $w^+$ and $w^-$ ,
        \item $p^*_w(w^+) > 0$ and $p^*_w(w^-) > 0$ ,
        \item If $w^+ \in \gR_2$, $p^*_w(w^+ + \Delta^+) < \gamma$ ,
        \item If $w^- \in \gR_3$, $p^*_w(w^- - \Delta^-) < \gamma$ ,
        \item $\tau(w^+ + \Delta^+) + \tau(w^- - \Delta^-) = \tau(w^+) + \tau(w^-)$ ,
    \end{enumerate}
    then construct a new solution
    \begin{equation}
    \label{eq:_proof_min_T_gamma_bounded}
        p_w(w) = p^*_w(w) + d \, \One\{w = w^+ + \Delta^+\} - d \, \One\{w = w^+\} + d \, \One\{w = w^- - \Delta^-\} - d \, \One\{w = w^-\}
        \,,
    \end{equation}
    where $d > 0$ is the maximum value that keeps $p_w$ a valid probability density function. For instance, if $w^+ \in \gR_2$ and $w^- \in \gR_3$,
    \begin{equation*}
        d = \min \big\{ p^*(w^+) - 0, p^*(w^-) - 0, \gamma - p^*(w^+ + \Delta^+), \gamma - p^*(w^- - \Delta^-) \big\}
        \,.
    \end{equation*}
    It is straightforward to verify that the constructed~$p_w$ preserves~$T'_k$ and meets \cref{assump:gamma_bounded_density} at $\gR_2$ and $\gR_3$, but has a $T_k$ lower or equal to what $p^*_w$ could give. By repetitively applying this operation for all $w^+$, $w^-$, $\Delta^+$, and $\Delta^-$ that meet the conditions, we obtain a density function that follows one of the following structures:
    \begin{enumerate}
        \item All the mass in $\gR_2$ and $\gR_4$ has moved towards the right-most boundaries, and all the mass in $\gR_1$ and $\gR_3$ has moved towards the left-most boundaries (\cref{fig:_proof_min_T_gamma_bounded_3}):
        \begin{equation}
        \label{eq:_proof_min_T_gamma_bounded_structure}
            p_w(w) = \gamma \, \One \Big\{w \in \big[w^*_{(1-q)} - \frac{1-q-q'}{\gamma}, w^*_{(1-q)} + \frac{q - q'}{\gamma} \big] \Big\}
            + q' \, \delta(w) + q' \, \delta(w-1)
            \,.
        \end{equation}
        Here, $\delta(\cdot)$ is the Dirac delta function.
    
        \item All the mass in $\gR_2$ and $\gR_4$ has moved towards the right-most boundaries. Choose $w^+ \in \gR_3$, $w^- \in \gR_3$, $\Delta^+ \ge 0$, and $\Delta^- \ge 0$ such that all the previous conditions are met and additionally $w^+ > w^-$. Then applying a similar operation as \cref{eq:_proof_min_T_gamma_bounded} multiple times, move the mass in~$\gR_3$ towards its boundaries. Similarly, if we choose $w^+$ and $w^-$ from~$\gR_1$ and apply the operations multiple times, the mass in~$\gR_1$ will be concentrated on its boundaries. These operations maintain~$T'_k$. They also do not change~$T_k$ as $T_k$ only depends on~$p_w$ in~$\gR_2$ and~$\gR_4$. These operations make~$p_w$ look like \cref{fig:_proof_min_T_gamma_bounded_1}.
        Now consider increasing $w_{(1-q)}$ by moving the block around it forward while decreasing $w_{(1-q')}$ such that the resulting distribution preserves~$T'_k$. Then at some point two blocks of mass merge as \cref{fig:_proof_min_T_gamma_bounded_2}. Since the mass under~$\gR_2$ is only moving forward, this operation can only decrease~$T_k$. Keep moving the merged block forward while decreasing~$w_{(q')}$ to preserve~$T'_k$. This operation also can only decrease~$T_k$. As $w_{(q')}$ approaches zero, we obtain a density like \cref{fig:_proof_min_T_gamma_bounded_3}, which has a similar structure as the first case.
        
        \begin{figure}[h]
            \centering
            \subfigure[]{\includegraphics{figs/_proof_min_T_gamma_bounded_1.tex}\label{fig:_proof_min_T_gamma_bounded_1}}
            \subfigure[]{\includegraphics{figs/_proof_min_T_gamma_bounded_2.tex}\label{fig:_proof_min_T_gamma_bounded_2}}
            \subfigure[]{\includegraphics{figs/_proof_min_T_gamma_bounded_3.tex}\label{fig:_proof_min_T_gamma_bounded_3}}
            \caption{Demonstrating the proof of \cref{lem:min_T_gamma_bounded}.}
            \label{fig:_proof_min_T_gamma_bounded}
        \end{figure}
    
        \item All the mass in $\gR_1$ and $\gR_3$ has moved towards the left-most boundaries. We follow similar steps as the previous case. Choose $w^+ \in \gR_2$, $w^- \in \gR_2$, $\Delta^+ \ge 0$, and $\Delta^- \ge 0$ such that all the required conditions are met and additionally $w^+ > w^-$. Then applying a similar operation as \cref{eq:_proof_min_T_gamma_bounded} multiple times, move the mass in~$\gR_2$ towards its boundaries. Similarly, if we choose $w^+$ and $w^-$ from~$\gR_4$ and apply the operations multiple times, the mass in~$\gR_4$ will be concentrated on its boundaries. These operations maintain~$T'_k$. Define $T^c_k \coloneqq T'_k - T_k$. Since $T^c_k$ only depends on~$p_w$ in~$\gR_1$ and~$\gR_3$, these operations do not change~$T^c_k$, consequently leaving~$T_k$ intact.
        Now consider decreasing $w_{(1-q)}$ by moving the block around it forward while decreasing $w_{(q')}$ such that the resulting distribution preserves~$T'_k$. Then at some point two blocks of mass merge. This operation can only increase~$T^c_k$ and therefore decreases~$T_k$. Keep moving the merged block forward while increasing~$w_{(1-q')}$ to preserve~$T'_k$. This operation also can only decrease~$T_k$. As $w_{(1-q')}$ approaches one, we obtain a density like \cref{fig:_proof_min_T_gamma_bounded_3}, which has a similar structure as the first case.
    \end{enumerate}
    
    The above analysis shows there exists an optimal~$p_w$ with the structure following \cref{eq:_proof_min_T_gamma_bounded_structure} (\cref{fig:_proof_min_T_gamma_bounded_3}). This solution should give an average treatment effect of~$T'_k$ when everyone gets treated. We can write~$T'_k$ and~$T_k$ resulting from~$p_w$ of \cref{eq:_proof_min_T_gamma_bounded_structure} as
    \begin{align*}
        T'_k &= q'\delta' + \gamma \, \Gamma_{(1 - 2q')/\gamma}\big( w_{(1-q)} - (1-q-q')/\gamma\big)\,, \\
        T_k &= \gamma \, \Gamma_{(1 - q - q')/\gamma}\big( w_{(1-q)} - (1-q-q')/\gamma \big)
        \,.
    \end{align*}
    Then solving for $w_{(1-q)}$ from the first equation and plugging it into the second equation completes the proof. 
\end{proofEnd}

\begin{theoremEnd}[no proof end, proof here]{remark}
For every~$\gamma > 0$, \ifarxiv we have \fi $Q_\tau(t; q, q', \gamma) \ge t \, \frac{1 - q - q'}{1 - 2q'}$, where equality corresponds to $\gamma \rightarrow \infty$ 
\ifarxiv
(as we did not have \cref{assump:gamma_bounded_density}).
\fi
\end{theoremEnd}
\begin{proof}
    For $b \ge a > 0$ and a decreasing~$\tau$, $\Gamma_a(w)/\Gamma_b(w) \ge a/b$. Let $w = \Gamma^{-1}_b(t)$. Then $\Gamma_a\big(\Gamma^{-1}_b(t)\big) \ge (a/b) t$. 
    Defining $a = (1 - q - q')/\gamma$, $b = (1 - 2q')/\gamma$, and $t = (T'_k - q' \delta')/\gamma$ completes the proof.
\end{proof}
In \cref{lem:min_T_gamma_bounded}, $Q_\tau$ transforms $T'_k$ into $T_k$ for a single unit. A property of $\tau$ will enable us to utilize $Q_\tau$ for converting an average treatment effect~$\barT'_k$ to $V_\text{unit}$:

\begin{property}
\label{prop:decreasing_deriv}
Given $w_c$, $a$, and $b$ ($b \ge a > 0$), $\tau$ has \cref{prop:decreasing_deriv} if $r(w) \coloneqq \frac{\tau(w) - \tau(w + a)}{\tau(w) - \tau(w + b)}$ is decreasing for $w \le w_c$.
\end{property}
It can be immediately observed that a strictly decreasing $\tau$ has \cref{prop:decreasing_deriv} for any~$w_c$ if $a=b$. See \cref{prop:increasing_deriv} for a necessary characterization of general treatment effects that have \cref{prop:decreasing_deriv}. In fact, this property holds for many of the typical~$\tau$s,
for example:
\begin{example}
\label{ex:prop_l_lip}
Consider a treatment effect of~$\delta'$ at $w=0$ decreasing linearly with~$w$ and capped at maximum welfare:
\begin{equation*}
    \tau(w) = \min \, \{w + (\delta' - l w), 1\} - w
    \,.
\end{equation*}
Here, $\delta' = \delta + (1 - \delta)l$ and $l > 0$. One can verify that for any $a$ and $b$ ($b \ge a > 0$), $\tau$ has \cref{prop:decreasing_deriv} for $w_c = 1 -\delta - a$. 
\end{example}
In \cref{thm:heter_ula_lb}, we show that for a $\tau$ satisfying \cref{prop:decreasing_deriv},
\begin{equation}
\label{eq:heter_ula_lb_repeated}
    V_{\rm unit} \ge K Q_\tau \big(\barT'/(1 - G_T) - q'\delta'; \; q, q', \gamma\big)
    \,.
\end{equation}
So, effectively, ULA chooses a unit with $T' = \barT'/(1 - G_T)$ --- clearly showing how inequality can make ULA extremely more effective. We skip the details of the theorem here and directly utilize it to identify a dominant ULA next.
 

\ifarxiv
\subsection{Sufficient Conditions For a Dominant ULA}
\fi

Our specific interest lies in the \emph{elbow-shaped} treatment effect (\cref{fig:elbow_tau}), where, starting from $\tau(0) = \delta'$, an $l$-Lipschitz effect gradually decreases and reaches $\delta$. Beyond this point, the welfare after intervention will reach its cap of~$1$, limiting the effect. \cref{ex:prop_l_lip} is a special case of this model that satisfies \cref{prop:decreasing_deriv}. If $\tau$ of \cref{fig:elbow_tau} similarly satisfies \cref{prop:decreasing_deriv}, \cref{thm:heter_ula_lb} enables us to bound $V_\text{unit}$ as \cref{eq:heter_ula_lb_repeated}. Then a comparison to \cref{thm:heter_ila_ub} can further establish sufficient conditions for a dominant ULA. The next theorem proposes such conditions for a small~$l$. Consistent with our previous observations, high inequality or a reasonably bounded budget can ensure that ULA is effective and dominant.

\ifarxiv
\begin{figure}[h]
    \centering
    \includegraphics{figs/elbow_tau.tex}
    \caption{An elbow-shaped heterogeneous effect.}
    \label{fig:elbow_tau}
\end{figure}
\fi

\begin{thm}[Sufficient conditions for a dominant ULA in case of a heterogeneous effect]
\label{thm:heter_dominant_ula_short}
Consider ULA with a $(q, q')$-within-unit allocation and a budget~$B$ that is no more than the cost of treating $M \barT'/\delta'$ units. 
Under \cref{assump:gamma_bounded_density} and conditions of \cref{thm:heter_ila_ub}, suppose an elbow-shaped effect with \cref{prop:decreasing_deriv} (as in \cref{thm:heter_ula_lb}). Consider the inequality thresholds
\ifnotarxiv
\else
\begin{equation}
    G_T^{(1)} \coloneqq 1 - \frac{\barT'}{\delta' - q^2/(2\gamma)}
    \;\;\le\;\;
    G_T^{(2)} \coloneqq 1 - \frac{1}{4} \frac{\barT'}{\delta' - q^2/(2\gamma)} - \frac{3}{4} \frac{\barT'}{\delta'}
    \,.
\end{equation}
\fi

\begin{itemize}
    \item If $G_T > G_T^{(2)}$, neglecting $O(l^2)$ and $O(q'^2)$, ULA achieves $(1 - q'/(1-q) - l/\delta)$ of the optimal value. 
    \\If further, 
    \begin{equation}
    \label{eq:heter_dominant_ula_p_eps}
        \frac{p(\epsilon)}{B} > \frac{q'}{1-q} + \frac{l}{\delta} - \frac{l}{\delta} \frac{1}{8 \gamma} \big(\frac{B}{C} + \ugamma \epsilon/2 \big)
        \,,
    \end{equation}
    we have ${\rm ULA} \succ {\rm ILA}\,$.

    \item If $G_T > G_T^{(1)}$ and the budget meets 
    \begin{equation*}
        \frac{B}{C} < \frac{1}{2} (1-q)(2\gamma \delta'/q^2 - 1) (G_T - G_T^{(1)})
        \,,
    \end{equation*}
    we obtain similar results.
\end{itemize}
\end{thm}
The \hyperref[proof:heter_dominant_ula]{proof} hinges on the strong convexity of~$Q_\tau$ and presents a stronger statement of the theorem. 

\section{Learning Unit-Level Statistics}
\label{sec:learning}

Welfare is not directly observable. However, noisy observations of welfare may be available. For instance, grades or dropout rates can serve as informative signals of student welfare. We refer to this information as type~$1$, with a cost of~$c_1$ per query per individual. On the other hand, obtaining true welfare is often expensive and may necessitate fresh data. This might involve students taking some standard exams or patients undergoing a set of tests. We call this type of data type~$2$ with a cost of~$c_2$ per data point.
In practice, obtaining true welfare may incur a cost comparable to the cost of intervention~$c$, and we expect $c_2 \approx c \gg c_1$. So, we may collect type~$2$ only from a subset of individuals in select units to train a predictor acting on type~$1$.

In this section, we explore the efficient estimation of unit-level statistics, such as $\rho_k$ or $T'_k$, even when individual welfare estimation poses challenges. We consider general unit-level statistics that can be expressed as the average of a function~$u(\cdot)$ evaluated at $w_i$s. For example, $\rho_k$ is the average of $w \rightarrow \One \{w > 1-\delta\}$ at unit~$k$, or $T'_k$ is the average of $\tau$. For brevity, we discuss the results when $N = \Theta(M)$; however, \cref{sec:learning_appendix} provides the general results.

Starting with the simplest case, consider estimating unit-level statistics by sampling a few individual welfare from each unit. This estimation relies on type~$2$ data and does not involve learning. As discussed in \cref{sec:learning_with_sampling_all}, the relative decrease in ULA's value due to inaccuracy in estimation and the cost of obtaining type~$2$ data asymptotically becomes negligible in expectation when there are many units and the budget is sufficient to effectively treat $K = \Omega(M^{2/3})$ units.

Moving beyond estimating unit-level statistics with sampled welfare values, we aim to learn a predictor that acts on type~$1$ observations. This approach potentially reduces the sampling cost by relying on fewer type~$2$ examples. We first demonstrate that learning unit-level statistics from type~$1$ observations is no harder and often significantly easier than learning individual-level welfare (\cref{sec:learning_is_not_harder}). Formally, suppose there exists a high-probability generalization bound on the mean squared error of the individual welfare predictor in terms of the pseudo-dimension of the hypothesis class. Then, for a monotone concave or convex~$u$, a similarly accurate predictor of unit-level statistics can be obtained with the same or significantly better sample complexity if the predictor is additionally unbiased. Moreover, averaging within units makes the intrinsic unpredictability of individuals less relevant to unit-level statistics.

Next, we focus on methods that directly learn unit-level statistics without relying on an individual predictor. In the simplest model (\cref{sec:learning_noise_free}), we consider noise-free observations $o \in \gO$ as an unknown deterministic function of welfare: $o = f(w)$, assuming $f$ is strictly monotone. Unlike individual-level prediction, which requires learning $f^{-1}$, learning a single parameter $f^{-1}(1 - \delta)$ suffices for simple statistics like~$\rho$ in the context of ULA. \cref{thm:learning_rho_under_noisefree} shows that under \cref{assump:overall_bounded_density} and \ref{assump:gamma_bounded_density}, with an appropriately chosen number of type~$2$ samples and algorithm, the relative decrease of $V_\text{unit}$ due to this extra step of sampling and learning is $O(\sqrt{c/B} \, \ln(1/\alpha))$ with probability at least $1-2\alpha$. This change is negligible since we expect $B \gg c$. A weaker result holds for a general $1$-Lipschitz $u$, implying a negligible decrease in ULA's expected value (\cref{thm:learning_rho_under_noisefree_1_lipschitz_u}).

In a more general setting, observation is a function of welfare and an independent noise term $z$: $o = f(w, z)$. In \cref{sec:learning_noisy}, we discuss learning unit-level statistics under this model with different assumptions on $f$ and noise. Notably, we present a result on the learnability of a ranking of the units under a nonparametric model with minimal assumptions about noise. More precisely, if the noise does not distort the observation too much, meaning there exists a function~$v: \gO \rightarrow \R$ such that $\E_{p_w^k}[u(w)] \approx \E_{p_o^k}[v(o)]$ while $\|v\|_\infty = O(\|u\|_\infty)$, the risk of misranking can be minimized by an SVM-like algorithm minimizing an empirical margin loss (\cref{lem:ranking_generalization_bound}). Hinging on this result, we bound the relative decrease of~$V_\text{unit}$ due to ranking error and cost of learning. Asymptotically, for an appropriately chosen margin in the algorithm, when the number of treated units exceeds $M^{1/2}$, the relative decrease of $V_\text{unit}$ is negligible with high probability (\cref{thm:worst_case_ranking}). Although we perform a worst-case analysis in terms of where the ranking errors might occur, this result is still significant compared to the setting without learning. In sum, learning an accurate ranking of units in terms of their unit-level statistics is possible under minimal assumptions, and the cost of learning can be negligible in the analysis of ULA's value.
\section{Discussion}

Our contribution is two-fold: On the technical side, we provide a mathematical framework for evaluating the efficacy of predictions—the tools presented here point to lines of inquiry to investigate prediction as a solution concept for allocation. 
On the conceptual front, we surface inequality as a driving mechanism limiting the utility of prediction. 
Our insights run counter to the intuition that prediction may curb wasteful allocations in settings where resources are scarce. 

There are several open questions of interest. For instance, in all cases, we only model individual treatment effects. An extension could consider modeling individuals on a network to examine network effects. Likewise, both ULA and ILA treat individuals, and we do not leverage the presence of units beyond identification. A separate line of inquiry may investigate comparing individual versus group interventions. That is, some interventions may only be available at the unit level, e.g., increasing the number of doctors in a hospital or teachers in a school, which would impact the entire unit. 

There are known limitations to allocating by proxy, which individual risk scores often are. These limitations apply whether we are targeting individuals using predictions or unit-level information. Significant gaps exist in our understanding of the limits to prediction as a proxy, and we echo recent calls to examine this concern further. 
\ifarxiv

\fi
Finally, targeting itself is a contested means for intervention, as argued by various social scientists and policy-makers \citep{rawls1971atheory,sen1980equality,sen1981issues,mkandawire2005targeting}. While both ULA and ILA are ultimately used for targeting individuals, the insights in our work point to the persistence of structural forces in determining outcomes. 

\section*{Acknowledgements}
RA was supported by a grant from the Andrew Carnegie Fellowship Program. The authors would like to thank Nathan Srebro, Jon Kleinberg, Ariel Procaccia, Evi Micha, Max Kasy, André Cruz, Ishaq Aden-Ali, and MLO Reading Group at TTIC for extensive feedback and discussions on earlier versions of this work. 

\bibliography{refs}

\newpage
\appendix
\clearpage
\newpage
\section{Additional Related Work}
\label{sec:additional_related}
As a form of resource allocation based on prediction, ILA is widely adopted across various domains. These include allocating homelessness funds to minimize the predicted probability of household re-entry~\citep{kube2023community,kube2023fair} or expected cost~\citep{toros2018prioritizing}, reducing eviction rate by targeting assistance to renters at greater predicted risk~\citep{mashiat2024beyond,abebe2020subsidy}, making ICU discharge decisions based on readmission probability~\citep{chan2012optimizing}, improving emergency responses using incident prediction models~\citep{mukhopadhyay2017prioritized}, or allocating educational resources based on early warning systems that identify students at risk of dropping out~\citep{faria2017getting,mac2019efficacy,dews}. For a more complete discussion on the role of predictive tools in education refer to \citet{liu2023reimagining} and \citet{rismanchian2023four}.

Despite widespread implementation of ILA, in various domains is little evidence of its efficacy~\citep{mac2019efficacy}.  
Reviewing existing algorithmic approaches to reducing homelessness as a resource allocation problem, \citet{moon2024human} found broad issues when optimization is based on individual-level predictors. They suggest a more ``human-centered approach'' that involves key stakeholders to understand how structural and individual factors impact hard-to-quantify facets of an individual, such as one’s resilience and willingness to improve. In our model, this kind of mechanism could be seen as an instance of ULA, where within-unit allocation is delegated to the unit administrators.

\citet{dews} recently demonstrated that early warning systems, when utilized to target interventions at individual students by predicting the risk of dropout, have shown little improvement despite the use of accurate predictors. \citet{dews} and \citet{hardt2023is} also point out how risk scores draw on unit-level information. Consistent observations were made by \citet{faria2017getting} and \citet{mac2019efficacy}, indicating no statistically significant impact of the intervention. In particular, \citet{mac2019efficacy} attribute these observations to the lack of contrast between what happened in treatment and control schools, as the control schools were effectively implementing the same mechanism internally. This supports our assumption that units can implement a reasonable within-unit allocation when directed and provided resources. 
Closely related to our work, \citet{perdomo2023relative} asks how improvements in welfare arising from better predictions compare to those of other policy levers, such as expanding access to resources. In line with our results, it often does not help to invest in greater accuracy when resources are limited.

In policy contexts, historically planning has relied on aggregate data. However, the promise of better resource allocation, lower costs, and more preventative intervention has increasingly motivated individual-level algorithmic solutions in public sector applications~\citep{levy2021algorithms}. \citet{abaigar2023bridging} present an overview of key technical challenges where discrepancies between policy goals and machine learning models commonly arise. In their resource allocation setting, policy goals can be formalized through an allocation principle that chooses the target population based on predicted individual outcomes of interest. As they highlight, such individual estimation may not always be helpful for the decision-making process, and the choice of aggregation level is of utmost importance. 

There is extensive work on the use of causal inference to identify suitable policies and interventions, see, e.g.,~\citep{imbens2009recent, bertsimas2020predictive, athey2021policy} and references therein. \citet{zezulka2024fair} argue in social contexts predictions that inform policies should be evaluated based on the change in social goods that arises after deploying the algorithm.
\citet{kleinberg2015prediction} discuss a class of policy problems that do not necessitate causal inference but rather rely on predictive inference.

Empirical evaluation of \emph{index-based} allocations~\citep{mate2023improved}, where individuals estimated to be in utmost need receive resources, such as the mechanisms we study, requires careful design and estimators~\citep{mate2023improved,boehmer2024evaluating}. 
This is because of the interdependencies among samples, as what an individual receives under these policies depends on her prediction and predictions about others~\citep{zheng2020modeling,shirali2022sequential}.

The extensive debate around the use of risk scores touched on the role of predictive tools as the basis for consequential decisions and interventions. For example, \citet{wang2022against} raise concerns about the legitimacy of decision-making based on predictive optimization.
 In the context of the criminal justice system,
\citet{barabas2018interventions} argued that risk assessment should be intervention-driven and ideally work as a diagnostic tool, identifying causal structures that drive crime. \citet{hofman2021integrating} suggest that explanatory modeling can benefit predictive modeling (and vice versa), for example, by encouraging more robust models that generalize better under interventions. \citet{liu2023reimagining} discuss improving outcome prediction may not necessarily help decision-makers in selecting a more effective intervention option. Related to this topic is the question of actionability of predictions as studied by~\citet{liu2023actionability}. 

Even with access to abundant data on individuals' lives, it remains challenging to forecast outcomes at the individual level~\citep{salganik2020measuring}. In part due to these inherent limits, some literature shows reluctance by individuals to use predictive systems to make inferences about individuals. \citet{raviv2023citizens} found citizens exhibit aversion to the use of algorithmic decision systems when they are required to make inferences about individuals rather than collectives. This might root, for instance, in the disproportionate advantage individuals receive in return for sharing their personal data; in fact, it is challenging to determine whether such participation ultimately benefits or harms them~\citep{monteiro2022on}.

Related to our discussion of intervention-driven predictions is the predict-then-optimize paradigm in the Operations Research community. Here, a predictive tool is first acquired to maximize accuracy, followed by the optimization of allocations or decisions based on it. 
Our definition of ILA closely aligns with this paradigm. The disentanglement of these stages has been a subject of study for a long time; refer to \citet{manski2004statistical} as one of the first attempts where a predictive model is directly optimized to maximize the social welfare resulting from a prediction-based treatment choice. Informing prediction with its downstream applications is widely known as decision-focused learning~\citep{mukhopadhyay2017prioritized,elmachtoub2020decision,elmachtoub2022smart}. Although following such an approach can potentially provide ILA with a predictor informed about the follow-up decision, the main deficiency of ILA in our model comes from the cost of prediction. Our generous assumption of strong per-individual guarantees for the predictor makes a decision-focused predictor less relevant for our study of ILA. 
However, our proposed algorithm for learning unit-level statistics under a nonparametric model in \cref{sec:learning} chooses a sampling strategy informed by downstream ranking decisions and generally operates within a decision-focused framework. Future works can also extend our framework to sequential learning under a resource-constrained setting~\citep{li2022efficient,verma2023restless}. 

Social service bureaucracies neither fully give street-level bureaucrats full discretion nor fully prioritize algorithmically in targeting assistance~\citep{johnson2022bureaucratic,pokharel2023discretionary}. The pre-algorithmic, rule-based part typically involves a dimensionality reduction, quantization, and deliberation by the stakeholders, which overall is called categorical prioritization by \citet{johnson2022bureaucratic}.  \citet{pokharel2023discretionary} discovered that discretionary decisions in resource allocation, not easily explained by simple decision rules, often lead to improved outcomes. This suggests that caseworkers leverage the knowledge gained through human interactions and evaluations in the process. The stronger these discretionary decisions are, the stronger our within-unit allocation, thereby enhancing ULA as well.







\clearpage
\newpage
\section{Additional Tables}

\begin{table*}[h!]
    \centering
    \caption{Sufficient conditions for ULA to dominate ILA. Refer to \cref{thm:dominant_ula_suff_conditions} for accurate statements.\smallskip}

    \begin{tabular}{lcccc}
        & \multirow{2}{*}{$\widehat{G}_\rho > \widehat{G}_\rho^{(2)}$} & \multicolumn{2}{c}{$\widehat{G}_\rho^{(2)} \ge \widehat{G}_\rho > \widehat{G}_\rho^{(1)}$} & \multirow{2}{*}{$\widehat{G}_\rho \le \widehat{G}_\rho^{(1)}$} \\
        \cline{3-4}
        & & $B/C$ meets Eq.~(\ref{eq:dominant_ula_suff_conditions_linear_budget_low_B}) & $B/C$ doesn't meet Eq.~(\ref{eq:dominant_ula_suff_conditions_linear_budget_low_B}) & \\
        \midrule
        $p(\epsilon)/B > q_c$ & $\text{ULA} \succ \text{ILA}$ & $\text{ULA} \succ \text{ILA}$ & $\text{ULA} \succ \text{ILA}$ & $\text{ULA} \succ \text{ILA}$ \\
        \cline{4-5}
        $p(\epsilon)/B > q'/(1-q)$ & $\text{ULA} \succ \text{ILA}$ & $\text{ULA} \succ \text{ILA}$  & \multicolumn{2}{|c|}{\multirow{2}{*}{?}} \\
        o.w. & \multicolumn{2}{c}{$\big(1+O(q')\big)\!\cdot\!\text{ULA} \succeq \text{ILA}$} & \multicolumn{2}{|c|}{} \\
        \cline{4-5}
    \end{tabular}
    \label{tab:dominant_ula_quantitative}
\end{table*}

\begin{table*}[h!]
    \centering
    \caption{Sufficient conditions for when ILA does not dominate ULA. Refer to \cref{thm:nondominated_ula_suff_conditions} for accurate statements.\smallskip}
    \begin{tabular}{lcc}
        & $\widehat{G}_\rho \ge \widehat{G}_\rho^{(0)}$ & $\widehat{G}_\rho < \widehat{G}_\rho^{(0)}$ \\
        \midrule
        $p(\epsilon)/B > \bar{\rho} \, (1 - \widehat{G}_\rho)$ & $\text{ILA} \nsucceq \text{ULA}$ & $\text{ILA} \nsucceq \text{ULA}$ \\
        \cline{3-3}
        $p(\epsilon)/B > q'/(1-q)$ & $\text{ILA} \nsucceq \text{ULA}$ & \multicolumn{1}{|c|}{\multirow{2}{*}{?}} \\
        o.w. & $\text{ILA} \nsucc \big(1+O(q')\big)\!\cdot\!\text{ULA}$ & \multicolumn{1}{|c|}{} \\
        \cline{3-3}
    \end{tabular}
    \label{tab:nondominated_ula_quantitative}
\end{table*}


\section{Additional Figures}

\ifarxiv
\cref{fig:demo_NY} extends our visualization to the case of school districts in New York State. These exhibit a significant level of inequality. As a result, ULA realizes a higher value than ILA.

\begin{figure}[h]
    \centering
    \includegraphics[width=0.9\linewidth]{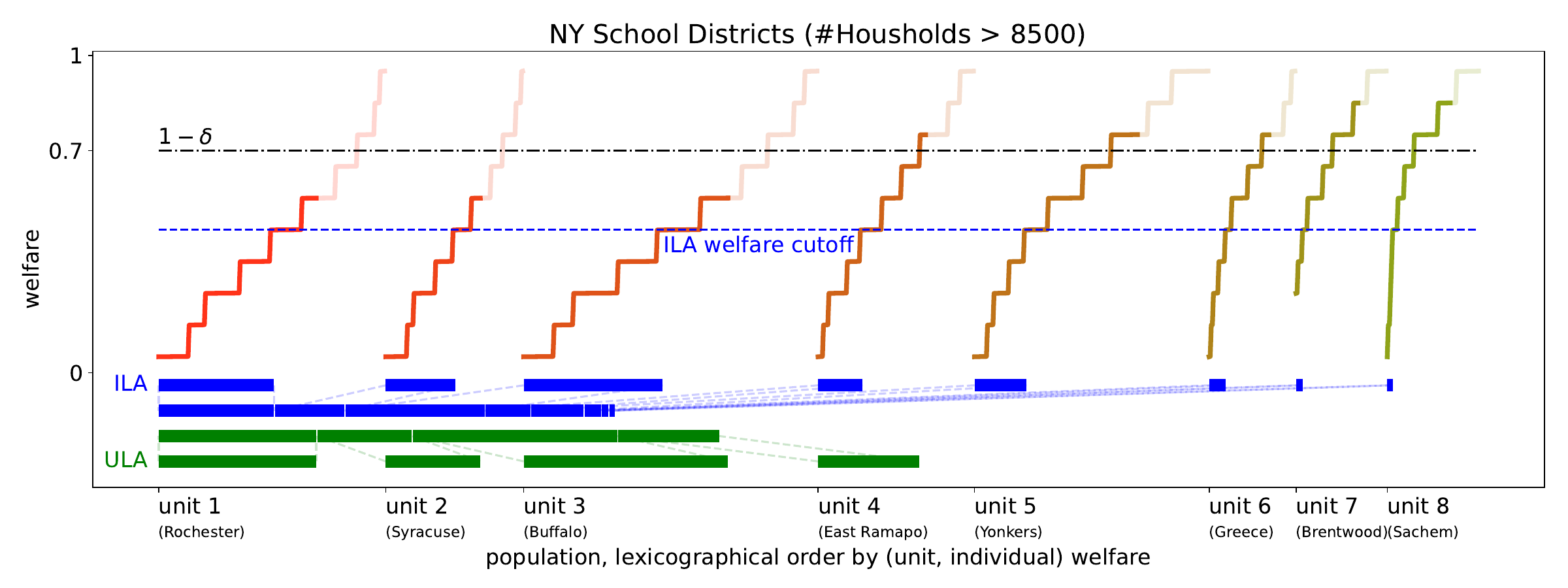}
    \caption{Similar to \cref{fig:acs_ed_demo_LA}, ULA outperforms ILA in a real-world high inequality setting. New York State's school districts with a household population of $8,500$ or more are considered (excluding the New York City Department of Education due to its sheer size). Household income is used as a proxy for individual welfare, categorized into $10$~income brackets. A within-unit allocation with $q=0.3$ and $q'\approx0$ is considered, and individuals among the top~$q$ fraction at each unit appear faded. The values obtained by ILA and ULA from each unit are depicted with horizontal bars, and their overall values are compared.}
    \label{fig:demo_NY}
\end{figure}

\cref{fig:demo_UT} presents an example with school districts from Utah. The school districts have lower inequality compared with those in New York and Los Angeles. As a result, the comparison between ILA and ULA is closer. We see a slight advantage for ILA. However, recall that for the ease of visualization, we assume that ULA gets no value from individuals above the welfare cutoff~$1-\delta.$ In other words, we round any treatment effect strictly smaller than~$\delta$ to zero. If we took such individuals into account, then the green bar corresponding to unit~$3$ would be significantly wider.

\begin{figure}[h]
    \centering
    \includegraphics[width=0.9\linewidth]{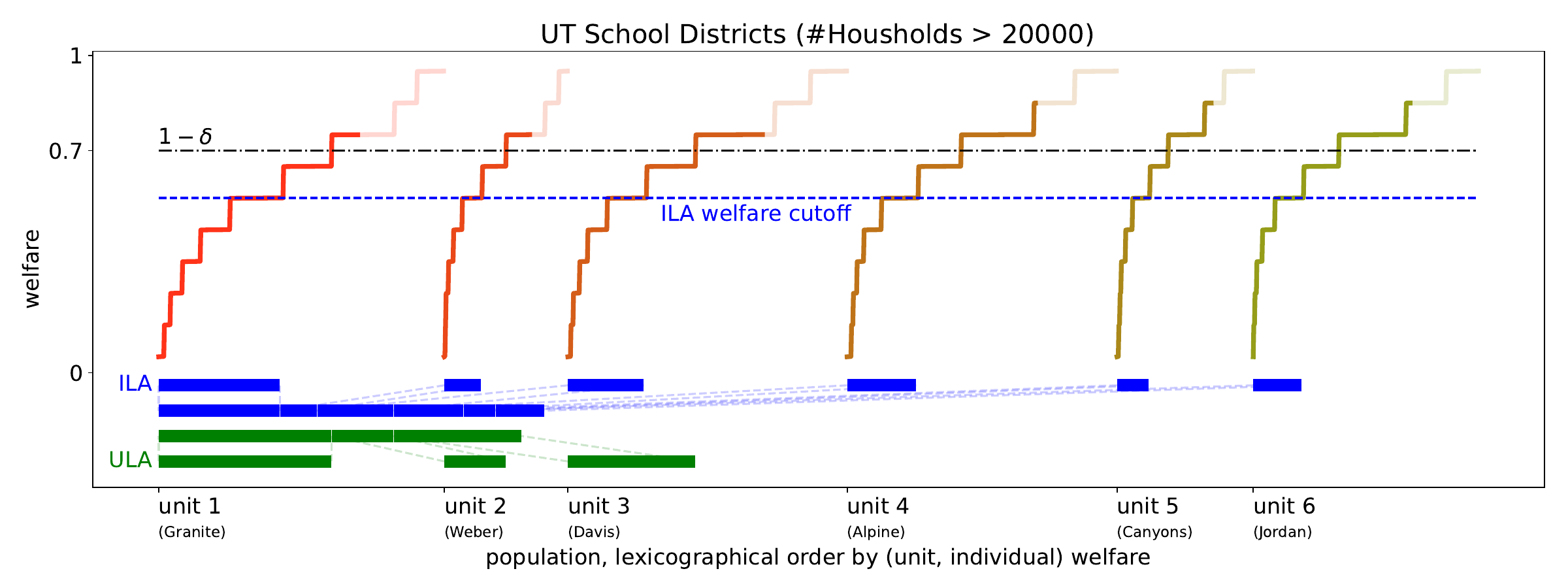}
    \caption{Unlike \cref{fig:acs_ed_demo_LA}, ILA outperforms ULA in a real-world low inequality setting. Utah's school districts with a household population of $20,000$ or more are considered. Household income is used as a proxy for individual welfare, categorized into $10$~income brackets. A within-unit allocation with $q=0.3$ and $q'\approx0$ is considered, and individuals among the top~$q$ fraction at each unit appear faded. The values obtained by ILA and ULA from each unit are depicted with horizontal bars, and their overall values are compared.}
    \label{fig:demo_UT}
\end{figure}
\fi


\clearpage
\newpage
\section{Further on Learning Unit-Level Statistics}
\label{sec:learning_appendix} 

\subsection{Estimating Unit-Level Statistics by Sampling From All Units}
\label{sec:learning_with_sampling_all}
Suppose we estimate unit-level statistics at each unit by sampling the welfare of $n$~individuals at each unit. These samples are classified as type~$2$ because they contain true welfare values. Let $\gS_k$ denote the set of sampled individuals from unit~$k$. Our estimate of the true unit-level statistic~$U_k \coloneqq \frac{1}{N} \sum_{i \in \gU_k} u(w_i)$ will be $\hat{U}_k \coloneqq \frac{1}{n} \sum_{i \in \gS_k} u(w_i)$. Given that the samples are randomly drawn, we can approximate~$\hat{U}_k$ with~$\gN(U_k, \sigma^2/n)$. Since ULA only considers the ranking of these statistics, we can normalize any~$u$ to the interval $[0, 1]$, implying $\sigma^2 \le 1/4$. 

Suppose ULA's value is $(N \delta')$-Lipschitz in unit-level statistics, which holds for $u(w) = \tau(w)/\delta'$ for example. Here, consistent with \cref{sec:het}, $\delta'$ denotes the maximum treatment effect an individual may experience. This property allows us to bound the potential decrease in $V_\text{unit}$ due to inaccurate estimations by
\begin{equation*}
    |\Delta V_\text{unit}| \le N \delta' \sum_{k \in [M]} |\hat{U}_k - U_k|
    \,.
\end{equation*}
Hence, in expectation with respect to the randomness in sampling, we have 
\begin{equation*}
    \E[|\Delta V_\text{unit}|] \le M N \delta' \, \E\big[|\gN(0, 1/(4n))|\big] 
    = 2 M N \delta' \int_0^{\infty} \sqrt{\frac{2 n}{\pi}} \, x \exp\big(-2nx^2\big) \dif x
    = \frac{M N \delta'}{\sqrt{2 \pi n}}
    \,.
\end{equation*}
On the other hand, collecting $n$ type~$2$ samples from each unit reduces the treatment budget by $M \, n \, c_2$. Assuming $c_2 \approx c$, this implies $M n$ fewer individuals will receive treatment. Consequently, we can bound the reduction o~ $V_\text{unit}$ due to estimation cost by~$M \, n \, \delta'$. Therefore, we can bound the overall reduction of ULA's value, accounting for both inaccurate estimation and the additional cost of estimation, by
\begin{equation*}
    \E[|\Delta V_\text{unit}|] \le M n \delta' + \frac{M N \delta'}{\sqrt{2 \pi n}}
    \,.
\end{equation*}
Optimally, choosing $n = \Theta\big(N^{2/3}\big)$ implies $\E[|\Delta V_\text{unit}|] = O\big(M N^{2/3} \delta'\big)$. Since for an efficient allocation we expect $V_\text{unit} = \Theta(K N \delta')$, the relative decrease of ULA's value is bounded by $O\big(\frac{M}{K N^{1/3}}\big)$. For this effect to become negligible asymptotically, the number of treated units should exceed $\Omega(M/N^{1/3})$.


\subsection{Learning Unit-Level Statistics Is No Harder Than Learning Individual Welfare}
\label{sec:learning_is_not_harder}

An individual-level predictor uses a set of features, denoted by~$x$, to make a prediction about welfare~$w$. The features include observations about the individual as well as any relevant contextual information accessible to the predictor. Consider a joint distribution~$\gD$ over $\gX \times \gW$, where $\gX$~is the feature space and $\gW = [0, 1]$. Assume $n_2$~independent samples~$\gS$ are drawn from~$\gD$. These samples go under type~$2$ data because they contain the true welfare values. For a choice of loss function $l: \gW \times \gW \rightarrow [0, 1]$, the risk and the empirical risk of~$h$ are defined as
\begin{align*}
    R(h) &\coloneqq \E_{(x,w)\sim \gD}[l(h(x), w)] \,, \\
    \widehat{R}(h) &\coloneqq \frac{1}{|\gS|} \sum_{(x,w)\in \gS} l(h(x), w) \,.
\end{align*}
To learn a predictor with low risk, the standard approach is to minimize the empirical risk.
If empirical risk minimization can be performed perfectly over a hypothesis class~$\gH$ consisting of functions with range in~$\gW$, then based on a well-known result (see \citet{mohri2018foundations}, Theorem 11.8), the risk of the empirical risk minimizer~$\hat{h}$ can be bounded with probability at least~$1-2\alpha$ by
\begin{equation}
\label{eq:risk_regression}
    R(\hat{h}) \le\min_{h \in \gH} R(h) + \sqrt{\frac{8d \log(e n_2 /d)}{n_2}} + \sqrt{\frac{2 \log(1/\alpha)}{n_2}} = \min_{h \in \gH} R(h) + \widetilde{O}\Big(\sqrt{\frac{d}{n_2}} + \sqrt{\frac{\log(1/\alpha)}{n_2}}\Big)
    \,.
\end{equation}
Here $d$~is the pseudo-dimension (\citet{mohri2018foundations}, Definition 11.5) of $\gG = \{(x, w) \rightarrow l(h(x), w): h \in \gH\}$. Let $l(\hat{w}, w) = (\hat{w} - w)^2$. Collecting $n_2 = \Omega(d/\epsilon^2)$ samples ensures the second term of the bounded risk will be $\widetilde{O}(\epsilon)$. Then $\hat{h}$ could have a mean squared error of~$\widetilde{O}(\epsilon)$ if $\min_h R(h)$ is sufficiently small. The best we can hope for this term is the Bayes risk from the Bayes predictor~$h_B(x) = \E[w | x]$. Unfortunately, in many practical settings, limited relevant observations from individuals and their intrinsic unpredictability can lead to a large Bayes risk, making accurate prediction challenging. Next, we discuss why this concern is less relevant for estimating unit-level statistics.

Consider a unit-level statistic that can be expressed as the average of a strictly monotone bounded function $u(\cdot)$ evaluated at $w_i$s. Note that since only the ranking of such statistics matters in ULA, we can normalize any~$u$ to be in $[0, 1]$. Redefine the loss as $l(\hat{w}, w) \coloneqq \big(u(\hat{w}) - u(w)\big)^2$. For strictly monotone convex or concave~$u$, one can verify that the pseudo-dimension based on the new~$l$ is similar to the pseudo-dimension based on the~$(\hat{w} - w)^2$ loss. We may assume that minimizing the empirical risk based on the new loss function is computationally as hard as minimizing the empirical risk based on the original $(\hat{w} - w)^2$~loss. Given this assumption, a similar argument demonstrates that the risk of the empirical risk minimizer~$\hat{h}$ can be bounded by \cref{eq:risk_regression}. Using~$\hat{h}$, we can estimate the unit-level statistic at unit~$k$ by calculating $\frac{1}{N} \sum_{i \in \gU_k} u(\hat{h}(x_i))$. Our primary interest lies in the~$L_2$ risk of the estimated unit-level statistics. Using the Jensen's inequality, we can bound this with the risk of~$\hat{h}$:
\begin{equation}
\label{eq:leanring_is_not_harder}
    \E\Big[\frac{1}{M} \sum_{k \in [M]} \big(\frac{1}{N}\sum_{i \in \gU_k} u(\hat{h}(x_i)) - \frac{1}{N}\sum_{i \in \gU_k} u(w_i) \big)^2\Big]
    \le \E\Big[ \frac{1}{M N} \sum_{k \in [M]} \sum_{i \in \gU_k} \big(u(\hat{h}(x_i)) - u(w_i)\big)^2 \Big] = R(\hat{h})
    \,.
\end{equation}
This shows learning unit-level statistics is no harder than learning individual-level welfare. 

Suppose $\hat{h}$ is additionally unbiased: $\E[u(\hat{h}(x)) \mid w] = u(w)$. Then a straightforward calculation shows that the upper bound of \cref{eq:leanring_is_not_harder} is loose and the $L_2$ risk of the estimated unit-level statistics can be bounded by~$\frac{1}{N} R(\hat{h})$.
Moreover, $R(\hat{h}) \le R(h_B) + \widetilde{O}\big(\sqrt{d/n_2}\big)$. 
Note that the Bayes predictor in this case outputs $h_B(x) = u^{-1}\big(\E[u(w) | x]\big)$. One can see for a large number of individuals per unit the Bayes risk is less relevant here. When this is the case, $n_2 = \Omega(d/(N \epsilon)^2)$ samples are sufficient to get a mean-squared error of $\widetilde{O}(\epsilon)$ which is significantly less number of samples compared to learning an individual-level predictor with a similar error. 

In sum, assuming the efficient learnability of individual welfare, learning unit-level statistics can be achieved using similar tools as for predicting individual welfare but in a more efficient way possibly requiring fewer assumptions about the intrinsic unpredictability of individuals. In the following, we show learning unit-level statistics can be done directly based on type~$1$ observations and without an individual welfare predictor under the hood. 

\subsection{Learning Unit-Level Statistics Under Noise-Free Observations $o = f(w)$}
\label{sec:learning_noise_free}

In the simplest case, observation~$o$ is an unknown deterministic function of welfare~$w$. We assume $f$ is a strictly increasing function of~$w$, however, similar results hold true for a strictly decreasing function. Learning a function from observation to individual welfare requires learning~$f^{-1}$. However, this is unnecessary for estimating unit-level statistics. For instance, $\rho_k$ is an average of threshold function $w \rightarrow \One\{w > 1-\delta\}$ evaluated at~$w_i$s. Since $f$ is an increasing function, there exists a threshold~$t$ such that $\rho_k$ can be calculated as an average of threshold function $o \rightarrow \One\{o > t\}$ evaluated at~$o_i$s. Therefore, the problem of learning~$\rho_k$s reduces to learning a single parameter~$t$. The next theorem demonstrates that for every available budget~$B$, by appropriately choosing the number of type~$2$ samples, we can estimate $\rho_k$s sufficiently accurately with minimal cost, such that our previous bounds on~$V_\text{unit}$ can be nearly realized with high probability.

\begin{theoremEnd}{thm}
\label{thm:learning_rho_under_noisefree}
Given a set of welfare-observation pairs $\{w_i, o_i\}$, choose threshold~$\hat{t}$ as the largest~$o_i$ for which $w_i \le 1-\delta$. Estimate~$\rho_k$ at each unit as the average of~$\One\{o > \hat{t}\}$. Consider ULA based on estimated $\rho_k$s, given a budget of~$B \gg c$ for treating $K$~units. Under \cref{assump:overall_bounded_density} and \ref{assump:gamma_bounded_density}, for any $0 < \alpha < 1/2$, by choosing an appropriate number of samples, the decrease of~$V_{\rm unit}$ due to the extra step of sampling and learning is bounded with probability at least~$1-2\alpha$ by
\begin{equation*}
    \Delta V_{\rm unit} \ge -\sqrt{K N} \delta \Big(\gamma/\ugamma + \ln(1/\alpha) + z_\alpha (K N)^{-1/4} (\gamma/\ugamma)^{1/2} \Big)
    \,,
\end{equation*}
where $z_\alpha$ is the z-score at $\alpha$~significance level. This implies that, with probability at least $1-2\alpha$, the relative decrease of~$V_{\rm unit}$ is bounded by
\begin{equation*}
    O\Big(\sqrt{\frac{c}{B}} \big(\gamma/\ugamma + \ln(1/\alpha) \big) \Big)
    \,.
\end{equation*}
\end{theoremEnd}
\begin{proofEnd}
    Choose a set~$\gS$ of $n_2$~individuals uniformly at random and collect their true welfare data (type~$2$ data). Let $\hat{t} = \max \,\{o_i: i \in \gS, w_i \le 1-\delta\}$ be the estimated threshold. Note that $\hat{t} \le t$. At every unit~$k$, estimate $\rho_k$ from observations by $\hat{\rho}_k = \frac{1}{N} \sum_{i \in \gU_k} \One\{o_i > \hat{t}\}$. The error of estimating~$\rho_k$ is
    \begin{equation*}
        \hat{\rho}_k - \rho_k = \frac{1}{N} \sum_{i \in \gU_k} \One\{o_i \in (\hat{t}, t]\} \ge 0
        \,.
    \end{equation*}

    Sort the $\rho_k$~values in ascending order and, without loss of generality, let $\rho_k$ be the $k^\text{th}$ smallest value. Similarly, sort the $\hat{\rho}_k$ values in ascending order and let $\hat{s}(k)$ denote the unit with the $k^\text{th}$ smallest value. In ULA, we intervene on first~$K$ units according to $\hat{s}(\cdot)$. For $k \in [K]$, if $k$ is among $\hat{s}(1), \dots, \hat{s}(K)$, then our previous bound on loss from allocation to this unit will be valid. Otherwise, a unit not among true bottom~$K$ units has substituted unit~$k$ in allocation. In this case, for any $k' \in [K]$, we have $\hat{\rho}_{\hat{s}(k')} \le \hat{\rho}_{k}$ and hence
    \begin{equation*}
        \rho_{\hat{s}(k')} \le \hat{\rho}_{\hat{s}(k')} \le \hat{\rho}_k \le \rho_k + \frac{1}{N} \sum_{i \in \gU_k} \One\{o_i \in (\hat{t}, t]\}
        \,.
    \end{equation*}
    So, in the worst case, for $k \in [K]$, units with effective unit-level statistic~$\tilde{\rho}_k = \rho_k + \frac{1}{N} \sum_{i \in \gU_k} \One\{o_i \in (\hat{t}, t]\}$ will be treated. Recall that $\text{loss}(\cdot)$ in \cref{eq:loss_definition} is $1$-Lipschitz. This suggests that for a fixed~$\hat{t}$, the increase of loss can be bounded by
    \begin{equation*}
        \Delta \text{loss} \le \frac{1}{N} \sum_{k \in [K]} \sum_{i \in \gU_k} \One\{o_i \in (\hat{t}, t]\}
        \,.
    \end{equation*}

    Define $D \coloneqq f^{-1}(t) - f^{-1}(\hat{t})$. The fact that $f$~is an increasing function and $\hat{t} \le t$ ensure $D \ge 0$. \cref{assump:gamma_bounded_density} limits the concentration of welfare at each unit by~$\gamma$. Therefore, in the worst case, we can think of $\One\{o_i \in (\hat{t}, t]\}$ as $\text{Ber}(\gamma D)$. Denoting the z-score at the $\alpha$ significance level by $z_\alpha$ and using the central limit theorem to approximate the sum of Bernoulli random variables with a Gaussian distribution, we have
    \begin{equation*}
        \Delta \text{loss}_K \le \frac{1}{N} \big( K N \gamma D + z_\alpha \sqrt{K N} \sqrt{\gamma D (1 - \gamma D)} \big) \le K \gamma D + z_\alpha \sqrt{K/N} \sqrt{\gamma D}
        \,,
    \end{equation*}
    with probability at least $1-\alpha$. Observing $D > d$ requires none of $n_2$~welfare samples to lie in an interval of length~$d$. Since the overall welfare density is lower bounded by~$\ugamma$ (\cref{assump:overall_bounded_density}), we have $\Pr(D > d) \le (1 - \ugamma d)^{n_2}$. Then a direct calculation shows
    \begin{equation*}
        D \le (1 - \alpha^{1/n_2})/\ugamma
        \,,
    \end{equation*}
    with probability at least~$1-\alpha$. Putting these together, the increase in loss due to inaccurate estimation of unit-level statistics is bounded by
    \begin{equation*}
        \Delta \text{loss}_K \le K \frac{\gamma}{\ugamma} (1 - \alpha^{1/n_2}) + z_\alpha \sqrt{K/N} \sqrt{\frac{\gamma}{\ugamma}} \sqrt{1 - \alpha^{1/n_2}}
        \,,
    \end{equation*}
    with probability at least $1-2\alpha$.
    Assuming $c_2 \approx c$, the cost $n_2$ type~$2$ samples at most reduces ULA's value by $n_2 \delta$. A unit change in $\text{loss}_K$ would also decrease ULA's value by $N \delta$. Therefore, with probability at least $1-2\alpha$, the decrease of ULA's value can be bounded by
    \begin{equation*}
        \Delta V_\text{unit}/\delta \ge -K N \frac{\gamma}{\ugamma} (1 - \alpha^{1/n_2}) - z_\alpha \sqrt{K N} \sqrt{\frac{\gamma}{\ugamma}} \sqrt{1 - \alpha^{1/n_2}} - n_2
        \,.
    \end{equation*}
    Choose $n_2 = \sqrt{K N} \, \ln(1/\alpha)$. Then $\alpha^{1/n_2} = \exp(\ln(\alpha)/n_2) = \exp(-1/\sqrt{K N}) \approx 1 - 1/\sqrt{K N}$, and we have
    \begin{equation*}
        \Delta V_\text{unit}/\delta \ge -\sqrt{K N} \big(\frac{\gamma}{\ugamma}\big) - z_\alpha (K N)^{1/4} \sqrt{\frac{\gamma}{\ugamma}} - \sqrt{K N} \, \ln(1/\alpha)
        \,.
    \end{equation*}
    Using $V_\text{unit} = \Theta(K N \delta)$ and $K = \Theta(B/(N c))$ complete the proof.
\end{proofEnd}

\cref{thm:learning_rho_under_noisefree} suggests learning~$\rho_k$s under noise-free observations comes at the relative cost of $O(\sqrt{c/B})$ which we expect to be very small. In fact, this is not limited to $\rho_k$s. The next theorem shows that for a $1$-Lipschitz~$u$, a slightly weaker bound of $\widetilde{O}\big(\sqrt{c/B}\big)$ on the relative decrease of ULA's value holds in expectation.

\begin{theoremEnd}{thm}
\label{thm:learning_rho_under_noisefree_1_lipschitz_u}
Consider ULA based on unit-level statistics defined as the average of a $1$-Lipschitz function~$u(\cdot)$ at each unit. Given a random set~$\gS$ of welfare-observation pairs, we can reconstruct a quantized~$f^{-1}$ and use that to estimate unit-level statistics based on type~$1$ observations (the details come in the proof). Assume $V_{\rm unit}$ is $N$-Lipschitz in the unit-level statistics (for example, this is the case when unit-level statistics are average treatment effects). Given a budget of~$B \gg c$ for treating $K$~units, under \cref{assump:overall_bounded_density}, by an appropriately chosen number of samples, the expected decrease of~$V_{\rm unit}$ due to the extra step of sampling and learning is bounded by
\begin{equation*}
    \E_\gS[\Delta V_\text{unit}] \ge -3\sqrt{KN} - \frac{\delta'}{2\ugamma} \sqrt{KN} \ln(KN)
    \,,
\end{equation*}
where $\delta'$ is the maximum treatment effect an individual may experience. This implies that, the expected relative decrease of~$V_{\rm unit}$ is bounded by
\begin{equation*}
    \widetilde{O}\Big(\sqrt{\frac{c}{B}} \big(1/\ugamma + 1/\delta'\big) \Big)
    \,.
\end{equation*}
\end{theoremEnd}
\begin{proofEnd}
    Our strategy is to reconstruct~$f$ by quantizing welfare into $L$~levels. Choose a set~$\gS$ of $n_2$~individuals uniformly at random and get their true welfare (type~$2$ data). Define
    \begin{equation*}
        \hat{t}_l = \begin{cases}
            \max \, \{o_i: i \in \gS, w_i \le l/L\} \, , & l \in [L-1]\,, \\
            1 \,, & l=L\,.
        \end{cases}
    \end{equation*}
    Here, if a set was empty for $l^\text{th}$ bin, we can assume $\hat{t}_l = \hat{t}_{l+1}$. Using the estimated thresholds, we can approximate $f^{-1}$ and estimate the welfare of an observation~$o$ by
    \begin{equation*}
        \hat{w} = \hat{f}^{-1}(o) = \frac{1}{L} \min \, \{l \in [L]: \hat{t}_l \ge o\}
        \,.
    \end{equation*}
    Note that by this definition, we have $\hat{w} \ge w$.
    Consider a welfare value~$w$ that lies in $[(l-1)/L, l/L]$ for an $l \in [L-1]$. If there exists $i \in \gS$ such that $w_i \in [l/L, (l+1)/L]$, then $\hat{w} - w \le 2/L$. Since $\gS$ are random draws and welfare density is at least~$\ugamma$ (\cref{assump:overall_bounded_density}), 
    \begin{equation*}
        \Pr_\gS(\hat{w} - w > 2/L) \le \beta \coloneqq (1 - \ugamma/L)^{n_2}
        \,.
    \end{equation*}
    For $l=L$, since $\hat{t}_L = 1$, we always have $\hat{w} - w \le 1/L$, and the above argument holds more strongly. Denote the unit-level statistics at unit~$k$ by~$U_k$: $U_k = \frac{1}{N} \sum_{i \in \gU_k} u(w_i)$. Our estimation from~$U_k$ is $\hat{U}_k = \frac{1}{N} \sum_{i \in \gU_k} u(\hat{w}_i)$, where $\hat{w}_i = \hat{f}^{-1}(o_i)$. For a $1$-Lipschitz~$u$, we have
    \begin{equation*}
        (1-\beta) (u(w_i) - 2/L) \le \E_\gS[u(\hat{w}_i)] \le (1-\beta) (u(w_i) + 2/L) + \beta
        \,.
    \end{equation*}
    Therefore, $|\E_\gS[u(\hat{w}_i)] - u(w_i)| \le \beta + 2/L$. This gives
    \begin{equation*}
        |\E_S[\hat{U}_k] - U_k| \le \beta + 2/L
        \,.
    \end{equation*}
    Now if $V_\text{unit}$ is $N$-Lipschitz in~$U_k$ for any~$k\in[K]$, the expected decrease of~$V_\text{unit}$ can be bounded by
    \begin{equation*}
        \E_\gS[\Delta V_\text{unit}] \ge -K N (\beta + 2/L) - n_2 \delta' = -K N \big((1 - \ugamma/L)^{n_2} + 2/L\big) - n_2 \delta'
        \,,
    \end{equation*}
    where $\delta'$ is the maximum treatment effect an individual can experience. Using the well-known property $\exp(x) \ge (1+x/n)^n$ for $n \ge |x|$, by choosing $n_2 = (L/\ugamma) \cdot \ln(L)$ and $L = \sqrt{K N}$, we obtain
    \begin{equation*}
        \E_\gS[\Delta V_\text{unit}] \ge -K N \big(1/L + 2/L\big) - (L/\ugamma) \ln(L) \, \delta' = -3\sqrt{KN} - \frac{\delta'}{2\ugamma} \sqrt{KN} \ln(KN)
        \,.
    \end{equation*}
    Using $V_\text{unit} = \Theta(K N \delta')$ and $K = \Theta(B/(N c))$ complete the proof.           
\end{proofEnd}


\subsection{Learning Unit-Level Statistics Under Noisy Observations $o = f(w, z)$}
\label{sec:learning_noisy}

\paragraph{Estimation Under Known Welfare-To-Observation Likelihood.}
In the simplest model, suppose the function~$f$ and the distribution of noise~$z$ denoted by~$p_z$ are known. Also, suppose welfare distribution can be represented as a member of a parametric statistical model parameterized by a few-dimensional parameter~$\theta$. Marginalizing over~$z$, we obtain a statistical model with parameter~$\theta$ over the space of observation. Maximizing likelihood, under mild assumptions, we can obtain a consistent estimator~$\hat{\theta}_k$ for unit~$k$ with true parameter~$\theta_k$. Assuming there are $n_1$~observations available from this unit, $\hat{\theta}_k$ will have a covariance matrix of~$\frac{I^{-1}(\theta_k)}{n}$, where $I(\theta_k)$ is the Fisher information matrix. We can then analytically calculate expected $T'_k$ or $\rho_k$ for this unit. For instance, denoting the cumulative density function of welfare associated with~$\theta$ as $P_\theta$, the estimated~$\hat{\rho}_k$ will be consistent with a variance of $\nabla_\theta P_\theta(1-\delta)^\sT \, \frac{I^{-1}(\theta_k)}{n} \,\nabla_\theta P_\theta(1-\delta)$. For a regular statistical model, we can assume $I(\theta_k)$ is well-conditioned (its smallest eigenvalue is $\Omega(1)$) and $P_\theta$ is $O(1)$-Lipschitz. These are sufficient conditions to guarantee $\|I(\theta_k)^{-1/2} \nabla_\theta P_\theta\|^2_2 = O(1)$. Then,  $\big|\E_w[\hat{\rho}_k] - \E_w[\rho_k]\big| = O\big(\frac{1}{\sqrt{n_1}}\big)$ with high probability. In fact, for each individual, there may already be a few type~$1$ samples almost freely available. Then, when $n_1 = \Omega(N)$, the $O\big(\frac{1}{\sqrt{n_1}}\big)$ error in estimating $\E_w[\rho_k]$ will be smaller or on par with the unavoidable $O\big(\frac{1}{\sqrt{N}}\big)$ finite sample error in $\big|\rho_k - \E_w[\rho_k] \big|$, making $\hat{\rho}_k$ optimal.

In summary, when the welfare-to-observation likelihood is known and the Fisher information of the specified statistical model is well-conditioned, no type~$2$ samples or learning is required to estimate unit-level statistics. It is worth noting that a known likelihood does not necessarily imply that observations can fully resolve uncertainty about individual welfare. In fact, for each individual, only a few observations may be relevant at the time of intervention. For instance, a student might have only a few grades early in their school when interventions are most effective. Furthermore, individual welfare is subject to change, and previous years' grades may not be helpful due to potential changes in the student's life circumstances. However, these issues are less relevant when considering units as a whole.

\paragraph{Learning Under Limited Distortion of Welfare Distribution.}
The following proposition provides an example where noisy observations of welfare are available with an unknown noise distribution, but the noise does not strongly distort the observations. In this case, by paying a fixed cost, we can obtain an $\epsilon$-accurate predictor for unit-level statistics.

\begin{proposition}
Consider noisy observation $o = f(w, z)$ where $z$~is an independent randomness (noise). Assume there exists a scaled and shifted version of observation~$\tilde{o}$ such that ${\rm EMD}(p_w, p_{\tilde{o}}) \le \epsilon$, where ${\rm EMD}$ is earth mover's distance. Suppose $u(\cdot)$ is $1$-Lipschitz and there are many type~$1$ observations available from each unit for free. By observing average treatment effects from $1/\epsilon^2$ units, we can obtain a predictor from observations to unit-level statistic that has an average absolute error of $O(\epsilon)$. 
\end{proposition}
\begin{proof}
    For a $1$-Lipschitz~$u(\cdot)$, Kantorovich-Rubinstein duality implies $\big| \E_w[u(w)] - \E_{\tilde{o}}[u(\tilde{o})] \big| \le \text{EMD}(p_w, p_{\tilde{o}}) \le \epsilon$. Therefore, when the input to~$u$ is appropriately shifted and scaled, the estimated unit-level statistic $\hat{U}_k \coloneqq \E_{\tilde{o}}[u(\tilde{o}) \mid \gU_k]$ will remain in the $\epsilon$-radius of the true statistic~$U_k \coloneqq \E_w[u(w) \mid \gU_k]$. We can find the scale and shift by solving a linear regression: Assume $U_k$s are available from $m$~units denoted by~$\gS$. Solve the regression to find the scale and shift that minimizes $\sum_{k \in \gS} |\hat{U}_k - U_k|$. Such estimates will have an $L_1$-norm risk of~$O(\epsilon + \sqrt{1/m})$. Choosing $m = 1/\epsilon^2$ will complete the proof. 
\end{proof}


\paragraph{Learning Under a Nonparamteric Model.}

Next, we show efficient learning of unit-level statistics is possible under more general nonparametric models. Assume that observation~$o \in \gO$ is an unknown noisy function of~$w$: $o = f(w, z)$. Units differ in how welfare is distributed within each unit, but the function~$f$ and the distribution of noise~$z$ are common across units. We denote the density of welfare and observation at unit~$k$ by~$p_w^k$ and~$p_o^k$, respectively. For demonstration purposes, we assume that $\gO$ is a closed interval of~$\R$ with~$O(1)$ length. 

We assume the distortion from the noise is limited in the sense that there exists~$v: \gO \rightarrow \R$ such that for every unit~$k$ we have $\E_{p_w^k}[u(w)] \approx \E_{p_o^k}[v(o)]$ while $\max_o |v(o)| = O(\max_w |u(w)|)$.
Consider a Hilbert space~$\sH = \R^\gO$ equipped with an inner product~$\langle p, q \rangle_\sH = \int_\gO p(o) \, q(o) \dif o$, for $p, q \in \sH$. Define a kernel~$\gK: [M]\times[M] \rightarrow \R$ by its feature mapping~$\phi: [M] \rightarrow \sH$ where $\phi(k) = p_o^k$. Hence $\gK(k, k') = \int_\gO p_o^k(o) \, p_o^{k'}(o) \dif o$.

Since only the ranking of statistics matters in ULA, we can normalize any~$u$ to be in $[0, 1]$. Our assumptions on the observation model ensure that there exists~$v^* \in \sH$ with $\|v^*\|_\infty = \Lambda = O(1)$ such that the unit-level statistic at unit~$k$ can be computed by $h^*(k) = \langle v^*, \phi(k) \rangle_\sH = \E_{p_w^k}[u(w)]$. Note that $h^*$ is linear in~$v^*$. Consider hypothesis class
\begin{equation}
\label{eq:def_lin_hypothesis_class}
    \gH = \{h: k \rightarrow \langle v_h, \phi(k) \rangle_\sH, \; \|v_h^2\|_\sH \le \Lambda \}
    \,.
\end{equation}
The next assumption enables efficient learning of~$v_h$ such that $h(k)$ induces an accurate ranking of units, as presented in a follow-up lemma.

\begin{assumption}
\label{assump:kernel_r}
We limit the concentration of~$p_o^k$ by assuming $r \coloneqq \sup_{k \in [M]} \gK(k, k)$ is $O(1)$.
\end{assumption}

\begin{lemma}[\citet{mohri2018foundations}, Corollary 10.2]
\label{lem:ranking_mohri_generalization_bound}
Consider $n$~independent samples~$\gS$ drawn from an arbitrary joint distribution~$\gD$ over~$[M]\times[M]$. For $\beta > 0$ and a sample~$(k, k', y)$, where $y = \chi\{h^*(k) > h^*(k')\}$, define the $\beta$-margin loss $l_\beta\big(y(h(k) - h(k'))\big) \coloneqq \min \, \big\{1, \max \, \{0, 1 - y(h(k) - h(k'))/\beta\} \big\}$. Also, define the zero-one risk and empirical $\beta$-margin risk by
\begin{align*}
    R(h) \coloneqq \E_{(k, k', y) \sim \gD}\big[l_{0^+}\big(y(h(k) - h(k'))\big)\big] \,,\\
    \hat{R}_\beta(h) \coloneqq \frac{1}{n} \sum_{(k, k', y) \in \gS} l_\beta \big(y(h(k) - h(k'))\big) \,.
\end{align*}
Under \cref{assump:kernel_r}, with probability at least~$1-\alpha$, for every~$h$ in the hypothesis class~$\gH$ of \cref{eq:def_lin_hypothesis_class}, we have
\begin{equation*}
    R(h) \le \hat{R}_\beta(h) + \frac{4r}{\beta} \sqrt{\frac{1}{n}} + \sqrt{\frac{\log(1/\alpha)}{n}}
    \,,
\end{equation*}
where $r = O(1)$ by assumption.
\end{lemma}

Direct application of this lemma is not possible since we only have finite-sample estimates of~$h^*(k)$ and~$h(k)$. Let $z_\alpha$ be the z-score of $\alpha$ significance level. If there are $n_1$~type~$1$ samples available for calculating~$h(k)$, then $\hat{h}(k) = \frac{1}{n_1} \sum_{j, o^{(j)} \sim p_o^k} v_h(o^{(j)})$ will be in $z_\alpha \sqrt{\Var(v_h)/n_1}$ radius of~$h(k)$ with probability at least~$1-\alpha$. Observing $n_2$~type~$2$ samples from each unit, we can estimate~$h^*(k)$ with $\tilde{h}^*(k) = \frac{1}{n_2} \sum_{j, w^{(j)} \sim p_w^k} u(w^{(j)})$ up to an error of $z_\alpha \sqrt{\Var(u)/n_2} \le z_\alpha \sqrt{1/(4n_2)}$ with probability at least~$1-\alpha$. Using these finite-sample estimates, we can present the following lemma.

\begin{lemma}[Extension of \cref{lem:ranking_mohri_generalization_bound}]
\label{lem:ranking_generalization_bound}
Given $\beta$, $n_2$, and $\alpha$, consider $n$~independent samples drawn from an arbitrary joint distribution~$\gD$ over~$([M]\times[M]) \setminus \big\{(k, k'): |h^*(k) - h^*(k')| < \max \, \{\beta,  z_\alpha / \sqrt{n_2}\} \big\}$. Let $h$ be the minimizer of empirical $\beta$-margin risk calculated with $n_1$~type~$1$ and $n_2$~type~$2$ samples. Under \cref{assump:kernel_r}, with probability at least~$1-8\alpha$, we have
\begin{equation*}
    R(h) \le \frac{4 z_\alpha}{\beta} \sqrt{\frac{\sqrt{r}\Lambda}{n_1}} + \frac{4r}{\beta} \sqrt{\frac{1}{n}} + \sqrt{\frac{\log(1/\alpha)}{n}}
    \,.
\end{equation*}
\end{lemma}
\begin{proof}
    Consider a sample $(k, k')$ from~$\gD$. Our finite-sample evaluation of this sample's label based on type~$2$ data is $\tilde{y} = \chi\{\tilde{h}^*(k) > \tilde{h}^*(k')\}$ where $\tilde{h}^*(k) = \frac{1}{n_2} \sum_{j, w^{(j)} \sim p_w^k} u(w^{(j)})$. Since $\gD$ only encompasses $(k, k')$ such that $|h^*(k) - h^*(k')| > z_\alpha / \sqrt{n_2} $ and $|\tilde{h}^*(k) - h^*(k)| \ge z_\alpha \sqrt{1/4n_2}$ with probability at least~$1-\alpha$, we can conclude $\tilde{y} = y$ with probability at least $1-2\alpha$.
    
    Let $\hat{h}^*(k) = \frac{1}{n_1} \sum_{j, o^{(j)} \sim p_o^k} v^*(o^{(j)})$ be the empirical evaluation of~$h^*(k)$ based on type~$1$ observations. For $(k, k')$ drawn from~$\gD$ we know $|h^*(k) - h^*(k')| \ge \beta$. Since $|\hat{h}^*(k) - h^*(k)| \le z_\alpha \sqrt{\Var(v^*)/n_1}$ with probability at least~$1-\alpha$, we can argue $l_\beta\big(y(\hat{h}^*(k) - \hat{h}^*(k'))\big) \le \frac{2 z_\alpha}{\beta} \sqrt{\Var(v^*)/n_1}$ with probability at least $1 - 2\alpha$. 
    
    Putting it together, the finite-sample evaluation of empirical margin risk of $h^*$ will be bounded by $\frac{2 z_\alpha}{\beta} \sqrt{\Var(v^*)/n_1}$ with probability at least $1-4\alpha$. Let $h$ be the minimizer of the finite-sample evaluation of empirical margin risk. A similar argument shows the true empirical margin risk of~$h$, denoted by $\hat{R}_\beta(h)$, cannot differ from its finite-sample evaluation more than $\frac{2 z_\alpha}{\beta} \sqrt{\Var(v_h)/n_1}$ with probability at least~$1-4\alpha$. Therefore, since $h$ was the minimizer of the finite-sample evaluation of empirical margin risk, we have
    \begin{equation}
    \label{eq:_proof_ranking_generalization_bound}
        \hat{R}_\beta(h) \le \frac{2 z_\alpha}{\beta} \sqrt{\frac{\Var(v^*)}{n_1}} + \frac{2 z_\alpha}{\beta} \sqrt{\frac{\Var(v_h)}{n_1}}
        \,,
    \end{equation}
    with probability at least~$1-8\alpha$. By definition of the hypothesis class, $\|v_h^2\|_\sH \le \Lambda$. Then for any unit~$k$, \cref{assump:kernel_r} and Cauchy–Schwarz inequality imply
    \begin{equation*}
        \Var(v_h) \le \max_k \, \E_{p_o^k}[v_h^2(o)] = \max_k \, \langle \phi(k),  v_h^2\rangle_\sH \le \max_k \, \|\phi(k)\|_\sH \, \|v_h^2\|_\sH \le \sqrt{r} \Lambda
        \,.
    \end{equation*}
    Since $h^* \in \gH$, the same bound holds on $\Var(v^*)$. Using these variance bounds to further bound~$\hat{R}_\beta(h)$ in \cref{eq:_proof_ranking_generalization_bound}, and plugging this into \cref{lem:ranking_mohri_generalization_bound} complete the proof. 
\end{proof}

Thus far, we have shown that efficiently ranking units is possible for a general bounded $u(\cdot)$. The following assumption will enable us to connect the ranking risk with the ULA's value in subsequent analysis.

\begin{assumption}
\label{assump:statistics_not_concentrated}
Denote the unit-level statistics at unit~$k$ by~$U_k$.
We assume $U_k$s are not very concentrated: There exists $\lambda = O(1)$ such that $\Pr(U_k \in [a, a + b]) \le \lambda \,b$. 
\end{assumption}

Since type~$1$ data is abundant and almost freely available, we neglect finite-sample issues with~$n_1$ and assume~$c_1 \approx 0$. The next theorem shows that by appropriately choosing the number of samples and margin~$\beta$, the relative cost of learning an accurate ranking of units will be negligible for a sufficiently large budget.

\begin{thm}
\label{thm:worst_case_ranking}
Suppose ULA treats $K$~units and ULA's value is $(N\delta')$-Lipschitz in unit-level statistics. By appropriate choice of the algorithm and the number of samples as detailed in the proof, the relative decrease in ULA's value due to the cost of learning a ranking of units is $O\big(K^{-8/9} M^{2/3} N^{-2/9}\big)$ with high probability, where $M$ is the total number of units and~$N$ is the number of individuals per unit. As a result, for $K = \Omega(M^{3/4}/N^{1/4})$, the effect of learning starts to become negligible asymptotically.
\end{thm}
\begin{proof}
    From $m$ random units, sample $n_2$ true welfare from each unit. Without loss of generality, assume ULA allocates resources to the units with the smallest statistics. We assume the $\xi$-quantile of the unit-level statistics is known and denoted by~$U_{(\xi)}$. One can later verify that quantiles can effectively be estimated with our sampled data without incurring a significant additional cost. Define $\Delta \coloneqq \max \, \{\beta,  z_\alpha / \sqrt{n_2}\}$. Targeting $K$~units, define $\gD$ as the product of two distributions~$\gD_1$ and~$\gD_2$, where $\gD_1$ is a uniform distribution over any unit with a statistic less than $U_{(K/M)}$ and $\gD_2$ is a uniform distribution over any unit with a statistic more than $U_{(K/M)} + \Delta$. Since \cref{assump:statistics_not_concentrated} limits the concentration of $U_k$s, the $m$~random samples from the units give $n = \Theta\big(\frac{K}{M} m^2\big)$ samples from~$\gD$. 
    
    We minimize the empirical $\beta$-margin risk of \cref{lem:ranking_generalization_bound} using finite-sample evaluations of welfare. Since by definition of~$\gD$ this problem is realizable, assuming empirical risk minimization can be performed perfectly, we have $\hat{R}_\beta(h) = 0$. Next, we bound $R(h)$. For notational brevity, we normalize ULA's value by $N \delta'$, where $\delta'$ is the maximum treatment effect an individual may experience. 
    \begin{itemize}
        \item For $K'$~units ($K' \le K$) in the support of~$\gD_2$ to substitute $K'$~units in the support of~$\gD_1$, at least $K'^2$ pairwise misranking is required. Since every pairwise misranking at least contributes $1/(K M)$ to the risk, we have $K' \le \sqrt{K M R(h)}$, where \cref{lem:ranking_generalization_bound} requires $R(h) = O\big(\frac{1}{\beta}\sqrt{\frac{M}{K m^2}}\big)$ with high probability. Therefore, the change in ULA's value due to misranking is $O(K^{1/4} M^{3/4} m^{-1/2} \beta^{-1/2})$.
    
        \item Assuming $c_2 \approx c$, the cost of sampling wastes a budget equivalent to treating $m n_2$ individuals. Hence, the change in ULA's (normalized) value due to the sampling cost is $O(m n_2 / N)$.
    
        \item Finally, assuming ULA's (normalized) value is $1$-Lipschitz in unit-level statistics, the change in ULA's value due to misranking of individuals with their statistic in $[U_{(K/M)}, U_{(K/M)} + \Delta]$, is $O\big(\Delta (M \lambda \Delta)\big) = O(M \Delta^2) = O(M \beta^2 + M/n_2)$.
    \end{itemize}
    Adding these all together, the decrease in ULA's (normalized) value is bounded by
    \begin{equation*}
        O\big(K^{1/4} M^{3/4} m^{-1/2} \beta^{-1/2} + m n_2 / N  + M \beta^2 + M/n_2\big)
        \,.
    \end{equation*}
    Choose $\beta = K^{1/10} M^{-1/10} m^{-1/5}$ and $n_2 = \sqrt{M N / m}$. Then, the above bound reduces to
    \begin{equation*}
        O\big(K^{1/5} M^{4/5} m^{-2/5} + \sqrt{M m / N}\big)
        \,.
    \end{equation*}
    Finally, choose $m = K^{2/9} M^{1/3} N^{5/9}$. Then the above bound simplifies to
    \begin{equation*}
        O\big(K^{1/9} M^{2/3} N^{-2/9}\big)
        \,.
    \end{equation*}
    Since ULA's (normalized) value is $\Theta(K)$, the relative change in ULA's value is bounded by $O\big(K^{-8/9} M^{2/3} N^{-2/9}\big)$. Hence, this effect becomes negligible when $K$ exceeds $\Omega(M^{3/4}/N^{1/4})$.
\end{proof}

In a comparison with naive sampling from all units (\cref{sec:learning_with_sampling_all}), a direct calculation shows that \cref{thm:worst_case_ranking} provides more efficient learning if $N = O(M^3)$. In particular, for $M = \Theta(N)$, the effect of learning a ranking according to \cref{thm:worst_case_ranking} becomes negligible when $K = \Omega(\sqrt{M})$, while naive sampling from all units requires $K = \Omega(M^{2/3})$. It is also worth noting that the guarantee in \cref{thm:worst_case_ranking} holds with high probability, which is a stronger result than the one presented in \cref{sec:learning_with_sampling_all} that holds in expectation.

\cref{thm:worst_case_ranking} has a pessimistic view of where incorrect ranking can occur. Optimistically, if for any pair of units, the probability of mistake is $O(R(h))$, then a simple calculation shows sampling from~$O(\log(M))$ units is enough to get a sufficiently accurate ranking for ULA.



\clearpage
\newpage
\section{Additional Statements}

\begin{proposition}
\label{prop:ila_bound_is_tight}
In ILA with an $\epsilon$-accurate predictor, if there are at least $I = \lfloor \frac{B - p(\epsilon)}{c} \rfloor$ individuals with a welfare of $(1 - \delta - 2\epsilon)$ or less, the ILA's maximum value of 
\begin{equation*}
    \max \Big\{ \frac{B - p(\epsilon)}{c},\; 0 \Big\} \, \delta
\end{equation*}
can be realized.
\end{proposition}
\begin{proof}
To see this, first, observe that for the $I$~individuals with the lowest welfare, the estimated welfare will be less than or equal to~$(1 - \delta - \epsilon)$. This ensures the $I^\text{th}$ estimated welfare after sorting, $\hat{w}_{s(I)}$, will be less than or equal to~$(1 - \delta - \epsilon)$. Therefore, the true welfare of $s(1), \dots, s(I)$ should be less than or equal to~$(1-\delta)$. For such a group of targeted individuals, the treatment effect is at its maximum of~$\delta$. 
\end{proof}

\begin{thm}[Sufficient conditions for a dominant ULA in case of a sublinear budget]
\label{thm:dominant_ula_suff_conditions_sublinear_budget}
Consider ULA with a $(q,q')$-within-unit allocation and a budget~$B=o(C)$. Assume $q' \le (1 - \bar{\rho})q$. Define normalized inequality $\widehat{G}_\rho \coloneqq G_\rho / (1 - \bar{\rho}) \in [0, 1]$, and consider inequality threshold
\begin{equation*}
    \widehat{G}_\rho^{(1)} \coloneqq 1 - \frac{q}{\bar{\rho}} \frac{1 - \bar{\rho}}{1 - q}
    \,.
\end{equation*}
\begin{itemize}
    \item If $q > \bar{\rho}$, within-unit allocation is effective and ULA achieves $\big(1 - q'/(1-q)\big)$ of the maximal value of~$B \frac{\delta}{c}$. This implies $\big(1 + O(q')\big)\!\cdot\!{\rm ULA} \succeq {\rm ILA}$.
    \\If further $p(\epsilon)/B > q'/(1-q)$, we have ${\rm ULA} \succ {\rm ILA}\,$.

    \item If $q \le \bar{\rho}$ but a minimal inequality is present as $\widehat{G}_\rho > \widehat{G}_\rho^{(1)}$, within-unit allocation is still effective and similar results hold true as the previous case.

    \item If none of the above conditions hold, but $p(\epsilon)$ consumes a significant portion of the budget as $p(\epsilon)> B \frac{q_c - (q-q')}{1-q}$, we have ${\rm ULA} \succ {\rm ILA}\,$.
\end{itemize}
\end{thm}
\begin{proof}
    First of all, note that for a sublinear budget, the condition $K \le M (1-\bar{\rho})$ of \cref{lem:ub_loss_unit} (and so \cref{cor:lb_V_unit_sublinear_budget}) is met. In the first two cases, within-unit allocation is effective, and \cref{cor:lb_V_unit_sublinear_budget} shows ULA at max loses a value of $B \frac{\delta}{c} \big(\frac{q'}{1-q}\big)$ compared to the optimal. On the other hand, \cref{eq:V_ind_ub} shows ILA at least loses a value of $p(\epsilon) \frac{\delta}{c}$. Therefore, ULA's value will exceed ILA if $B \frac{q'}{1-q} < p(\epsilon)$. For a least effective within-unit allocation, \cref{cor:lb_V_unit_sublinear_budget} shows that ILA would lose a maximum value of $B \frac{\delta}{c} \big(\frac{q_c - (q-q')}{1-q}\big)$. Therefore, a similar argument shows that ULA will dominate ILA if $B \big(\frac{q_c - (q-q')}{1-q}\big) < p(\epsilon)$.
\end{proof}

\begin{proposition}
\label{prop:increasing_deriv}
If 1) $\tau$ is strictly decreasing, 2) $\tau \in \C^1$, 3) $\tau$ is either convex or concave, and 4) $\od{\tau}{w}$ is convex, then $r(w)$ of \cref{prop:decreasing_deriv} will be increasing in~$w$ for $w_c = 0$ and every valid choice of~$a$ and~$b$. Therefore, under these conditions, $\tau$ won't have \cref{prop:decreasing_deriv}.
\end{proposition}
\begin{proof}
    Since $\tau$ is strictly decreasing, for every choice of~$b > 0$, $r(w)$ is well-defined. Denote the derivative of~$\tau$ by~$\tau'$. Taking the derivative of~$r$ w.r.t.~$w$ and rearranging the terms, we obtain
    \begin{equation*}
        \frac{1}{\big(\tau(w) - \tau(w + b)\big)^2} \Big\{ 
        \tau'(w) \big( \tau(w + a) - \tau(w + b) \big)
        - \tau'(w + a) \big( \tau(w) - \tau(w + b) \big)
        + \tau'(w + b) \big( \tau(w) - \tau(w + a) )
        \Big\}
        \,.
    \end{equation*}
    The sign of the derivative is determined by its numerator, denoted by~$n(w)$. Define 
    \begin{align*}
        c_1(w) &\coloneqq \tau(w + a) - \tau(w + b)\,, \\
        c_2(w) &\coloneqq \tau(w) - \tau(w + a)\,.
    \end{align*}
    Then, the numerator of the derivative can be written as
    \begin{equation*}
        n(w) = c_1(w) \tau'(w) + c_2(w) \tau'(w + b) - \big(c_1(w) + c_2(w)\big)\tau'(w + a)
        \,.
    \end{equation*}
    Note that a strictly decreasing~$\tau$ and $b \ge a > 0$ implies $c_1$ and~$c_2$ are positive. Then, for a convex~$\tau'$, Jensen's inequality implies
    \begin{equation}
    \label{eq:_proof_increasing_deriv_n_lb}
        n(w) \ge  \big(c_1(w) + c_2(w)\big) \tau'\big(w + b \frac{c_2(w)}{c_1(w) + c_2(w)} \big) - \big(c_1(w) + c_2(w)\big)\tau'(w + a)
        \,.
    \end{equation}
    \begin{itemize}
        \item For a concave~$\tau$ and $b \ge a > 0$, the Jensen's inequality gives $\tau(w + a) \ge (1 - a/b) \tau(w) + (a/b) \tau(w + b)$. This allows us to write
        \begin{equation}
        \label{eq:_proof_increasing_deriv_ratio_jensen}
            \frac{c_2(w)}{c_1(w) + c_2(w)} = \frac{\tau(w) - \tau(w + a)}{\tau(w) - \tau(w + b)} \le \frac{\tau(w) - (1 - a/b)\tau(w) - (a/b)\tau(w + b)}{\tau(w) - \tau(w + b)} = \frac{a}{b}
            \,.
        \end{equation}
        Plugging this into \cref{eq:_proof_increasing_deriv_n_lb} and using the fact that $\tau'$ is decreasing since $\tau$ is concave, we obtain
        \begin{equation}
        \label{eq:_proof_increasing_deriv_mid_ge}
            \tau'\big(w + b \frac{c_2(w)}{c_1(w) + c_2(w)} \big) \ge \tau'(w + a)
            \,.
        \end{equation}

        \item For a convex~$\tau$ and $b \ge a > 0$, the Jensen's inequality gives $\tau(w + a) \le (1 - a/b) \tau(w) + (a/b) \tau(w + b)$. So, \cref{eq:_proof_increasing_deriv_ratio_jensen} holds in the reverse direction. Plugging this into \cref{eq:_proof_increasing_deriv_n_lb} and using the fact that $\tau'$ is increasing since $\tau$ is convex, we again obtain \cref{eq:_proof_increasing_deriv_mid_ge}.
    \end{itemize}
    Since in both cases \cref{eq:_proof_increasing_deriv_mid_ge} holds true, 
    \cref{eq:_proof_increasing_deriv_n_lb} implies $n(w) \ge 0$, and the proof is complete. 

\end{proof}

\begin{thm}
\label{thm:heter_ula_lb}
Consider ULA with a $(q, q')$-within-unit allocation and a budget~$B$ that is no more than the cost of treating $M \barT'/\delta'$ units. Under \cref{assump:gamma_bounded_density}, suppose $\tau$ has \cref{prop:decreasing_deriv} for $w_c$, $a = (1 - q - q')/\gamma$, and $b = (1 - 2q')/\gamma$. Define $T'_c \coloneqq \gamma \Gamma_b(w_c) + q' \delta'$
and two inequality thresholds
\begin{equation}
\label{eq:heter_ula_lb_ineq_th}
    \widetilde{G}_T^{(1)} \coloneqq 1 - \frac{\barT'}{T'_c}
    \;\;\le\;\; 
    \widetilde{G}_T^{(2)} \coloneqq 1 - \frac{1}{4} \frac{\barT'}{T'_c} - \frac{3}{4} \frac{\barT'}{\delta'}
    \,.
\end{equation}
Then, if any of the following conditions holds, for every treated unit~$k$, we will have $T'_k \ge T'_c$ :
\begin{itemize}
    \item If $G_T > \widetilde{G}_T^{(2)}$ and the budget is not excessively large (not surpassing the cost of treating $M \barT'/2\delta'$~units).

    \item If $G_T > \widetilde{G}_T^{(1)}$ and the budget does not exceed
    \begin{equation}
    \label{eq:heter_ula_lb_small_budget}
        \frac{B}{C} < (1 - q)\frac{\barT' - T'_c + \max \, \{\barT' - T'_c, \, G_T T'_c\} }{2(\delta' - T'_c)}
        \,.
    \end{equation}
\end{itemize}
Then, defining $T'_{\rm eff} \coloneqq \barT'/(1 - G_T)$ and neglecting $O(B^2/C^2)$, if any of the above conditions holds, we have
\begin{equation*}
    V_{\rm unit} \ge K Q_\tau (T'_{\rm eff} - q'\delta'; \; q, q', \gamma)
    \,.
\end{equation*}
\end{thm}
\begin{proof}
    Denote the profile of all $T'_k$s by $\vT'$. Let $s(\cdot)$ be the ascending sort of $T'_k$s. For notational simplicity, assume $s(k) = k$ without loss of generality. We also drop the dependency of $Q_\tau(\cdot; q, q', \gamma)$ on $q$, $q'$, and $\gamma$ as these are the constants of the problem. \cref{lem:min_T_gamma_bounded} guarantees that unit~$k$ if treated, provides a value of at least $Q_\tau(T'_k - q'\delta')$. Then, treating $K$~units ensures a value of 
    \begin{equation*}
        V_\text{unit} \ge \widetilde{V}(\vT') \coloneqq \sum_{k = M - K + 1}^M Q_\tau(T'_k - q' \delta')
        \,.
    \end{equation*}
    Next, we look for~$\vT'$ minimizing~$\widetilde{V}(\vT')$ while satisfying a mean of~$\barT'$ and a Gini index of~$G_T$. The proof has two parts: First, we find sufficient conditions under which all the treated units have a~$T'$ larger than or equal to
    \begin{equation*}
        T'_c \coloneqq \gamma \Gamma_{(1 - 2q')/\gamma}(w_c) + q' \delta'
        \,.
    \end{equation*}
    Then, for such units, we further lower bound $\widetilde{V}(\vT')$. 
    
    \paragraph{Ensuring for All Treated Units, $T'_k \ge T'_c$.}
    As the first step, we relax the problem, allowing for unit profiles with a Gini index of~$G_T$ or more. This requires
    \begin{equation}
    \label{eq:_proof_lb_V_unit_heter_gini_constraint}
        2 \sum_{k \in [M]} k T'_k \ge M (M + 1) \barT' + M^2 \barT' G_T
        \,.
    \end{equation}
    Consider two units~$l$ and~$u$ ($u > l$). For a fixed mean, the Gini index or equivalently the left-hand side of \cref{eq:_proof_lb_V_unit_heter_gini_constraint} can be increased if we increase~$T'_u$ and decrease~$T'_l$ by the same amount. Now assume there are $K_+$~units with $T'_k \ge T'_c$. Neglecting two units at max, for a fixed~$K_+$, the Gini index will be maximized when units are concentrated on the boundaries $T'_\text{min}$, $T'_c$, and $T'_\text{max}$. We denote the number of units on these boundaries by $K_0$, $K_1$, and $K_2$. These numbers should be integer but by neglecting at max three units, we can assume they can take any nonnegative real value. We further relax the problem and set $T'_\text{min} = 0$ and $T'_\text{max} = \delta'$. We do a proof by contradiction. If any of the $K$~treated units has $T'_k < T'_c$, we should have $K > K_+$. Since $K_+ \ge K_2$, necessarily $K > K_2$. The mean constraint requires $K_1 T'_c + K_2 \delta' = M \barT'$. Using this and $K_0 + K_1 + K_2 = M$, we can solve for~$K_0$ and~$K_1$ in terms of~$K_2$:
    \begin{align}
    \label{eq:_proof_lb_V_unit_heter_K_0_K_1}
        K_1 &= \frac{M \barT' - K_2 \delta'}{T'_c}\,, \\
        K_0 &= \frac{M(T'_c - \barT') + K_2(\delta' - T'_c)}{T'_c}
        \,.
    \end{align}
    Then, the nonnegativity of $K_0$~and~$K_1$ requires
    \begin{equation}
    \label{eq:_proof_lb_V_unit_heter_K_2_nonneg}
        \frac{\barT' - T'_c}{\delta' - T'_c} \le \frac{K_2}{M} \le  \frac{\barT'}{\delta'}
        \,.
    \end{equation}
    The Gini index constraint also imposes a lower bound on~$K_2$. For a unit profile concentrated on the boundaries, \cref{eq:_proof_lb_V_unit_heter_gini_constraint} can be written as
    \begin{equation*}
        (K_0 + 1 + M - K_2) K_1 T'_c + (M - K_2 + 1 + M) K_2 \delta' \ge M(M + 1) \barT' + M^2 \barT' G_T
        \,.
    \end{equation*}
    Plugging $K_0$ and~$K_1$ from \cref{eq:_proof_lb_V_unit_heter_K_0_K_1} into this and doing a direct calculation, we obtain
    \begin{equation}
    \label{eq:_proof_lb_V_unit_heter_quadratic}
        -\Big(\frac{K_2}{M}\Big)^2 \frac{\delta'}{\barT'} + 2 \Big(\frac{K_2}{M}\Big) + \frac{T'_c(1 - G_T) - \barT'}{\delta' - T'_c} \ge 0
        \,.
    \end{equation}
    If $B=o(c)$ and consequently $K/M = o(1)$, asymptotically this condition requires $G_T \le \widetilde{G}_T^{(1)} \coloneqq 1 - \barT/T'_c$. Therefore, in the case of a sublinear budget, $G_T > \widetilde{G}_T^{(1)}$ results in contradiction and ensures for no treated unit $T'_k < T'_c$. In general, since $K_2/M \le \barT'/\delta'$ (\cref{eq:_proof_lb_V_unit_heter_K_2_nonneg}), the quadratic function of \cref{eq:_proof_lb_V_unit_heter_quadratic} is increasing in the valid range of~$K_2$. Under the assumption $K/M \le \barT'/\delta'$, then $K > K_2$ requires \cref{eq:_proof_lb_V_unit_heter_quadratic} to be holding for $K_2 = K$ as well. Now consider two cases:
    \begin{itemize}
        \item If the budget is not excessively large and $K/M \le \barT'/(2\delta')$, \cref{eq:_proof_lb_V_unit_heter_quadratic} for $K_2 = K = \barT'/(2\delta')$ requires
        \begin{equation*}
            G_T \le \widetilde{G}_T^{(2)} \coloneqq 1 - \frac{1}{4} \frac{\barT'}{T'_c} - \frac{3}{4} \frac{\barT'}{\delta'}
            \,.
        \end{equation*}
        Therefore, $G_T > \widetilde{G}_T^{(2)}$ results in a contradiction.
    
        \item Neglecting the squared term in \cref{eq:_proof_lb_V_unit_heter_quadratic}, $K/M$ should be at least $(\barT' - T'_c(1 - G_T))/(2\delta' - 2T'_c)$, which is binding only if $G_T > \widetilde{G}_T^{(1)} \coloneqq 1 - \barT'/T'_c$. Furthermore, \cref{eq:_proof_lb_V_unit_heter_K_2_nonneg} and $K \ge K_2$ requires $K/M$ to be also no less than $(\barT' - T'_c)/(\delta' - T'_c)$. Putting these together, 
        \begin{equation*}
            \frac{K}{M} \ge \frac{1}{2}\frac{\barT' - T'_c + \max \, \{\barT' - T'_c, \, G_T T'_c\} }{\delta' - T'_c}
            \,.
        \end{equation*}
        Hence, a $K$ less than the above amount results in a contradiction. 
    \end{itemize}
    This completes the first part of the proof.
    
    \paragraph{Lower Bounding $\widetilde{V}(\vT')$.}
    We next lower bound $\widetilde{V}(\vT')$, under the conditions that $T'_k \ge T'_c$ for $k > M - K$, subject to the mean and Gini index constraints. As the first step, we further relax the Gini constraint of \cref{eq:_proof_lb_V_unit_heter_gini_constraint} and require
    \begin{equation}
    \label{eq:_proof_lb_V_unit_heter_gini_constraint_relaxed}
        2 \sum_{k = 1}^{M - K} k T'_k + 2 M \sum_{k=M - K + 1}^M T'_k \ge M (M + 1) \barT' + M^2 \barT' G_T
        \,.
    \end{equation}
    Let $\vT'^*$ be the optimal solution of the relaxed problem. We now construct a solution $\vT'$ from $\vT'^*$ that does not increase $\widetilde{V}$ while maintaining a similar mean and meeting the relaxed Gini constraint of \cref{eq:_proof_lb_V_unit_heter_gini_constraint_relaxed} with a lower or equal margin. As the first step, construct a new solution as
    \begin{equation*}
        T'_k = \begin{cases}
            T'^*_k\,, & k \le M - K\,, \\
            \frac{1}{K} \sum_{k > M - K} T'^*_k & \text{o.w.}
        \end{cases}
    \end{equation*}
    This solution maintains the same mean and meets the relaxed Gini constraint with the same margin. We argue $\widetilde{V}(\vT') \le \widetilde{V}(\vT'^*)$: First, we show $Q_\tau(t)$ is a convex function for $t \ge T'_c - q' \delta'$ or equivalently $T' \ge T'_c$. To see this, note that 
    \begin{equation}
    \label{eq:_proof_heter_ula_lb_convex_Q}
        \od{Q_\tau}{t} = (\Gamma_b^{-1})'(t/\gamma) \, \Gamma'_a\big( \Gamma^{-1}_b(t/\gamma) \big) = \frac{\Gamma'_a(w)}{\Gamma'_b(w)} = \frac{\tau(w) - \tau(w+a)}{\tau(w) - \tau(w+b)}
        \,,
    \end{equation}
    where $w = \Gamma^{-1}_b(t/\gamma)$. Then \cref{prop:decreasing_deriv} implies $\od{Q_\tau}{t}$ is decreasing in~$w$ or equivalently increasing in~$t$ for $w \le w_c$ or equivalently $t = T' - q' \delta' \ge \gamma \Gamma_b(w_c) = T'_c - q' \delta'$. Since we have already shown $T'_k \ge T'_c$ for all of the top~$K$ units, Jensen's inequality implies $\widetilde{V}(\vT') \le \widetilde{V}(\vT'^*)$. Therefore, if $\vT'^*$ was optimal, the constructed $\vT'$ should be optimal as well.
    
    In the second step, choose any two units $l, u$ from $[M - K]$ such that $T'_l < T'u$. Decrease~$T'_l$ and increase~$T'_u$ for the same amount while the order of units is preserved. Such a change will increase the margin of the relaxed Gini index constraint but does not change the mean of the units and the objective. By repetitively applying this operation, we obtain an optimal solution where all units except for one are concentrated around either~$0$ or~$T'_u \coloneqq \frac{1}{K} \sum_{k > M - K} T'_k$. Hence, we can represent this solution by two numbers~$K_0$ and~$K_1$ which are the number the number of units with~$T'_k$ of~$0$ and~$T'_u$. These numbers should be integers however neglecting the value of two units in maximum, we can assume they are nonnegative real numbers satisfying the sum constraint $K_0 + K_1 = M$. The mean constraint also requires $K_1 T'_u = M \barT'$. Solving for $K_0$ and $K_1$ in terms of~$T'_u$ we obtain $K_1/M = \barT'/T'_u$ and $K_0/M = 1 - \barT'/T'_u$. The nonnegativity of~$K_0$ and $K_1 \ge K$ imposes
    \begin{equation*}
        \max \, \{\barT', T'_c\} \le T'_u \le \frac{M}{K} \barT
        \,.
    \end{equation*}
    The relaxed Gini constraint also imposes another lower bound on~$T'_u$. From \cref{eq:_proof_lb_V_unit_heter_gini_constraint_relaxed} we have
    \begin{equation*}
        (K_0 + 1 + M - K) (K_1 - K) T'_u + 2M K T'_u \ge M(M + 1) \barT' + M^2 \barT' G_T
        \,.
    \end{equation*}
    Substituting $K_0$ and $K_1$ in terms of~$T'_u$ and simplifying equation, we obtain
    \begin{equation*}
        \frac{K^2 - K}{M^2} \Big(\frac{T'_u}{\barT'}\Big)^2 + (1 - G_T) \Big(\frac{T'_u}{\barT'}\Big) - 1 \ge 0
        \,.
    \end{equation*}
    Then a direct calculation shows $T'_u$ should be at least as large as
    \begin{equation}
    \label{eq:_proof_lb_V_unit_heter_lb}
        T'_u \ge \frac{2 \barT'}{1 - G_T + \sqrt{(1 - G_T)^2 + 2(K^2 - K)/M^2}}
        \,.
    \end{equation}
    Using this lower bound in $\widetilde{V}(\vT') = K Q_\tau(T'_u - q' \delta')$ then completes the proof. 
    
\end{proof}



\clearpage
\newpage
\section{Missing Proofs}
\label{sec:missing}

\theoremstyle{plain}
\newtheorem*{thm:heter_dominant_ula}{Theorem \ref{thm:heter_dominant_ula_short}}
\begin{thm:heter_dominant_ula}[Sufficient conditions for a dominant ULA in case of a heterogeneous effect]
\label{thm:heter_dominant_ula}
Consider ULA with a $(q, q')$-within-unit allocation and a budget~$B$ that is no more than the cost of treating $M \barT'/\delta'$ units. Under \cref{assump:gamma_bounded_density} and conditions of \cref{thm:heter_ila_ub}, suppose an elbow-shaped effect (\cref{fig:elbow_tau}) that satisfies \cref{prop:decreasing_deriv} for $w_c = 1 - \delta - a$, $a = (1 - q - q')/\gamma$, and $b = (1 - 2q')/\gamma$. Then $T'_c \le \delta' - q^2/2$. Consider inequality thresholds
\begin{equation*}
    G_T^{(1)} \coloneqq 1 - \frac{\barT'}{\delta' - q^2/(2\gamma)}
    \le
    G_T^{(2)} \coloneqq 1 - \frac{1}{4} \frac{\barT'}{\delta' - q^2/(2\gamma)} - \frac{3}{4} \frac{\barT'}{\delta'}
    \,.
\end{equation*}
If either $G_T > G_T^{(2)}$, or $G_T > G_T^{(1)}$ and the budget meets \cref{eq:heter_dominant_ula_budget_constraint}, then we have $T'_k \ge \delta$ for every treated unit~$k$. Furthermore,
\begin{itemize}
    \item Defining $g \coloneqq (G_T - G_T^{(1)})/(1 - G_T)$ and
    \begin{equation*}
        \alpha \coloneqq 1 - \frac{1-q-q'}{4\gamma} - \frac{1}{2(q - q')} g - \frac{\gamma}{4} \frac{1 - 2q}{(q - q')^3} g^2
        \,,
    \end{equation*}
    and neglecting $O(l^2)$, ULA achieves $(1 - q'/(1-q))(1 - \alpha l/\delta)$ of the optimal value.
    \item Neglecting $O(q'^2)$, if
    \begin{equation*}
        \frac{p(\epsilon)}{B} > \frac{q'}{1-q} + \alpha \frac{l}{\delta} - \frac{l}{\delta} \frac{1}{8 \gamma} \big(\frac{B}{C} + \ugamma \epsilon/2 \big)
        \,,
    \end{equation*}
    we will further have ${\rm ULA} \succ {\rm ILA}\,$. 
\end{itemize}
\end{thm:heter_dominant_ula}
\begin{proof}
\label{proof:heter_dominant_ula}
    Since $\tau$ is $l$-Lipschitz when $w < 1 - \delta$, for $w_c = 1 - \delta - a$, we have
    \begin{align*}
        T'_c = \gamma \Gamma_b(w_c) + q'\delta' 
        &= \gamma \Gamma_a(w_c) + \gamma (\delta - (b-a)/2)(b - a) + q'\delta' \\
        &\le \gamma \delta' a + \gamma (\delta' - (b-a)/2)(b - a) + q'\delta'
        = \gamma \delta' b - \gamma (b-a)^2/2 + q' \delta' \\
        &\le \delta' - q^2/(2\gamma)
        \,.
    \end{align*}
    Plugging this into \cref{eq:heter_ula_lb_ineq_th} give inequality thresholds $G_T^{(1)}$ and $G_T^{(2)}$.
    
    Examining \cref{eq:heter_ula_lb_small_budget}, if $\barT' - T'_c \ge G_T T'_c$, then the upper bound on the budget decreases with increasing $T'_c$. On the other hand, if $\barT' - T'_c < G_T T'_c$, utilizing the property of the Gini index that $G_T$ cannot exceed $1 - \barT'/\delta'$, one can verify that the upper bound is again decreasing in~$T'_c$. Therefore, to satisfy \cref{eq:heter_ula_lb_small_budget}, it suffices to have
    \begin{align}
        \frac{B}{C} &< (1 - q)\frac{\barT' - \delta' + q^2/(2\gamma) + \max \, \{\barT' - \delta' + q^2/(2\gamma), \, (\delta' - q^2/(2\gamma)) G_T\} }{2(\delta' - \delta' + q^2/(2\gamma))} \nonumber \\
        \label{eq:heter_dominant_ula_budget_constraint}
        &=\frac{1}{2} (1-q)(2\gamma \delta'/q^2 - 1) \big(- G_T^{(1)} + \max \, \{-G_T^{(1)}, G_T\}\big)
        \,.
    \end{align}

    In the proof of \cref{thm:heter_ula_lb} (in particular, \cref{eq:_proof_heter_ula_lb_convex_Q}) we showed 
    \begin{equation*}
        \od{Q_\tau}{t} = \frac{\tau(w) - \tau(w + a)}{\tau(w) - \tau(w + b)}
        \,,
    \end{equation*}
    for $w = \Gamma^{-1}_b(t/\gamma)$. Then we argued 
    \cref{prop:decreasing_deriv} implies $Q_\tau(t)$ is a convex function for $t \ge T'_c - q'\delta'$. Now we further show that $Q_\tau$ is strongly convex in the regime of interest where $w + b \ge w_c$ and $w + a \le w_c$:
    \begin{align}
        \od[2]{Q_\tau}{t} &= \od{w}{t} \od{}{w}\Big(\od{Q_\tau}{t}\Big) \nonumber \\
        &= \frac{1}{\gamma} \frac{
            -\tau'(w)\big(\tau(w+a) - \tau(w+b)\big)
            + \tau'(w+a)\big(\tau(w) - \tau(w+b)\big)
            - \tau'(w+b)\big(\tau(w) - \tau(w+a)\big)
        }{\big(\tau(w) - \tau(w+b)\big)^3} \nonumber \\
        &\ge \frac{1}{\gamma} \frac{
            \frac{l}{2}\big(\tau(w+a) - \tau(w+b)\big)
            - l \big(\tau(w) - \tau(w+b)\big)
            + \big(\tau(w) - \tau(w+a)\big)
        }{\big(\tau(w) - \tau(w+b)\big)^3} \nonumber \\
        &= \frac{1}{\gamma} \frac{
            \big(\tau(w) - \tau(w+a)\big) 
            - \frac{l}{2} \big(\tau(w) - \tau(w+b)\big)
            + O(l^2)
        }{\big(\tau(w) - \tau(w+b)\big)^3} \nonumber \\
        \label{proof_heter_cor_suff_sec_deriv}
        &\ge \frac{l}{2\gamma} \frac{
            2 a - b
        }{(b - a)^3}
        + O(l^2)
        \,.
    \end{align}
    Here we denoted the derivative of~$\tau$ by~$\tau'$ and used $l \ge |\tau'(w)| \ge l/2$ for $w \le w_c + a$ and $\tau'(w + b) = -1$. Neglecting $O(l^2)$, this second derivative is positive for $a > b/2$ or equivalently $q < 1/2$. We denote the lower bound of \cref{proof_heter_cor_suff_sec_deriv} by~$\beta_2$. Using a similar technique, we can lower bound the first derivative as well:
    \begin{equation*}
        \od{Q_\tau}{t} \ge \frac{l}{2} \frac{a}{b - a} + O(l^2)
        \,.
    \end{equation*}
    Neglecting $O(l^2)$, we denote this lower bound by~$\beta_1$. At $w = w_c$, we have
    \begin{equation*}
        Q_\tau(T'_c - q'\delta') = \gamma \Gamma_a(w_c) \ge \gamma a (\delta + la/4)
        \,.
    \end{equation*}
    Also note that for a nonnegative~$\widetilde{G}_T^{(1)}$, we have
    \begin{equation*}
        T'_\text{eff} - T'_c = \frac{T'_c}{1 - G_T} \big(\frac{\barT'}{T'_c} - 1 + G_T \big) \ge \barT' \frac{G_T - \widetilde{G}_T^{(1)}}{1 - G_T} \ge \barT' \frac{G_T - G_T^{(1)}}{1 - G_T}
        \,.
    \end{equation*}
    Now we have all the pieces to lower bound $Q_\tau(T'_\text{eff} - q'\delta')$:
    \begin{align*}
        Q_\tau(T'_\text{eff} - q'\delta') &\ge Q_\tau(T'_c - q'\delta') + \beta_1 (T'_\text{eff} - T'_c) + \frac{1}{2} \beta_2 (T'_\text{eff} - T'_c)^2 \\
        &\ge \gamma a \Big( \delta + \frac{la}{4} + \frac{\beta_1}{\gamma} \barT' \frac{G_T - G_T^{(1)}}{1 - G_T} + \frac{\beta_2}{2\gamma} \big(\barT' \frac{G_T - G_T^{(1)}}{1 - G_T} \big)^2  \Big)
        \,.
    \end{align*}
    Plugging $\beta_1$, $\beta_2$, $a$, and $b$ into the above equation and using $K \approx \frac{B}{N c (1 - q)}$ in \cref{thm:heter_ula_lb} will complete the proof of the near-optimality of ULA.
    
    We next drop $O(q'^2)$ from our calculation and obtain sufficient conditions for the dominance of ULA. If ULA is not dominant, necessarily,
    \begin{equation*}
        (B - p(\epsilon)) \frac{\delta'}{c} > B \frac{\delta'}{c}\big(1 - O(l, q')\big)
        \,.
    \end{equation*}
    This requires $p(\epsilon)/B = O(l, q')$. For such a small price of prediction, using \cref{thm:heter_ila_ub}, it is sufficient to have
    \begin{equation*}
        B \delta' \big(1 - \frac{q'}{1-q} - \alpha \frac{\delta}{\delta}\big) > (B - p(\epsilon)) \big(\delta' - \frac{l}{8 \gamma}(B/C + \ugamma \epsilon/2) \big) \approx B \delta' - p(\epsilon) \delta - B \frac{l}{8 \gamma}(B/C + \ugamma \epsilon/2) \big)
        \,.
    \end{equation*}
    Rearranging the terms and solving for $p(\epsilon)/B$ will prove the stated sufficiency condition.

\end{proof}

\printProofs

\end{document}